\pdfoutput=1
\documentclass{article}

\usepackage[main, final]{neurips_2026}

\setcitestyle{numbers,square,comma,compress}
\usepackage[utf8]{inputenc}
\usepackage[T1]{fontenc}

\usepackage{url}
\usepackage{booktabs}
\usepackage{amsfonts}
\usepackage{nicefrac}
\usepackage{microtype}
\usepackage{xcolor}
\usepackage{amsmath, amssymb}
\usepackage{tabularx}
\newcolumntype{Y}{>{\raggedright\arraybackslash}X}
\usepackage{float}
\usepackage{bm}
\usepackage{tikz}
\usepackage{graphicx}
\usepackage{subcaption}
\graphicspath{ {./fig/} }
\usepackage{wrapfig}
\usetikzlibrary{positioning}
\usetikzlibrary{arrows.meta}
\usetikzlibrary{shapes.geometric}
\usetikzlibrary{fit}
\usetikzlibrary{backgrounds}
\usetikzlibrary{patterns}
\usepackage[table]{xcolor}
\definecolor{citecolor}{HTML}{0071BC}
\usepackage[colorlinks, citecolor=citecolor, linkcolor=red, urlcolor=gray]{hyperref}
\usepackage{pgfplots}
\pgfplotsset{compat=1.18}
\definecolor{accent}{HTML}{B42B2B}
\definecolor{accent2}{HTML}{185FA5}
\definecolor{neutral}{HTML}{888880}
\usepackage{booktabs}
\usepackage{caption}
\usepackage{enumitem}
\setlist{topsep=2pt, itemsep=1pt, parsep=0pt, partopsep=0pt}

\title{Emergent Vision-Shaped Thinking in Text-Only Reasoning through Multimodal Training}

\author{%
  Qi Zhang\textsuperscript{1}\thanks{Equal contribution.} \quad
  Bohan Wang\footnotemark[1] \quad
  Qianru Sun\textsuperscript{2} \quad
  Hanwang Zhang\textsuperscript{1} \\
  \textsuperscript{1}Nanyang Technological University \quad
  \textsuperscript{2}Singapore Management University \\
  \texttt{qi015@e.ntu.edu.sg}, \texttt{bohan.wang97@gmail.com}, \\
  \texttt{qianrusun@smu.edu.sg}, \texttt{hanwangzhang@ntu.edu.sg}
}

\begin{document}

\maketitle

\begin{abstract}
Does multimodal training shape how a language model thinks, even when the input is text only?
We test this with targeted causal ablations across 10 multimodal
configurations from 6 model families, with matched text-only backbones and
cross-family transfer controls.
We contrast spatial-imagery tasks that invite mental scene construction with symbol-manipulation tasks solvable by symbolic rules alone.
We identify a vision--language axis in hidden-state space and ablate its projection at inference.
In Chameleon-7B, this ablation drops accuracy on BIG-Bench Hard (BBH) \emph{navigate} by $27.6$ pp while leaving \emph{boolean expressions} intact ($+5.6$ pp), a $33.2$ pp double dissociation.
The effect is direction-specific and layer-localized, extends to
mental-simulation tasks such as mental rotation, paper folding, and
perspective taking, and is not explained by task difficulty, reasoning
depth, or sequential update load.
Together, these results reveal \emph{emergent vision-shaped thinking} (EVT):
text-only reasoning that causally recruits representations introduced or
strengthened by vision--language training, without assuming an image-like
internal format. The measured effect is spatially concentrated at 7--8B
and broadens to non-spatial visual attribute binding in the tested 32B
configuration.
In dual-path unified models (BAGEL, Janus-Pro), the causal handle localizes
to the generation pathway as a compact one-dimensional carrier. The
understanding pathway also encodes spatial information in a higher-rank,
probe-decodable form, but is not the causal handle. To separate
decodability from causal use, we analyze concentration, alignment, and
semantic location. These diagnostics explain why behavior, probing, and
intervention can diverge. Multimodal training thus reshapes the
geometry of text-only reasoning, not just how visual input is processed.
\end{abstract}

\section{Introduction}
Multimodal training has become standard for large language models, producing
diverse architectures for coupling vision with language. These range from shared-vocabulary fusion
(Chameleon)~\citep{chameleon2024} and detached-encoder bridging
(Qwen3-VL)~\citep{bai2025qwen3vl} to vision-fine-tuned backbones
(Liquid)~\citep{wu2024liquid} and dual-path unified models that jointly
train visual generation and understanding
(BAGEL, Janus-Pro)~\citep{deng2025bagel,chen2025januspro}. Yet how such training affects the language backbone itself remains poorly understood. Architectural choices describe where visual information enters
the model. They say little about what the backbone learns to do with it,
or whether the linguistic representations used for text-only reasoning are
reshaped in the process. This question is especially sharp in unified models, where visual generation, visual understanding, and language modeling share a single backbone. But it has not been mechanistically tested at scale across modern multimodal architectures.

Behavioral evidence suggests that multimodal training affects text-only
reasoning. VLMs outperform LLM backbones on text-only spatial
reasoning~\citep{wang2024picture}, and visual training can shift internal
binding even under text-only evaluation on synthetic
tasks~\citep{buzeta2026seeing}. Yet existing evidence is behavioral or
limited to synthetic mid-scale settings. Whether multimodal models rely on
visual-spatial structure during naturalistic text-only reasoning, where
this dependency lives, and how it varies across architectures remain
unknown. We answer these questions causally.

To make the causal question testable, we contrast two families of
text-only tasks. \emph{Spatial-imagery tasks}
have textual inputs but naturally invite mental scene construction.
Examples include navigating a 2D grid described in words and tracking
object locations through textual updates.
\begin{wrapfigure}[20]{r}{0.60\textwidth}
    \centering
    \vspace{-8pt}
    \includegraphics[width=\linewidth]{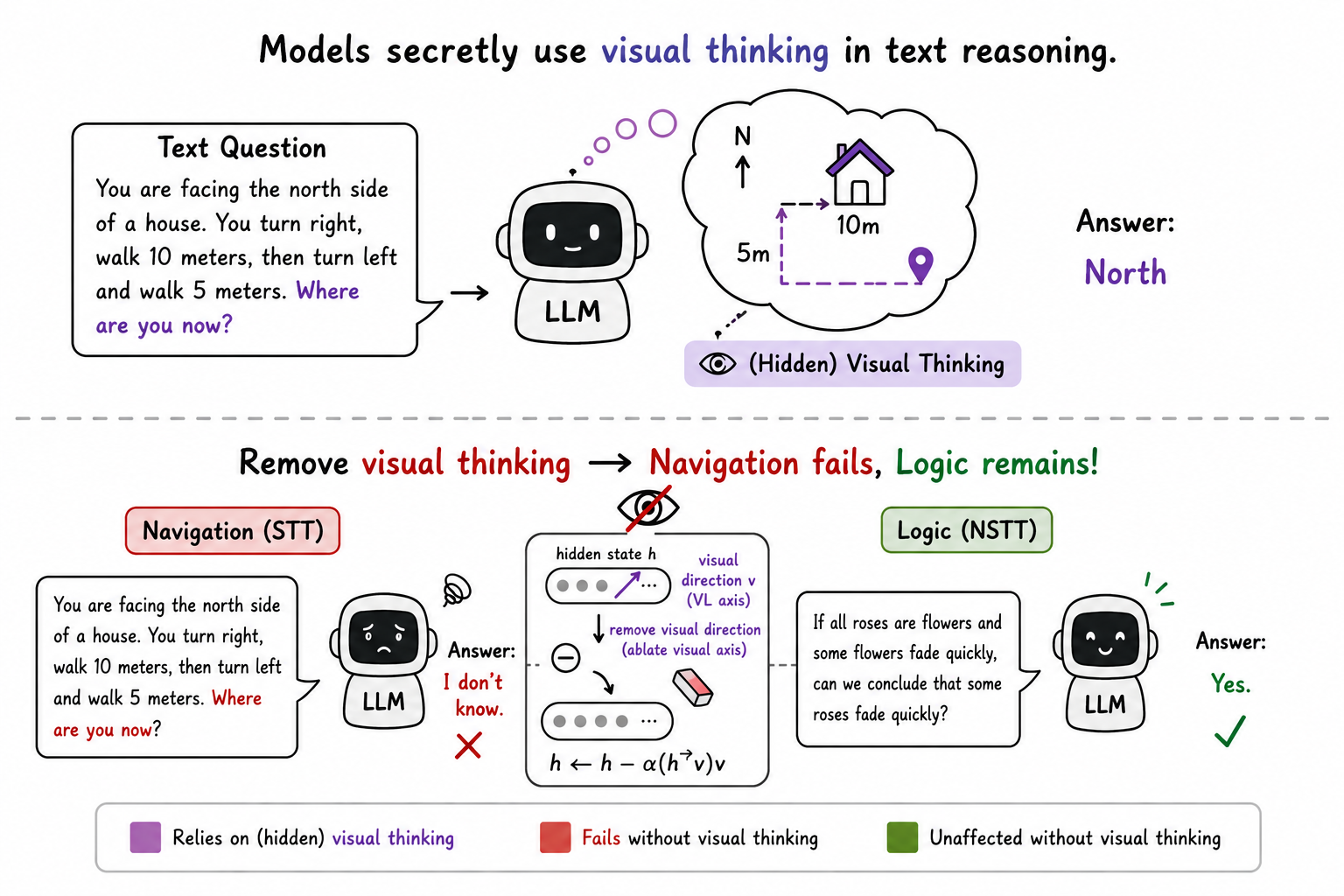}
    \caption{\textbf{Emergent Vision-Shaped Thinking.} Multimodal models use vision-shaped structure even on text-only spatial reasoning. Ablating a vision--language direction selectively damages spatial-imagery tasks (Navigation) while leaving symbol-manipulation tasks (Logic) intact.}
    \label{fig:teaser}
    \vspace{-8pt}
\end{wrapfigure}
\emph{Symbol-manipulation
tasks} are solvable by formal rule operations alone, such as evaluating
boolean expressions or checking logical consistency. Both families are
text-only. What differs is the representational form each task calls
for. If removing a hidden-state direction selectively collapses
spatial-imagery tasks while leaving symbol-manipulation tasks intact,
the model uses that direction to carry visual-spatial structure in
text-only reasoning. This shows functional use, not mere representational presence.

For each model, we identify a vision--language axis in hidden-state
space and ablate its projection during inference
(Figure~\ref{fig:teaser}). Inspired by human spatial imagery, we call this collapse
\emph{emergent vision-shaped thinking} (EVT), meaning text-only reasoning that
causally recruits representations introduced or strengthened by
vision--language training. The term deliberately leaves the internal format
open: our interventions establish causal use of vision-shaped structure,
not the construction of an image-like mental scene
(\S\ref{sec:related}, \emph{Terminology and scope}).
This definition yields three contributions.

\noindent\textbf{(1) Multimodal training reshapes text-only reasoning geometry.}
Across 9 VL-axis-positive configurations, VL-axis ablation selectively
damages spatial-imagery tasks while leaving symbol-manipulation tasks
largely preserved; in the tenth, Chameleon-30B, an input-faithful modality axis is causal at mid/final layers (mid-block DD $=+33.2$ pp; Appendix~\ref{app:cham-fullseq}). The effect is direction-specific, graded across tasks, and localized to a small range of layers. It extends beyond position- and navigation-style
tasks to mental-simulation tasks (mental rotation, paper folding,
perspective taking), and a controlled battery crossing spatial demand with
visual content shows that it is not explained by task difficulty,
reasoning depth, or sequential update load
(Appendix~\ref{app:construct-validity}). Matched-backbone controls
show that multimodal training can either \emph{introduce} a new
visual-spatial scaffold or \emph{amplify} one already present in the base
LLM. The intervenable effect resides inside the shared LLM hidden state,
not in an isolated visual front-end.

\noindent\textbf{(2) In dual-path models, the causal carrier concentrates in the generation pathway.}
In BAGEL and Janus-Pro, the visual-generation path
concentrates spatial structure into a compact one-dimensional causal
carrier whose removal selectively disrupts spatial-imagery reasoning.
The understanding path also contains spatial information, but in a
probe-decodable form that is entangled with general capacity: neither
its 1D contrast direction nor any tested rank-$k$ subspace yields a
spatially selective handle. In unified models, generation therefore writes a more directly
intervenable spatial geometry into the shared language backbone.

\noindent\textbf{(3) What a model encodes is not always what it uses.}
Behavior, probing, and intervention need not agree. We introduce three
diagnostics, concentration, alignment, and semantic location, that explain
why a readable feature can be causally weak, why a low-concentration
direction can still be behaviorally important, and why a modality axis can
fail while a task-contrastive axis succeeds. These diagnostics turn
probe-intervention mismatches into evidence about how multimodal training
organizes internal representations, so claims about whether a feature is
\emph{used} require all three.

\section{Terminology, scope, and related work}
\label{sec:related}

\noindent\textbf{Terminology and scope.}
We define the three terms in ``emergent vision-shaped thinking'' once,
and use them in this fixed sense throughout.
\emph{Emergent} denotes causal dependence that appears, or becomes
substantially stronger, after multimodal training: matched-backbone
controls show the dependence is largely \emph{introduced} in Chameleon
(the same axis that drops \texttt{navigate} by $27.2$ pp leaves
LLaMA-7B essentially unchanged at $-2.0$ pp) and \emph{amplified and
sharpened} from a scaffold already present in the matched Qwen text
backbones (DD $=+18.5$ to ${\approx}+34$ pp on text-only backbones;
\S\ref{sec:attribution}).
It does not mean de novo creation of task ability.
\emph{Vision-shaped} denotes representations introduced or strengthened by
vision--language training and causally recruited during text-only
reasoning, without implying an explicitly image-like internal scene;
``visual thinking'' would overstate the format evidence, while ``spatial''
would be too narrow for the 32B result below.
\emph{Thinking} refers to the internal computation used during text-only
reasoning, as measured by interventions on hidden states.
\emph{Functional range:} the measured effect is spatially concentrated
within the visual domain at 7--8B and extends to non-spatial visual
attribute binding in the tested 32B configuration, while the core
causal-recruitment result holds at both scales
(Appendix~\ref{app:construct-validity}).

\noindent\textbf{Multimodal training and text-only reasoning.}
Prior work suggests that multimodal training affects more than visual
perception. Spatial reasoning remains brittle for both LLMs and VLMs, and
visual input does not automatically help~\citep{wang2024picture}.
Multimodal neurons and cross-modal structure have been observed in
CLIP-style models and text transformers connected to visual
encoders~\citep{goh2021multimodal,schwettmann2023multimodal}, and visual
training can alter binding strategies and improve text-only
generalization~\citep{buzeta2026seeing}. Related work also studies
vision--language compatibility and language degradation after multimodal
tuning~\citep{merullo2023linearly,huh2024platonic,luo2025vlmtask,
gan2025textualsteering,lu2025trainingfree}. These studies show that
multimodal training can change representations and behavior, but they do
not test whether visual-spatial structure is causally recruited during
image-absent text-only reasoning. Unlike visual-thought or latent-transfer
methods that add, generate, or inject visual/CoT-like
intermediates~\citep{thinkingwithimages2025,wu2024vot,gu2025thinkmorph,
qin2025covt,zhan2025l2vcot}, EVT removes an internal VL direction during
text-only inference.

\noindent\textbf{Mechanistic interpretability of multimodal models.}
Mechanistic interpretability separates decodability from causal
use~\citep{belinkov2022probing}, motivating causal mediation, activation
patching, representation editing, and steering
vectors~\citep{vig2020investigating,geiger2021causal,zou2023repe,
li2023iti}. Prior causal VLM analyses largely focus on image-present
settings, including causal tracing, VQA information flow, visual-token
processing, hallucination mitigation, and cross-modal
corruption~\citep{palit2023blipcausal,basu2024mmllm,neo2025towards,
liu2025fcct,notice2024}. Concurrent work studies image-present spatial
mechanisms through linear spatial/temporal IDs~\citep{kang2026linear} and
interchange interventions that separate vision-encoder from LM-backbone
mechanisms~\citep{cui2026dual}. EVT instead studies the image-absent
setting, where a model may still recruit visual-spatial geometry written
into the language backbone. This matters because VLMs can rely on
linguistic shortcuts even with images~\citep{vlmsneedwords2026}, and
because introduced-versus-amplified scaffolds cannot be inferred from
image-present behavior alone. Methodologically, we extend linear concept
directions, DiffMean steering, and rank-one editing to architecture-matched
VL directions~\citep{bolukbasi2016manprogrammer,markstegmark2023truth,
rimsky2024caa,arditi2024refusal,park2024linear}, following Othello-GPT in
treating latent states as both decodable and causal~\citep{li2023emergent,
nanda2023linear}.

\noindent\textbf{Unified multimodal models.}
EVT also connects to unified models such as Janus-Pro and BAGEL, where
understanding and generation share a language backbone but use different
routes~\citep{chen2025januspro,deng2025bagel}. Prior work asks whether
generation behaviorally improves understanding~\citep{su2026unimrg,
han2026unicorn,chen2025unihetero}, or whether the two capabilities merely
coexist~\citep{yan2025uae,shi2025realunify,niu2025unisandbox,
xie2025mmeunify}. EVT reframes this debate mechanistically. Rather than
asking only whether scores improve, we ask whether understanding and
generation write different reusable geometries into the same backbone.

\section{Method}
\label{sec:method}

Our method has three parts. First, we define a vision--language (VL)
direction that separates textual and visual representations. Second, we
remove each hidden state's projection onto this direction during text-only
inference and measure the performance change. Third, we evaluate the
effect with three primary causal checks, namely task selectivity, direction
specificity, and layer localization, and two supporting analyses,
cross-task gradedness and bidirectional control.

\subsection{Identifying Vision--Language directions}
\label{sec:axis}

We define a vision--language (VL) direction in the representation space
where textual and visual signals are comparable. Let
$\mu_{\text{text}}$ and $\mu_{\text{visual}}$ be the mean textual and
visual representations in that space. For hidden dimension $d$, we define
the VL direction as
$v=(\mu_{\text{text}}-\mu_{\text{visual}})/
\lVert\mu_{\text{text}}-\mu_{\text{visual}}\rVert_2
\in\mathbb{R}^d$. We use a standard difference-in-means estimator and refer to the resulting architecture-matched direction as the VL DiffMean axis. The key design
choice is where to compute this contrast. For each model, we use the
representation space in which textual and visual signals yield an effective
modality contrast. For \textbf{Chameleon-7B}, we average text-token and
visual-token rows in its shared embedding table to obtain a single global VL
direction. Although \textbf{Emu3} also has a shared discrete vocabulary, its
embedding-centroid direction was ineffective in our pilot experiments; for
the reported Emu3 results, we instead compute a separate VL DiffMean direction
at each layer from hidden states at text positions and synthetic visual-token
positions in Emu3-format multimodal sequences. For
\textbf{continuous-encoder models} (Qwen3-VL), visual features enter the
language backbone as continuous hidden states, so we compute per-layer VL
directions from hidden states at text and visual positions in multimodal
sequences. For \textbf{dual-path unified models} (BAGEL, Janus-Pro),
visual understanding and visual generation use different visual token
streams, so we compute separate VL directions for the two paths.

The construction is therefore heterogeneous by design, following one
shared principle: compute the text--visual contrast at the representation
site where the two signals are directly comparable within each
architecture, using the image source that each encoder can actually
process. Every DiffMean axis has split-half stability $\geq 0.9992$, and
swapping synthetic visual inputs for real COCO images leaves the axes
essentially unchanged and reproduces every hub dissociation
(Appendix~\ref{app:source-swap}). These checks support qualitative
cross-architecture replication, not calibrated effect-size ranking. The
canonical VL axis is causal in nine configurations; on Chameleon-30B,
only the input-faithful full-sequence construction recovers a causal
modality axis (Appendix~\ref{app:cham-fullseq};
\S\ref{sec:representation-properties}). Construction details and
stability checks are in Appendix~\ref{app:repro-axis}.

\subsection{Projection intervention}
\label{sec:intervention}

Given hidden state $h$ and normalized axis $v$, we intervene by $h \leftarrow h - \alpha\,(h^{\!\top} v)\,v$.
Here $\alpha{=}1$ removes the projection, $\alpha\in(0,1)$ partially removes it, and $\alpha<0$ multiplicatively amplifies it.
The hook acts on the post-block residual stream of each selected decoder
block (Appendix~\ref{app:repro-hw}). For $\alpha<0$, the update scales each
token's existing projection on $v$, whereas additive steering
$h\leftarrow h+\beta v$ adds the same offset to all tokens; we therefore use
multiplicative amplification for amplification controls.

\noindent\textbf{Dosage selection and held-out validation.}
The strength $\alpha$ is selected once per configuration and then frozen.
Six of ten configurations use $\alpha{=}1.0$ (full projection removal) and
require no dosage selection; the remaining positive configurations use a
coarse sweep that picks the smallest $\alpha$ yielding a stable selective
effect without non-selective collapse. To rule out sample overfitting, we
repeated the sweep under a pre-fixed 125/125 calibration/evaluation split:
the selected dosages remain effective on held-out samples, with every
reported 95\% CI excluding 0, and the frozen settings transfer unchanged
to the 27-task, external-benchmark, and mental-simulation evaluations.
Positive effects persist over broad dosage ranges in the swept
configurations; Liquid is the sole cliff-like exception
(Appendix~\ref{app:fail-liquid-cliff}). Full dosage choices, sweeps, and
the held-out protocol are in Appendices~\ref{app:alpha-dosage}
and~\ref{app:heldout-alpha}.

All evaluation inputs are text-only. Thus, the intervention does not remove visual tokens from the input; it tests whether text-only hidden states \emph{use} a VL direction introduced or amplified by multimodal training.

For higher-rank analyses, we replace $vv^{\!\top}$ with $BB^{\!\top}$:
$h \leftarrow h - \alpha\,B B^{\!\top}h$.
For each multimodal example $i$, we compute
$\delta_i=\mu_{\text{text},i}-\mu_{\text{visual},i}$, where the two
centroids are averaged over text and visual positions, respectively.
Here $B\in\mathbb{R}^{d\times k}$ contains the top-$k$ left singular
vectors of the matrix formed by $\{\delta_i\}$.
This extension is used only for the dual-path geometry analysis in
\S\ref{sec:gen-vs-und}.

\subsection{Evaluation protocol}
\label{sec:eval}

We evaluate on two families of text-only tasks.
\textbf{Spatial-imagery} tasks naturally construct an internal spatial scene,
whereas \textbf{symbol-manipulation} controls rely on formal symbolic operations. Tasks are drawn from
BBH~\citep{suzgun2022challenging} and external spatial-reasoning datasets,
including bAbI, SpartQA, and
StepGame~\citep{weston2015babi,mirzaee2021spartqa,shi2022stepgame}
(\S\ref{sec:setup}, Appendix~\ref{app:repro-data}).

For each task, we measure
$\Delta_{\text{task}}=\text{Acc}_{\text{intervened}}-
\text{Acc}_{\text{baseline}}$ and summarize selectivity by
$\text{DD}=\Delta_{\text{control}}-\Delta_{\text{spatial}}$, where
$\text{DD}>0$ means larger spatial degradation. For random-control
summaries, we also report
$z=(\mathrm{DD}_{\mathrm{axis}}-\mu_{\mathrm{rand}})/\sigma_{\mathrm{rand}}$,
where $\mu_{\mathrm{rand}}$ and $\sigma_{\mathrm{rand}}$ are computed from
the matched random-direction DD distribution. We use $z$ as
direction-specificity evidence rather than as a cross-model effect size
(Appendix~\ref{app:repro-stats}).
EVT is identified by three causal checks.
\textbf{Task selectivity}
requires larger degradation on spatial-imagery tasks than on
symbol-manipulation controls. \textbf{Direction specificity} compares the
VL axis against norm-matched random axes, with energy-matched controls in
representative cases to match removed projection energy. \textbf{Layer
localization} requires the effect to peak in a small block range. We also
test \textbf{cross-task gradedness}, where degradation should increase
with spatial-imagery demand, and \textbf{bidirectional control}, where
amplification or reinjection should reduce the deficit. Each task uses 250 samples;
decoding details and dosage sweeps are in Appendices~\ref{app:repro-hw}
and~\ref{app:alpha-dosage}.

\section{Multimodal training reshapes reasoning geometry}
\label{sec:experiment-main}
\label{sec:evt}

\subsection{Cross-Architecture double dissociation}
\label{sec:cross-arch-dd}
\label{sec:setup}
We evaluate 10 multimodal configurations. Table~\ref{tab:main-effect}
summarizes the 9 configurations where the canonical VL axis is causal;
Chameleon-30B is analyzed separately (\S\ref{sec:representation-properties};
Appendices~\ref{app:fail-cham30b} and~\ref{app:cham-fullseq}).
Across these 9 VL-axis-positive configurations, VL-axis intervention
yields a double dissociation, with spatial-imagery tasks degrading while
symbol-manipulation controls are preserved or less affected. We identify the \emph{causal hub}
by block-wise sweeps followed by single-layer scans. Hyperparameters,
splits, and axis details are in Appendix~\ref{app:setup}.

\begin{table}[t]
\centering
\small
\setlength{\tabcolsep}{4.2pt}
\renewcommand{\arraystretch}{1.05}
\caption{\textbf{Cross-architecture double dissociation under VL-axis intervention.}
For each model, we report the peak causal hub, intervention strength,
DD$=\Delta_{\text{control}}-\Delta_{\text{spatial}}$ when available,
and direction-specificity control. DD is in percentage points; DD $> 0$
indicates stronger degradation on spatial-imagery tasks than on
symbol-manipulation controls. $z$ reports random-null direction specificity.}
\label{tab:main-effect}
  \begin{tabularx}{\linewidth}{@{}l c c c c c c Y@{}}
  \toprule
  \textbf{Model} & \textbf{Fam.} & \textbf{Axis} & \textbf{Hub}
  & $\boldsymbol{\alpha}$ & \textbf{DD} & $\boldsymbol{z}$ & \textbf{Specificity} \\
  \midrule
  Cham-7B
  & VQ & Emb. & L5 & 1.0
  & $+33.2$ pp
  & $+12.22$
  & $0/101$ rand.;\newline CI $[+24.8,+45.4]$ pp \\
  Emu3
  & VQ & Diff. & L3+F & .50/.50
  & $+20.8$ pp
  & --
  & $0/22$ early/final rand.;\newline CI $[+9.2,+27.0]$ pp \\

  Liquid
  & FT & Diff. & L27 & 1.0
  & $+21.6$ pp
  & --
  & $0/22$ rand.;\newline CI $[+12.0,+31.2]$ pp \\

  BAGEL
  & Dual & Gen-Diff. & L21--27 & .5
  & $+16.8$ pp
  & $+3.19$
  & $0/22$ rand.;\newline CI $[+10.7,+23.0]$ pp \\

 Janus
& Dual & Gen-Diff. (VQ) & L22--28 & .7
& $+8.0$ pp
& $+3.58$
& $0/22$ rand.;\newline CI $[+2.93,+13.00]$ pp \\

  Qwen3-2B
  & ViT & Diff. & L3 & 1.0
  & --
  & $+27.7$
  & $0/22$ rand.; CI n.e. \\

  Qwen3-8B
  & ViT & Diff. & L8 & 1.0
  & --
  & $+38.3$
  & $0/100$ rand.; CI n.e. \\

  Qwen3-32B
  & ViT & Diff. & L8 & 1.0
  & --
  & $+262.5$
  & $0/22$ rand.;\newline CI $[+92.0,+118.0]$ pp \\

  Qwen3-A3B
  & MoE & Diff. & L6--18 & 1.0
  & --
  & $+48.8$
  & $0/5$ rand.; CI n.e. \\

  \bottomrule
  \end{tabularx}
\vspace{2pt}
\begin{minipage}{0.98\linewidth}
\footnotesize
\raggedright
\emph{Notes.} ``0/$N$ rand.'' means no random direction matched the real-axis
effect. Abbrev.: Emb./Diff./Gen-Diff. = embedding-centroid/per-layer/
generation-path DiffMean; the VQ tag denotes VQ-token generation states.
F = final block; CI n.e. = not estimated. Dashes mean no paired DD or
continuous $z$ is reported. Qwen3-A3B uses a
smaller random-null budget; Qwen3-32B's large $z$ reflects small null
variance and is used as pass/fail evidence, not cross-model effect size.
Chameleon-7B: DD is the single-layer L5 peak (\texttt{navigate} $-27.6$,
\texttt{boolean} $+5.6$); $z$ and the CI use the matched L0--7 block
(block DD $+32.0$). For the Qwen configurations, $z$ is the
random-null direction-specificity statistic on the
\texttt{colored\_objects}/\texttt{web\_of\_lies} pair (at 8B,
\texttt{navigate} itself is nearly unaffected, so no
\texttt{navigate}-based DD or CI is reported); for Chameleon/BAGEL/Janus,
$z$ is the paired DD significance statistic, with direction specificity
carried by the 0/$N$ random-control column.
\end{minipage}
\end{table}

\begin{wrapfigure}{r}{0.56\textwidth}
    \centering
    \vspace{-6pt}
    \includegraphics[width=\linewidth]{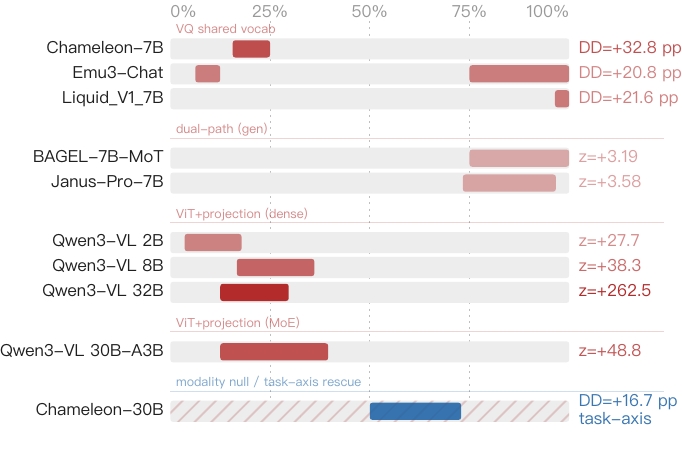}
    \caption{Causal hubs across model architectures.}
    \label{fig:causal-hubs}
    \vspace{-7pt}
\end{wrapfigure}

\noindent\textbf{Hub locations reflect modality integration points.}
Causal hubs are architecture-dependent and align with where visual
information enters the language backbone (Figure~\ref{fig:causal-hubs}).
Hubs appear early for shared-vocabulary or projection-based models such as Chameleon
(L5) and Qwen3-VL (L3--L8), but later for post-hoc visual fine-tuning and
dual-path generation models such as Liquid (L27), BAGEL, and Janus-Pro
(L21--L28). This argues against a generic perturbation account. The hub is
tied to each architecture's vision--language interface. Emu3 further shows
two carrier windows, early (L0--L7) and late (L24--L31), with highly
correlated per-task drops ($r=0.97$), suggesting redundant spatial copies
(Appendix~\ref{app:emu3-dual-carrier}). In BAGEL and Janus-Pro,
selectivity appears mainly in generation-path layers; the
understanding-path VL axis overlaps with dominant hidden-state components,
so removal causes broad rather than selective damage
(\S\ref{sec:gen-vs-und}).

\subsection{Specificity and Cross-Benchmark robustness}
\label{sec:four-checks}
We next ask whether the effect is spatial-specific rather than generic
damage or a benchmark artifact.

\noindent\textbf{Specificity beyond Table~\ref{tab:main-effect}.}
Energy-matched controls test a stricter alternative. The VL axis may look
special simply because it removes more residual-stream energy. This
alternative fails. On the matched Chameleon-7B L0--L7 block, energy-matched directions yield
DD $=-1.9\pm3.6$ pp, whereas the VL axis yields $+15.2$ pp
($z=+4.8$, one-sided $p<10^{-4}$ under a normal approximation;
Appendix~\ref{app:energy-matched},
Table~\ref{tab:energy-matched-cham7b}). Thus, the effect reflects direction-specific spatial structure rather than
removal magnitude or generic damage.

\begin{figure}[H]
    \centering
    \vspace{-6pt}
    \captionsetup[subfigure]{font=footnotesize,labelfont=bf,skip=1pt}
    \begin{subfigure}[t]{0.55\textwidth}
        \centering
        \includegraphics[width=\textwidth]{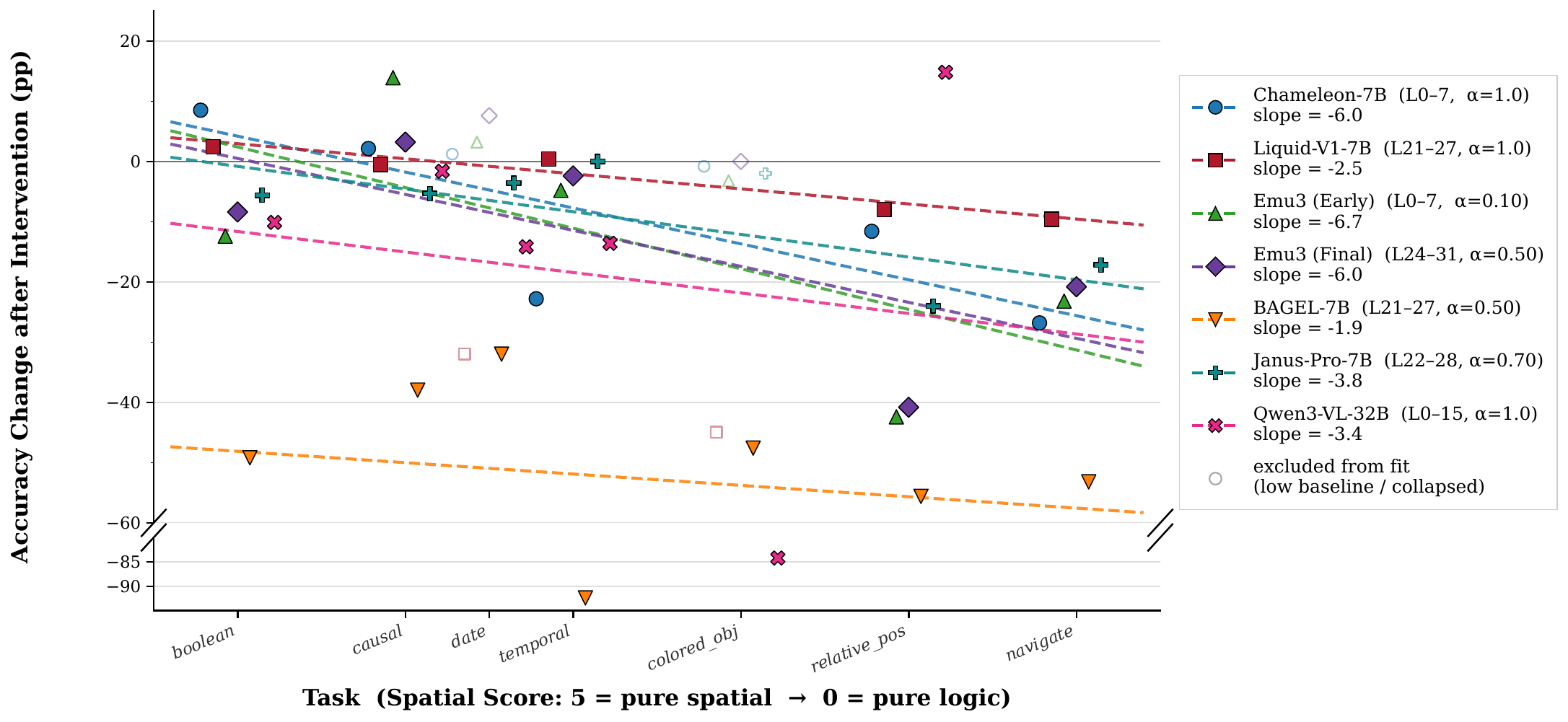}
        \caption{Spatial task spectrum.}
        \label{fig:seven_task_spectrum}
    \end{subfigure}
    \hfill
    \begin{subfigure}[t]{0.41\textwidth}
        \centering
        \includegraphics[width=\textwidth]{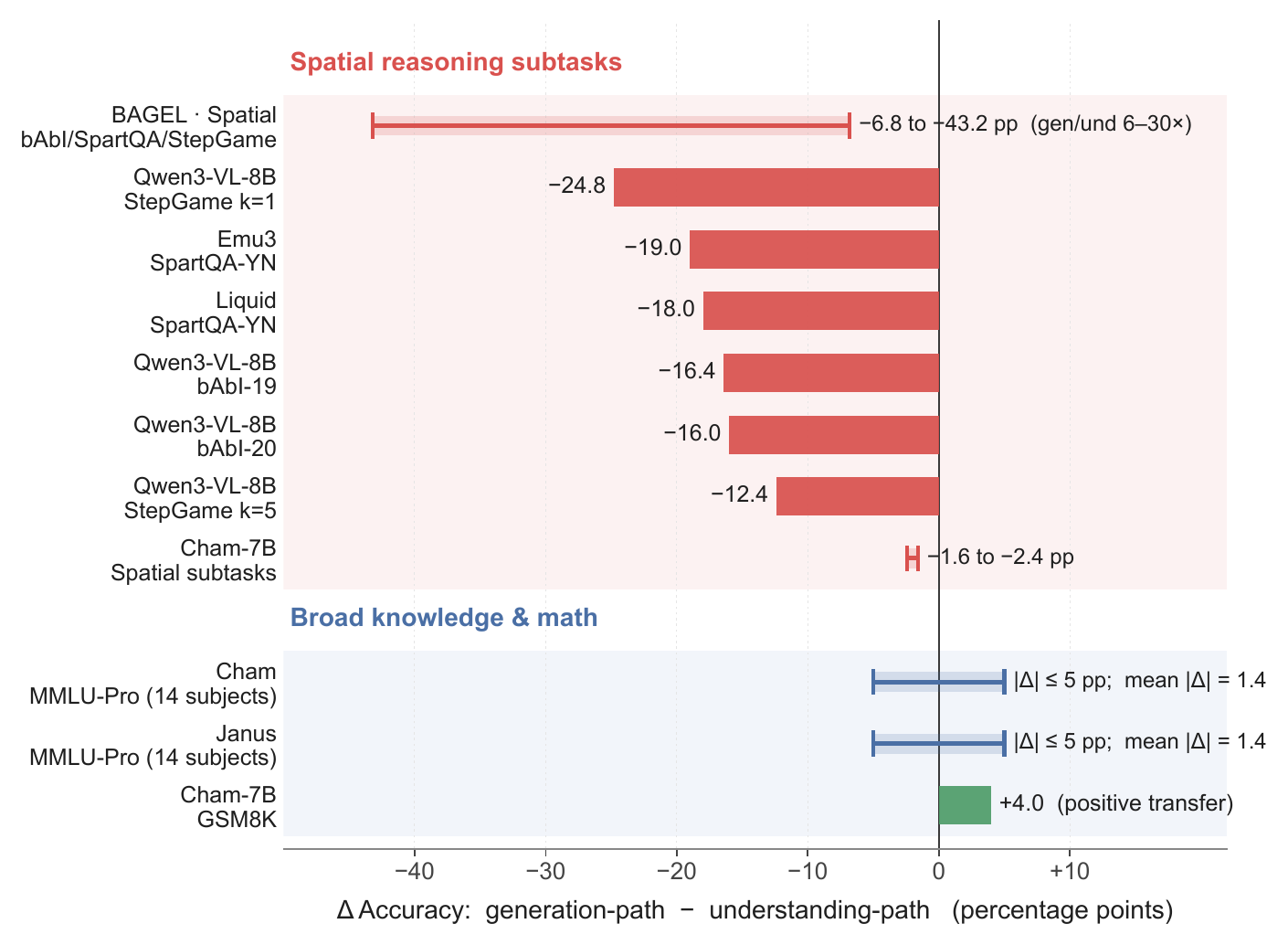}
        \caption{Cross-benchmark transfer.}
        \label{fig:cross_benchmark}
    \end{subfigure}
    \vspace{-2pt}
    \caption{\textbf{Task selectivity beyond the main benchmark.}
(a) VL-axis intervention is increasingly harmful from logic-like to spatial
tasks; points with collapsed baselines are excluded from the regression fits.
(b) Effects transfer to external spatial benchmarks, while knowledge/math
benchmarks stay nearly unchanged.}
    \label{fig:panel-spectrum}
    \vspace{-6pt}
\end{figure}

\noindent\textbf{Cross-task gradedness and external benchmarks.}
Figure~\ref{fig:panel-spectrum}a reports per-task accuracy changes on a seven-task spectrum ordered by a $0$--$5$ spatial-imagery score annotated by four frontier LLMs and sanity-checked by two AI-background humans.
Inter-annotator agreement is high (Kendall's $W\!\approx\!0.98$): six of
seven tasks receive unanimous LLM ratings, and the human checks show the
same pattern (Appendix~\ref{app:spatial-score}). Across configurations, the intervention shows predominantly negative
slopes, indicating that degradation increases with spatial demand rather
than uniformly across tasks. The negative slope is not an artifact of baseline
difficulty: after controlling for baseline accuracy across the seven
tasks, the spatial-score slope remains negative in 6 of 7 configurations
($-1.6$ to $-7.3$ pp per score point; Chameleon-7B Spearman
$\rho=-0.83$, $p=0.042$; Appendix~\ref{app:construct-validity}).
Qwen3-VL-8B is the exception: it passes direction specificity strongly
($z=+38.3$) but shows a positive slope, because its axis also damages
low-score control tasks under this protocol
(Appendix~\ref{app:spectrum-control}).

Figure~\ref{fig:panel-spectrum}b shows that this selectivity transfers
beyond the main benchmark. Drops concentrate on spatial reasoning
benchmarks such as bAbI, SpartQA, and StepGame, whereas broad knowledge
and math benchmarks such as MMLU-Pro remain nearly unchanged. This pattern
argues against generic format disruption and supports a spatial-specific
causal effect. One boundary case qualifies this transfer: BAGEL's
generation-path intervention at its main dosage also substantially
damages GSM8K ($-53.6$ pp) and LogiQA ($-25.2$ pp), so knowledge/math
preservation holds for the configurations shown in
Figure~\ref{fig:panel-spectrum}b but is not universal across models; we
treat the BAGEL external-benchmark cells as a non-selective dosage
boundary rather than as transfer evidence.

\subsection{Attribution and functional use}
\label{sec:attribution-control}
\label{sec:attribution}

We next test where the causal direction comes from and whether the removed signal
is functionally \emph{used}, using matched-backbone controls, hub-specific
ablation, same-direction rescue, and multiplicative amplification
(Appendices~\ref{app:fail-transfer} and~\ref{app:rescue-layer}--\ref{app:rescue-dosage}).

\begin{figure}[t]
    \centering
    \captionsetup[subfigure]{font=footnotesize,labelfont=bf,skip=2pt}
    \begin{subfigure}[b]{0.5\textwidth}
        \centering
        \includegraphics[width=\textwidth]{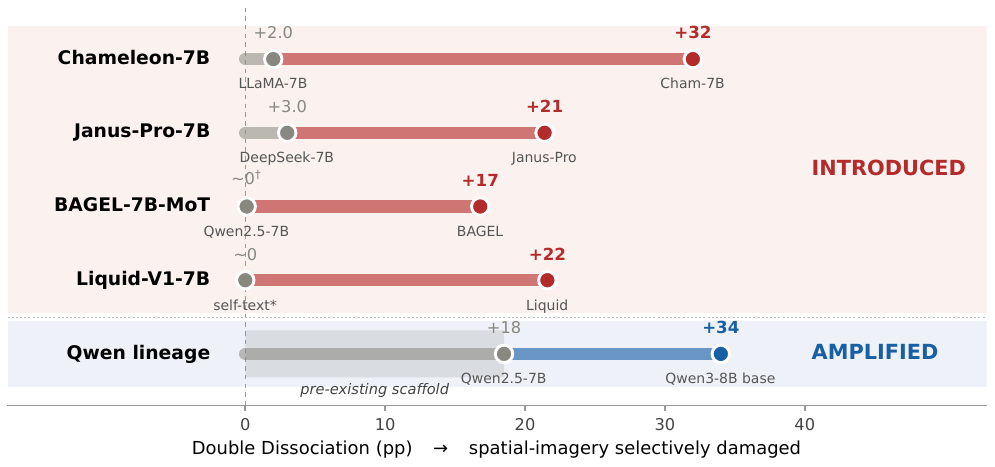}
        \caption{Cross-architecture attribution.}
        \label{fig:panel-attribution}
    \end{subfigure}
    \hfill
    \begin{subfigure}[b]{0.48\textwidth}
        \centering
        \includegraphics[width=\textwidth]{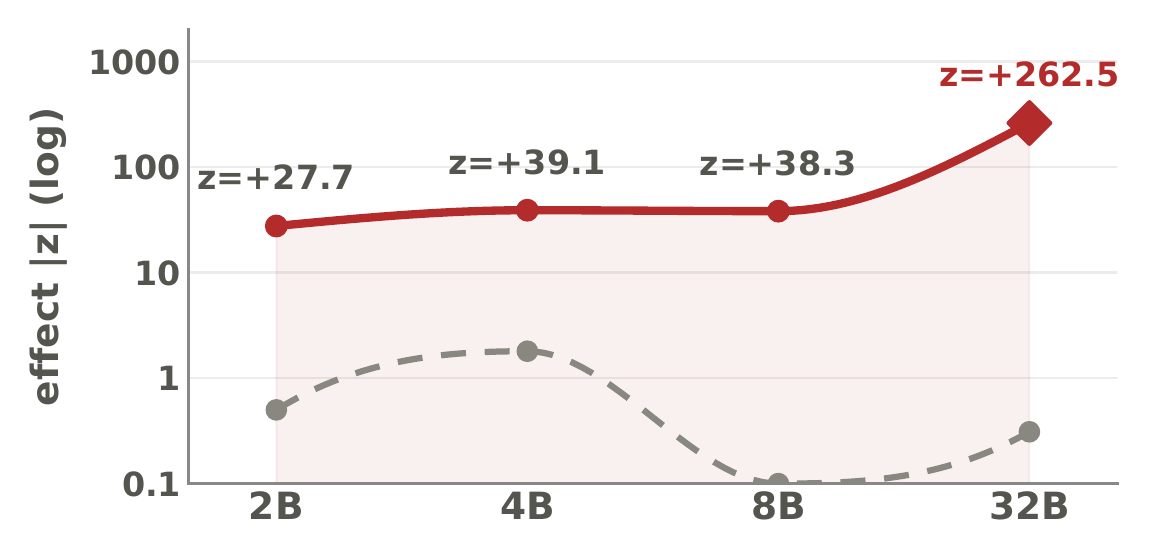}
        \caption{Qwen3-VL scaling.}
        \label{fig:panel-qwen-scaling}
    \end{subfigure}
    \caption{\textbf{Attribution: introduction vs.\ amplification.}
(a) Matched-base transfer collapses for four architectures; only Qwen carries
a pre-existing scaffold (+18.5 pp on Qwen2.5-7B). (b) In Qwen3-VL,
spatial-effect $|z|$ is already large at 2B--8B and becomes strongest at 32B, while control-task $|z|$ stays near zero, indicating scale-strengthened selectivity.}
\end{figure}

\noindent\textbf{Attribution: introduction vs.\ amplification.}
Multimodal training can either introduce a new causal scaffold or amplify one that already exists in the text-only backbone. For \emph{Chameleon}, the same axis leaves LLaMA-7B essentially
unchanged (\texttt{navigate} $-2.0$ pp; every task within $\pm3.3$ pp,
comparable to random directions), versus a $-27.2$ pp \texttt{navigate}
drop and DD $=+32.0$ pp in Chameleon-7B, indicating an introduced effect
(Figure~\ref{fig:panel-attribution}).
For the \emph{Qwen} family, the matched-backbone control reveals an existing
scaffold: a Qwen2.5-VL axis yields DD $=+18.5$ pp on the text-only
Qwen2.5-7B backbone, and a Qwen3-VL-8B axis transferred to the Qwen3-8B
base backbone yields DD $\approx+34$ pp---but with a $\sim$5 pp
non-selective damage floor on all tasks, whereas in Qwen3-VL-8B itself
the same intervention is cleaner (DD $=+30.0$ pp with controls
unaffected). Multimodal training thus amplifies and \emph{purifies} a
scaffold already present in the base LLM
(Figure~\ref{fig:panel-attribution}).\footnote{Qwen3 pretraining-data composition is not fully public, so introduction-vs.-amplification cannot be cleanly distinguished for the Qwen3-VL$\to$Qwen3-8B-base
transfer in Figure~\ref{fig:panel-attribution}. Qwen2.5-7B, documented as
text-only pretrained, provides the cleaner matched-family anchor.}
Within Qwen3-VL, direction-specificity $z$ clusters between $+27.7$ and
$+39.1$ across 2B--8B and becomes strongest at 32B ($+262.5$). The
intermediate 4B run is included only as a scaling sanity check. The affected
layer window expands from $\approx 4$ to $\approx 11$ layers
(Figure~\ref{fig:panel-qwen-scaling}), supporting amplification.

This amplification trend is family-dependent rather than a universal
scaling law: in Chameleon, scaling relocates the carrier deeper (L5 at 7B
$\to$ mid/final blocks at 30B; \S\ref{sec:representation-properties}).
Cross-family transfer further shows that these carriers are
model-specific rather than universal hidden-space directions
(Appendix~\ref{app:fail-transfer}, Table~\ref{tab:transfer}).

\noindent\textbf{Functional use: rescue and steering controls.}\label{sec:rescue}
We ask whether the VL direction is functionally used rather than merely
diagnostic. Because the L5 ablation causes a layer-specific impairment,
we next ask whether the removed information can be recovered.
Canonical-axis ablation at the Cham-7B L5 hub gives the headline
\texttt{navigate} drop of $-27.6$ pp. In a paired rescue run, L6
reinjection with $\beta=0.5\sigma$ recovers $+8.0$ pp, while the matched
non-hub L20 ablation is null ($+0.8$ pp) and the L20--L21 reinjection
control shows no analogous recovery. Rescue is therefore hub-specific
rather than a generic residual-stream effect
(Appendices~\ref{app:rescue-layer}--\ref{app:rescue-direction};
Tables~\ref{tab:rescue-layer}--\ref{tab:rescue-dosage}).

Additive shifts provide a negative steering control. The update $h\leftarrow h+\beta v$
does not yield a reliable positive rescue regime
(Appendix~\ref{app:additive-vs-multiplicative},
Table~\ref{tab:additive-vs-multiplicative}). Thus, EVT is carried by a
token-wise projection pattern, not by a prompt-independent steering bias.

Finally, BAGEL provides a cross-path rescue test. After removing the
compact generation carrier, we ask whether the distributed
understanding-path subspace can compensate. We analyze this in
\S\ref{sec:gen-vs-und}.

\section{The generation path forms a compact spatial causal carrier}
\label{sec:generation-carrier}

Section~\ref{sec:evt} shows that EVT appears across architectures and is
introduced or amplified by multimodal training. We now locate the carrier
in dual-path models. BAGEL-7B-MoT and Janus-Pro-7B provide an internal
comparison, because both have a continuous ViT/SigLIP understanding path and a
VAE/VQ generation path sharing the same LLM backbone.

\noindent\textbf{One-dimensional carriers are generation-dominant.}\label{sec:cross-task-path}
We construct a path-specific VL axis for each path and intervene over the
late LLM blocks used by each model, L21--27 for BAGEL and L22--28 for Janus.
Across both models, the one-dimensional generation-path axis yields the
stable spatially selective effect, while the understanding-path axis is
weak, null, or dominated by broad input-statistics differences. In
BAGEL, the generation axis gives a spatial-control double dissociation of
$+16.8$ pp ($z=+3.19$), whereas the understanding axis gives $-10.9$ pp
(n.s., slightly reversed). The same pattern is not confined to the BBH-style tasks. On
bAbI, SpartQA, and single-hop StepGame, the generation-to-understanding damage ratio
ranges from $6{\times}$ to $31{\times}$ (Appendix~\ref{app:cross_task_path}).
On a 27-task suite, Janus-Pro independently reproduces the qualitative effect.
\texttt{navigate} drops by $-28.7$ pp, whereas
\texttt{boolean\_expressions} drops by only $-7.3$ pp
(Appendix~\ref{app:janus-bbh27}, Table~\ref{tab:janus-bbh27}).

This 1D failure does not mean the understanding path lacks spatial
information: in Janus-Pro, its DiffMean axis is dominated by a tenfold
text/visual norm gap (Appendix~\ref{app:fail-janus-early}), so a null 1D
intervention cannot distinguish absent from distributed information,
motivating rank-$k$ PCA tests.

\begin{wrapfigure}[18]{r}{0.60\textwidth}
    \centering
    \vspace{-8pt}
    \includegraphics[width=\linewidth]{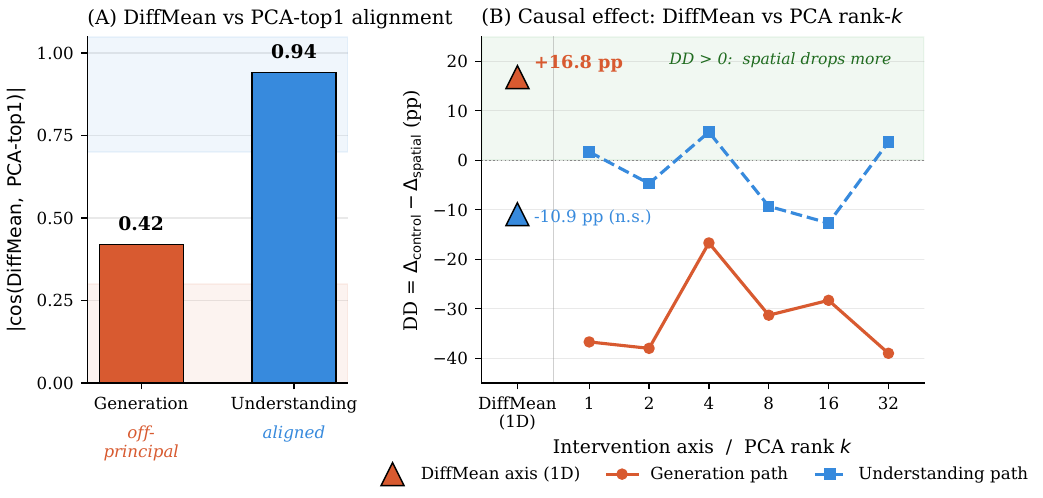}
    \caption{\textbf{BAGEL dual-path DiffMean vs.\ PCA geometry.}
(A) Hub-block DiffMean is off-principal in generation ($|\cos|\approx0.42$)
but top-PC-aligned in understanding ($0.94$). (B) Positive DD = stronger
spatial degradation. Only the 1D generation-path carrier is selective
($+16.8$ pp); the understanding axis is null ($-10.9$ pp), and rank-$k$
PCA ablation is non-selective in both paths (all DD $\leq 0$).}
    \label{fig:large_img}
    \vspace{-8pt}
\end{wrapfigure}

\noindent\textbf{DiffMean and PCA reveal complementary geometries.}
\label{sec:gen-vs-und}
DiffMean and PCA ask different questions. DiffMean tests whether a single
vision--text contrast direction forms a clean causal carrier; PCA tests
whether related information is distributed over a higher-rank difference
subspace. The two can therefore disagree. A 1D contrast can be directly
intervenable, while a high-rank subspace can contain stronger but more
entangled effects. Figure~\ref{fig:large_img} visualizes this
DiffMean--PCA distinction.

For PCA, we center per-sample vision/text mean-difference vectors, compute
the top-$k$ basis $\mathbf{B}$, and apply
$h \leftarrow h - \alpha \mathbf{B}\mathbf{B}^{\top}h$,
$k \in \{1,2,4,8,16,32\}$. In BAGEL's hub block, DiffMean sits off the
dominant variance direction in generation (mean $|\cos|=0.42$ over
L21--26; $0.19$ averaged over all layers) but is aligned with it in
understanding ($|\cos|>0.85$ at every hub layer;
\S\ref{sec:representation-properties};
Appendix~\ref{app:anchors}, Table~\ref{tab:anchor-summary}). Together, they expose complementary path geometries, summarized in
Table~\ref{tab:dual_path_geometry}.

\begin{table}[t]
\centering
\footnotesize
\setlength{\tabcolsep}{3.5pt}
\renewcommand{\arraystretch}{1.0}
\begin{tabularx}{\linewidth}{@{}l c c c >{\raggedright\arraybackslash}X@{}}
\toprule
\textbf{Path}
& \textbf{DiffMean (1D, $\alpha{=}0.5$)}
& \textbf{Top-1 EV}
& \textbf{Top-16 EV}
& \textbf{PCA rank-$k$ ablation} \\
& \textbf{DD (pp)}
&
&
& \textbf{(selectivity)} \\
\midrule
Generation
& $+16.8$ ($z{=}+3.19$)
& 19\%
& 69.5\%
& non-selective at all $k$; controls drop 56--74 pp \\
Understanding
& $-10.9$ (n.s., $z{=}-1.84$)
& 53\%
& 86.2\%
& non-selective at all $k$; $k{=}16$: controls $-32.3$ vs.\ spatial $-19.7$ pp \\
\bottomrule
\end{tabularx}
\caption{
\textbf{Two distinct geometries in BAGEL dual paths.}
DD $> 0$ denotes larger spatial than control degradation. DiffMean gives a
clean rank-1 carrier in the generation path and is null (slightly
reversed) in the understanding path under the same block-level path
intervention. PCA rank-$k$ ablation is significant but non-selective in
both paths: at every tested $k$, control tasks degrade at least as much
as spatial tasks, indicating entanglement with general capacity rather
than a clean carrier (Appendix~\ref{app:anchors}).
}
\label{tab:dual_path_geometry}
\end{table}

Table~\ref{tab:dual_path_geometry} summarizes the asymmetry. In
generation, DiffMean isolates a compact 1D carrier, while rank-$k$ PCA
ablation damages controls more than spatial tasks at every $k$. In
understanding, neither the 1D DiffMean direction nor any tested rank-$k$
subspace yields a spatially selective effect; at $k{=}16$ removal is
clearly damaging, but controls drop more than spatial tasks
($-32.3$ vs.\ $-19.7$ pp;
Appendix~\ref{app:anchors}, Table~\ref{tab:anchor-summary}).
Thus, generation exposes a compact causal handle, whereas the
understanding path's spatial information---though present---is not
selectively intervenable at any tested rank.

\noindent\textbf{Generation paths compress; understanding-path information stays entangled.}
These results support a geometric division of labor. Generation training
appears to compress spatial structure into a compact 1D DiffMean signal
that can drive visual token generation. The understanding path preserves
spatial information in decodable form (\S\ref{app:within-task-probes}),
but neither DiffMean nor PCA isolates a spatial-selective carrier there.

Cross-path rescue is consistent with this reading. After
generation-carrier removal, amplifying the rank-16 understanding PCA
subspace recovers spatial accuracy ($+6.92$ pp), but also controls
($+5.07$ pp; Appendix~\ref{app:cross-path-rescue},
Table~\ref{tab:cross-path-rescue}). The recovery is real but not
spatially selective: the understanding path carries recoverable
task-relevant information in an entangled, non-selective form.

Encoder differences alone are unlikely to explain this, since both paths inject into the same late LLM backbone with the same hidden dimension, and the DiffMean–PCA contrast appears inside the shared backbone rather than only at injection. We interpret this as path-level division of labor, with generation supplying a compact selective carrier and understanding supplying a distributed backup. Controlled training ablations are needed to test whether generation alone causes this compression.

\section{From decodability to use: three geometric diagnostics}
\label{sec:representation-properties}

Sections~\ref{sec:evt} and~\ref{sec:generation-carrier} show that
decodability and causal use need not coincide. In Chameleon-7B, the
VL-axis probe peaks at the same layer where VL-axis ablation is most
effective, whereas in BAGEL's understanding path, the probe-visible
VL-axis direction has little selective causal effect
(Appendix~\ref{app:probe-vs-hub}). We therefore use three geometric
diagnostics to explain when a readable direction becomes a functional
carrier.

First, \emph{concentration} measures how much hidden-state energy lies
along an axis:
$c_L(\hat a)=\mathbb{E}_{i}[(h_{i,L}^{\top}\hat a)^2/\|h_{i,L}\|^2]$.
Second, \emph{alignment} measures whether a readable direction matches
the causal direction:
$A(\hat a)=|\cos(\hat a_{\mathrm{readout}},\hat u^{\mathrm{causal}}_{1})|$.
Third, \emph{semantic location} asks which candidate contrast exposes the
functional carrier. Operationally, we scan candidate axes or contrasts
$a\in\mathcal{C}$ across layers, keep settings that pass the
direction-specificity check without causing large control-task damage, and
then select the layer--axis pair with the largest double dissociation.
For $|\mathcal{C}|>1$ or task-defined contrasts, we require held-out
validation, as in Chameleon-30B (Appendix~\ref{app:fail-cham30b}).
This matters when a modality-defined axis is null but a task-contrastive
axis succeeds.

\noindent\textbf{Concentration} ($c$) measures how much of the hidden state
lies along the causal axis. We report both raw concentration and a centered
version that removes the layer mean, because a large raw value can reflect
a global mean offset rather than sample-varying signal. This distinction
matters for BAGEL. Its generation-path DiffMean axis has very large raw
concentration ($47$--$58\%$ across L21--L26), but this is mostly due to the
hidden-state mean aligning with the axis ($|\cos(\mu,a)|=0.71$--$0.79$).
After centering, the concentration drops to $0.43$--$3.16\%$, still
$15$--$118\times$ above random directions (Appendix~\ref{app:anchors}, Table~\ref{tab:concentration}).

Chameleon shows a different geometry. Its L5 VL axis has low centered
concentration ($0.14\%$, $5.3\times$ random), yet its ablation produces a
spatial-selective effect. Across the L3--L7 hub window, centered
concentration ranges from $0.04$--$0.38\%$, while the non-hub L20 control
falls to the random floor after centering ($0.4\times$ random) (Appendix~\ref{app:anchors}, Table~\ref{tab:concentration}). Thus,
concentration separates two mechanisms. Chameleon uses a low-concentration
but selective causal signal, whereas BAGEL uses a more concentrated
generation-path signal that is mixed with broader task information.

\noindent\textbf{Alignment} ($A$) asks whether the direction that makes a
feature linearly readable is also the direction the model causally uses.
BAGEL's understanding path shows a clear dissociation. A 1D probe on the
DiffMean axis reaches 0.73 accuracy ($+2.65\sigma$ above random), yet
removing the same direction is null ($\mathrm{DD}=-7.3$ pp;
Appendix~\ref{app:probe-vs-hub}, Table~\ref{tab:bagel-und-probe}), and no
tested rank-$k$ subspace is selective either (\S\ref{sec:gen-vs-und}):
readability and causal alignment dissociate.

We summarize this relation using
$|\cos(\mathrm{DiffMean},\mathrm{PCA\text{-}top\text{-}1})|$, which yields
three regimes. In the \emph{concentric} regime (BAGEL understanding,
hub-layer $|\cos|>0.85$), DiffMean lies inside the dominant variance
subspace, but 1D intervention is null. In the \emph{projected-diffuse}
regime (BAGEL generation, hub mean $|\cos|\approx0.42$, all-layer mean
$0.19$), DiffMean is separated from the dominant
variance direction yet isolates the spatial-selective carrier. In the
\emph{mixed} regime (Qwen3-VL), the relation changes across layers,
producing non-monotonic rank-$k$ curves that never reach spatial
selectivity---the per-layer 1D DiffMean axis is the only selective
handle (Appendix~\ref{app:anchors}, Table~\ref{tab:anchor-summary}).

\noindent\textbf{Semantic location} ($\ell^{*}$) asks which contrast
reveals the causal carrier. A null result on one contrast does not mean
the carrier is absent; it may simply be organized along another semantic
axis --- or the contrast may have been \emph{estimated} in the wrong
space. Chameleon-30B illustrates both. Its modality axis, estimated from
the embedding table or synthetic token-ID sequences, is non-causal across
19 block--dosage cells, whereas a \texttt{navigate}--\texttt{boolean}
task-contrastive axis is causal (held-out DD $=+6.28$ pp, CI
$[+0.13,+12.37]$; Appendix~\ref{app:fail-cham30b}). Re-estimating the
modality axis input-faithfully, with images through Chameleon's own VQ
tokenizer in full multimodal sequences, yields a rotated axis
($\cos \approx 0.4$--$0.6$) that is strongly causal at mid/final blocks
(mid DD $=+33.2$ pp, \texttt{navigate} $z=-88.7$;
Appendix~\ref{app:cham-fullseq}). The carrier is thus not lost at 30B; it
relocates deeper and is exposed by a task contrast or a correctly
estimated modality contrast.

Chameleon-7B further shows that the two contrasts are distinct. The
modality and task axes are nearly orthogonal ($|\cos|<0.04$), yet both can
be causal at L5 (Appendix~\ref{app:fail-cham30b},
Table~\ref{tab:cham-axis-cos}). Semantic location therefore depends on capacity,
architecture, and layer, not simply on model capability.

\noindent\textbf{Diagnostic implication.}
Together, these cases show why decodability alone is insufficient.
Mechanistic claims should pair causal interventions with concentration,
alignment, and semantic location. Extended anchor cases are in
Appendix~\ref{app:anchors}.

\section{Discussion and limitations}\label{sec:limitations}

Our results suggest that multimodal training reshapes text-only reasoning
inside the language backbone, not merely via a visual front-end, in an
architecture- and pathway-dependent form.

Our study is limited to linear or low-rank interventions and the model
families we could access. Our tasks (BBH, bAbI, SpartQA, StepGame, a mental-simulation battery,
and a controlled visual$\times$spatial battery;
Appendix~\ref{app:construct-validity}) are diagnostic rather than a
complete taxonomy. Two boundaries qualify this
scope. First, mental-simulation evidence is measurable only in
configurations that clear a capability floor: Chameleon-7B is at or below
chance on every mental-simulation task, so that evidence comes from the
Qwen3-VL configurations. Second, sufficiently long non-spatial state
tracking (8--12 sequential updates) shows partial overlap with the same
carrier, so the carrier is spatially concentrated rather than exclusively
spatial. The analysis may miss nonlinear,
highly distributed, or cross-attention-specific mechanisms, and Qwen attribution
remains constrained by public uncertainty about pretraining data, although
matched-backbone controls reduce this ambiguity. Moreover, axis
constructions that bypass the model's real multimodal input pathway can
miss true carriers (Chameleon-30B); our conclusions rest on constructions
validated against real-image inputs wherever the architecture permits.

Finally, our interventions test necessity rather than the full
computation: they do not show whether the model constructs an image-like
scene, nor which circuits write and read this signal.

\noindent\textbf{Broader impacts.}
This work is diagnostic: it introduces no new model, dataset, or
generation capability, and improves transparency about when multimodal
models rely on visual-spatial representations in text-only reasoning.
Similar intervention tools could be misused to manipulate model behavior;
we release only analysis code and benchmark scripts.

\section{Conclusion}

EVT provides causal evidence that multimodal training can make
vision-shaped representations usable during text-only reasoning. Across
architectures, removing a VL direction selectively harms spatial text tasks
while sparing logical controls, with the carrier taking the form of a
compact axis, a broader subspace, or an amplified pre-existing geometry.
Multimodal training thus changes which internal representations text-only
inference causally routes through, not just what models can process.

\begin{ack}
This research is supported by the National Research Foundation,
Singapore, under grant NRF-NRFI10-2024-0004; by the DSO research grant
awarded by DSO National Laboratories, Singapore; and by the Ministry of Education,
Singapore, under its Tier-1 Academic Research Fund (No.~24-SIS-SMU-040).
The authors thank the anonymous reviewers and the area chair for
suggestions that led to the mental-simulation battery, the
visual$\times$spatial control battery, and the within-task probe
analyses added in this version.
\end{ack}

{
\small
\bibliographystyle{plainnat}
\bibliography{refs}
}

\clearpage
\appendix

\section*{Supplementary Material}

\appendix

\section{Reproducibility}
\label{app:setup}
\label{app:reproducibility}

This section consolidates the experimental setup, compute budget,
axis-construction provenance, and random-seed accounting for all main-text
claims, in the format expected by the NeurIPS reproducibility checklist.
We release the intervention code, per-sample correctness JSONs used
for bootstrap CI recomputation, and the scripts needed to reproduce the
main experiments (\S\ref{app:repro-release}).

\subsection{Hardware and software}
\label{app:repro-hw}

All experiments use \textbf{8$\times$ data-center GPUs with $\approx$140\,GB HBM each} (H200-class), with PyTorch in \texttt{bfloat16} throughout. A 7B model in \texttt{bfloat16} occupies $\approx$14--16\,GB, so most experiments use 1--2 GPUs per model instance. Three lightweight components (axis construction, position-steering pilots) ran on a single 48\,GB or 80\,GB GPU. Decoding is greedy; \texttt{max\_new\_tokens}${=}128$ for chain-of-thought formats and $16$ for direct-answer formats. Bootstrap confidence intervals use $n_{\text{boot}}{=}10000$ paired-by-index resamples.

\subsection{Compute budget}
\label{app:repro-compute}

Table~\ref{tab:compute-budget} reports per-experiment GPU-hours. The total of \textbf{$\approx$150 GPU-hours} is a conservative estimate; Emu3, Liquid, and BAGEL main runs lack itemized wall-clock logs and are absorbed into the 7-task spectrum line. Janus random-control numbers are reported at $n{=}100$ per task rather than $n{=}250$ because the per-example latency on Janus ($\approx$7--8\,s) made $22 \times 5 \times 250 \approx 61$ GPU-hours infeasible; we re-computed real-axis DD on the matched $n{=}100$ subset for the random-control comparison and retained $n{=}250$ as the headline DD.

\begin{table}[h]
\centering
\footnotesize
\setlength{\tabcolsep}{3.5pt}
\renewcommand{\arraystretch}{1.0}
\caption{\textbf{Compute budget.} Per-experiment GPU-hours and configurations. Hardware column denotes the GPU count used; \texttt{$\Sigma$} denotes total inference forward passes (samples $\times$ tasks $\times$ seeds/dirs). Per-experiment configurations are in the released codebase under \texttt{experiments/<model>/}.}
\label{tab:compute-budget}
\begin{tabularx}{\linewidth}{@{} l l r r r r X @{}}
\toprule
\textbf{Experiment} & \textbf{Model} & \textbf{$n$} & \textbf{tasks} & \textbf{seeds/dirs} & \textbf{GPU-h} & \textbf{Hardware} \\
\midrule
Cham-7B core (Exp1) & Cham-7B & 250 & 2 & $1{+}20$ & ${\sim}22$ & 3$\times$GPU \\
Cham-7B 60-dir random control & Cham-7B & 250 & 2 & 60 & ${\sim}50$ & 8$\times$GPU \\
Cham-7B axis-config sweep & Cham-7B & 250 & 2 & 3 & ${\sim}2.5$ & GPU \\
Cham-30B Exp1 + Exp2A & Cham-30B & 250 & 2 & $1{+}20$ & ${\sim}14$ & 4$\times$GPU \\
Cham-30B Exp2B (LOO axes) & Cham-30B & 250 & 2 & 12 pairs & ${\sim}6$ & 2$\times$GPU \\
Janus random control & Janus & 100 & 5 & 22 & ${\sim}24$ & 3$\times$GPU \\
Janus main effect & Janus & 250 & 5 & 1 & ${\sim}2.5$ & GPU \\
EXP4 phase 0 (axis construction) & 3 models & 50/64 & --- & --- & $<0.5$ & GPU \\
EXP4 phase 1 ($k$-sweep, 6 values) & 3 models & 100 & 6 & 6 & ${\sim}6$ & GPU \\
EXP4 dose-response & BAGEL und & 100 & 6 & 3 & ${\sim}0.2$ & GPU \\
7-task spectrum (7 configs) & 6 models & 250 & 7 & 1 & ${\sim}18$ & GPU \\
Position steering (additive/multiplicative) & Cham-7B & 80--250 & 7 & 5 & ${\sim}3.5$ & GPU \\
Cross-family transfer (V.1--V.4) & 4 backbones & 250 & 5 & --- & ${\sim}4$ & GPU \\
\midrule
\textbf{Total (conservative estimate)} & \multicolumn{4}{l}{} & $\mathbf{\sim 150}$ & \\
\bottomrule
\end{tabularx}
\end{table}

\subsection{Axis-construction provenance}
\label{app:repro-axis}

The 10 model configurations use three axis-construction methods (embedding centroid, per-layer DiffMean, per-layer PCA on per-sample DiffMean vectors). Sample-source heterogeneity is the most important cross-experiment caveat: BAGEL gen-path and Janus axes use real COCO val2017 images, while all other axes use synthetic random-pixel images plus text prompts. We document this asymmetry explicitly; in pilot experiments we compared synthetic-vs-real source on BAGEL und-path and found split-half cosine $\geq 0.99$ across both sources, so the choice does not affect axis identity at our $N{=}50$ sample size. Appendix~\ref{app:source-swap} extends this check to all synthetic-axis models: rebuilding each axis from real COCO images reproduces both the axis direction at the causal hub and the hub intervention effect.

Shared vocabulary does not by itself determine the construction space. Chameleon uses the embedding-centroid axis reported in the main experiments, whereas Emu3 uses per-layer hidden-state DiffMean axes. We tested the analogous global embedding-centroid axis for Emu3 as an initial pilot, but it produced no meaningful intervention effect (all changes within approximately $2$ percentage points; all $p\geq0.5$), so it is not the axis reported for Emu3 in the main results.

\begin{table}[h]
\centering
\footnotesize
\setlength{\tabcolsep}{3pt}
\renewcommand{\arraystretch}{1.0}
\caption{\textbf{Axis construction provenance per model.} Method, sample source, sample count $N$, base random seed, and code reference.}
\label{tab:axis-provenance}
\begin{tabularx}{\linewidth}{@{} l l X r l @{}}
\toprule
\textbf{Model} & \textbf{Method} & \textbf{Sample source} & \textbf{$N$} & \textbf{Seed scheme} \\
\midrule
Chameleon-7B/30B (main) & Embedding centroid & Embedding rows only (no runtime samples) & --- & --- \\
Cham-7B (task-contrastive axis) & Per-layer DiffMean & Synthetic random-pixel images & 50 & \texttt{4096+i} for dir $i$ \\
Emu3-Chat & Per-layer DiffMean & Synthetic visual-token IDs & 50 & not seeded (fixed prompt loop) \\
BAGEL gen (main) & Per-layer DiffMean & \textbf{Real COCO val2017 images} & 50 & \texttt{4000+i} \\
BAGEL und (main) & Per-layer DiffMean & Synthetic random images & 50 & \texttt{1000+i} \\
BAGEL gen v2 (cross-check) & Per-layer DiffMean & Synthetic random images & 50 & \texttt{2000+i} \\
EXP4 (3 anchor models, phase 0) & Per-layer PCA on $\delta_i$ & Synthetic random pixel images & 50/64 & split1: \texttt{1000+i}, split2: \texttt{2000+i} \\
Janus-Pro-7B (gen) & Per-layer DiffMean & \textbf{Real COCO val2017 images} & 50 & not explicitly recorded \\
Liquid-V1-7B & Per-layer DiffMean & Synthetic sequences (Emu3-style) & 50 & not recorded \\
Qwen3-VL-8B (main) & Per-layer DiffMean & Synthetic random pixel images & 64 & \texttt{1000+i} \\
\bottomrule
\end{tabularx}
\end{table}

\paragraph{COCO usage.} The 50 images used for BAGEL gen-path and Janus axes are sampled from \textbf{COCO val2017} (CC-BY-4.0). To respect the license without distributing image bytes, the code release ships a \texttt{download\_coco\_50.sh} script with a hard-coded list of image IDs.

\paragraph{Split-half stability.} For all DiffMean-based axes, we doubled $N$ to $100$ and randomly split into two halves; per-layer cosine similarity between the two halves is $\geq 0.9992$ across all models and all layers. This justifies $N{=}50$ as our default budget. The high split-half stability justifies using $N{=}50$ as our default
axis-construction budget.

\subsection{Real-image source-swap validation of synthetic axes}
\label{app:source-swap}

A natural objection to synthetic axis inputs is that random pixels (or
uniformly random visual-token IDs) are out-of-distribution, so the
resulting direction might be an artifact of abnormal visual inputs rather
than a modality direction that real images would also produce. To test
this, we rebuilt the VL DiffMean axis for every synthetic-axis model with
the visual source swapped to $50$ real COCO val2017 images (the same
canonical image set used for the BAGEL gen-path and Janus axes), passed
through each model's own visual pathway: the ViT+projector for Qwen3-VL,
the Emu3-VisionTokenizer VQ encoder for Emu3, and Liquid's release
pipeline (pad-to-square, $512{\times}512$, Chameleon VQGAN, code
$+256000$ vocabulary offset) for Liquid. Text prompts, sequence
templates, position masks, axis estimator, hook mechanics, and the
evaluation harness are unchanged; only the visual source varies. We then
(i) compare the real-image axis $v_{\text{real}}$ to the synthetic axis
$v_{\text{syn}}$ per layer --- this direction comparison covers
\emph{every} configuration for which both axes exist, including BAGEL's
understanding path, Janus-Pro's generation path, Chameleon, and all five
Qwen3-VL sizes --- and (ii) re-run the causal-hub intervention with
$v_{\text{real}}$ on the identical task samples for the four
synthetic-axis models with paired references, including a fresh
$5$-direction random-null control.

\begin{table}[h]
\centering
\footnotesize
\setlength{\tabcolsep}{4pt}
\renewcommand{\arraystretch}{1.05}
\caption{\textbf{Axis identity under the source swap:} per-layer cosine
between the synthetic-input axis and the real-image (COCO) axis,
summarized at each model's causal hub and over the full stack. Low
minima occur only at layers where the axis itself is weakly defined
(centroid separation $1$--$2$ orders of magnitude below hub layers);
COCO-axis split-half stability is $\geq 0.96$ for all models at all
layers, so the reported cosines are not noise-limited.
$^{a}$Janus's canonical axis is already built from COCO; the swapped
source is random-RGB images through its VQ tokenizer (same comparison,
reversed direction). $^{b}$Chameleon rows compare the full-sequence
per-layer DiffMean axes of Appendix~\ref{app:cham-fullseq}; Chameleon's
main-result axis is the embedding-centroid direction, which uses no
input images at all.}
\label{tab:source-swap-cos}
\begin{tabularx}{\linewidth}{@{}l l c c l@{}}
\toprule
\textbf{Model (path)} & \textbf{Hub layers} & $\boldsymbol{\cos}$ \textbf{at hub} & \textbf{Median (all layers)} & \textbf{Min (all layers)} \\
\midrule
Qwen3-VL-2B      & L3      & $0.999$ & $0.997$ & $0.49$ (L0) \\
Qwen3-VL-4B      & L6--8   & $0.998$ & $0.998$ & $0.38$ (L0) \\
Qwen3-VL-8B      & L8      & $0.999$ & $0.995$ & $0.29$ (L0) \\
Qwen3-VL-32B     & L8      & $0.995$ & $0.995$ & $0.49$ (L0) \\
Qwen3-VL-30B-A3B & L6--18  & $0.983$--$0.989$ & $0.987$ & $0.79$ (L0) \\
Emu3-Chat        & L0--7   & $0.956$--$0.995$ & $0.955$ & $0.61$ (L31) \\
Liquid-V1-7B     & L27     & $0.951$ & $0.641$ & $0.23$ (L0) \\
BAGEL-7B-MoT (und) & L21--27 & $0.941$--$0.997$ & $0.998$ & $0.66$ (L1) \\
Janus-Pro-7B (gen)$^{a}$ & L22--28 & $0.849$--$0.932$ & $0.963$ & $0.81$ (L29) \\
\addlinespace
Chameleon-7B$^{b}$  & L0--7   & $0.73$--$0.79$ & $0.766$ & $0.67$ \\
Chameleon-30B$^{b}$ & L36--47 & $0.90$--$0.96$ & $0.861$ & $0.54$ \\
\bottomrule
\end{tabularx}
\end{table}

\textbf{Axis identity.} Table~\ref{tab:source-swap-cos} reports the
per-layer cosine between the two axes for every configuration with both
sources. At its causal hub, every positive-result model shows
$\cos \geq 0.94$ (Qwen3-VL family $\geq 0.98$); BAGEL's understanding
path, for which the original pilot reported only an aggregate
$\geq 0.99$, shows hub-window cosines of $0.941$--$0.997$ with a
full-stack median of $0.998$. Low cosines appear only where the axis is
weakly defined: Qwen3-VL layers L0--5 have centroid separation
$19$--$37$ versus $786+$ from L6 onward, and there the two sources
disagree ($\cos 0.29$--$0.60$) --- a property of an underdetermined
direction, not of the source. COCO-axis split-half stability is
$\geq 0.98$ for the four re-run models ($\geq 0.96$ for Chameleon), so
these cosines are limited by source agreement, not sampling noise. For
Liquid we additionally verified that a $256$-px encoding (matching the
$256$-token count of the synthetic sequences) agrees with the $512$-px
axis at $\cos = 0.986$, so token count is not a confound. The Chameleon
rows show the weakest direction agreement ($0.73$--$0.79$ at the 7B
hub); notably, even there both sources' axes produce a causal
spatial-selective effect of the same sign
(Table~\ref{tab:cham-fullseq}), and Chameleon's main-result axis is in
any case the embedding-centroid direction, computed from embedding rows
without any input images, so the synthetic-input objection does not
apply to it.

\begin{table}[h]
\centering
\footnotesize
\setlength{\tabcolsep}{4pt}
\renewcommand{\arraystretch}{1.05}
\caption{\textbf{Real-image source swap: the same intervention with the
synthetic axis and with the real-image (COCO) axis.} Each model has two
rows --- one per axis source --- evaluated on identical task samples. If
the synthetic axis were an artifact of out-of-distribution inputs, the
real-image rows would show no effect; instead, within every model the
two rows show the same pattern: the spatial task is damaged and the
control task is untouched. Accuracies in \%, base $\to$ intervened;
DD as in Table~\ref{tab:main-effect} (pp, DD${}>0$ = spatial-selective
damage). Spatial/control tasks: \texttt{colored\_objects}/%
\texttt{web\_of\_lies} for Qwen3-VL, \texttt{navigate}/\texttt{boolean}
for Emu3 and Liquid. $\cos(v_{\text{syn}},v_{\text{real}})$ is measured
at the hub layers. $^{a}$The matched synthetic 32B run did not evaluate
\texttt{web\_of\_lies}; the paper CI is shown instead of a DD point
estimate. $^{b}$Liquid's real-image axis is stronger per unit $\alpha$
(see text), so its dose-matched comparison uses $\alpha{=}0.5$.}
\label{tab:source-swap}
\begin{tabularx}{\linewidth}{@{}l >{\raggedright\arraybackslash}X c c c l@{}}
\toprule
\textbf{Model (hub, $\alpha$)} & \textbf{Axis built from} & $\boldsymbol{\cos(v_{\text{syn}},v_{\text{real}})}$ & \textbf{Spatial acc.} & \textbf{Control acc.} & \textbf{DD [95\% CI]} \\
\midrule
Qwen3-VL-8B  & random pixels & & $81.3 \to 52.7$ & $44.0 \to 45.3$ & $+30.0$ \\
(L0--8, $1.0$) & real COCO   & $0.999$ & $81.3 \to 3.3$ & $44.0 \to 45.3$ & $+79.3$ [$+68.0,+90.7$] \\
\addlinespace
Qwen3-VL-32B & random pixels & & $89.3 \to 4.0$ & ---$^{a}$ & CI $[+92,+118]^{a}$ \\
(L0--15, $1.0$) & real COCO  & $0.995$ & $89.3 \to 7.3$ & $26.0 \to 45.3$ & $+101.3$ [$+87.3,+114.7$] \\
\addlinespace
Emu3-Chat    & random token IDs & & $41.6 \to 0.0$ & $39.6 \to 18.8$ & $+20.8$ \\
(L0--7, $0.5$) & real COCO   & $0.956$--$0.995$ & $41.6 \to 0.0$ & $39.6 \to 11.6$ & $+13.6$ [$+4.0,+23.2$] \\
\addlinespace
Liquid-V1-7B & random token IDs ($\alpha{=}1.0$) & & $54.8 \to 42.0$ & $54.8 \to 63.6$ & $+21.6$ [$+12.0,+31.2$] \\
(L27)$^{b}$  & real COCO ($\alpha{=}0.5$) & $0.951$ & $54.8 \to 42.0$ & $54.8 \to 56.4$ & $+14.4$ [$+5.2,+23.6$] \\
\bottomrule
\end{tabularx}
\end{table}

\textbf{Causal reproduction.} Re-running each hub intervention with
$v_{\text{real}}$ reproduces the spatial-selective dissociation in every
configuration (Table~\ref{tab:source-swap}; all DD CIs exclude zero, and
no random direction matches the effect). Magnitudes match the synthetic
CIs for Qwen3-VL-32B, Emu3, and Liquid; for Qwen3-VL-8B the real-image
axis is \emph{stronger} ($+79.3$ vs.\ $+30.0$ pp), with the control task
still untouched (\texttt{web\_of\_lies} $+1.3$ pp, $z\approx-0.1$ against
the random null). Fine-grained patterns replicate point-for-point:
Emu3's \texttt{navigate} drop under the early-block intervention is
$-41.6$ pp (to $0.0$) under both sources, and Qwen3-VL-32B's paradoxical
\texttt{navigate} \emph{improvement} is $+39.3$ pp under both sources.

\textbf{Two source-sensitive components.} First, Liquid's L27 hub sits
in logit space (tied embeddings; \S\ref{app:fail-liquid-cliff}), and the real-image
axis carries a larger projection onto task hidden states there
($0.60$ vs.\ $0.46$), shifting the documented dose cliff to lower
$\alpha$: at $\alpha{=}0.5$ the COCO axis reproduces the synthetic
$\alpha{=}1.0$ headline effect almost exactly (\texttt{navigate}
$-12.8$ pp, identical to the synthetic drop; \texttt{boolean} unharmed at
$+1.6$ pp; DD $+14.4$), whereas $\alpha{\geq}0.75$ overshoots into
nonselective collapse. Second, Emu3's redundant \emph{final} carrier
(L24--31; Appendix~\ref{app:emu3-dual-carrier}) does not replicate under
the real-image axis (DD $-0.0$, CI $[-9.2,+8.8]$), consistent with the
larger axis divergence in that window ($\cos$ falls to $0.61$ by L31);
Emu3's primary early carrier is source-robust. Neither caveat affects
the headline configurations, which replicate under real images in all
four models. We conclude that the synthetic-input axes are not artifacts
of out-of-distribution visual inputs.

The swap also runs in the reverse direction for the two dual-path
models, whose canonical axes are already built from real COCO images.
For Janus-Pro, a generation-path axis rebuilt from seed-fixed
random-pixel images (through the VQ-16 encoder) agrees with the COCO
axis at $\cos = 0.85$--$0.93$ over the causal hub L22--28, and in an
environment-matched rerun produces the same causal profile
(\texttt{navigate} $-20.8$ pp with controls unchanged, versus $-25.2$ pp
for the COCO axis). For BAGEL's understanding path, the COCO-built axis
matches the noise-built axis at $\cos \geq 0.994$ (L21--26) and
reproduces the same \emph{negative} result under intervention
(control-dominated damage, $z=-1.33$, $p=0.91$): the understanding-path
null is a property of the path, not of the visual source. Likewise, the
Janus understanding-path axis rebuilt from random images retains the
pathological input-source norm gap ($\|\mu_{\text{text}}\|{=}14$ vs.\
$\|\mu_{\text{visual}}\|{=}131$ at L0, projection ratio $0.91$) and
collapses non-selectively already at $\alpha{=}0.1$, matching the
real-image failure in Appendix~\ref{app:fail-janus-early}. Positive and
negative conclusions are therefore both source-robust.

\subsection{Input-faithful full-sequence axes for Chameleon, and the resolution of the Chameleon-30B boundary case}
\label{app:cham-fullseq}

The source-swap protocol above also closes the one construction gap in
the Chameleon family. The canonical Chameleon axis is the embedding-table
centroid difference, and the per-layer DiffMean variants previously
tested on Chameleon (single-token forwards on 7B; $16$
random-token-ID synthetic sequences on 30B, method 7 of
Table~\ref{tab:cham30b-failures}) never passed real images through the
model's own VQ tokenizer inside a full multimodal sequence --- unlike
the Qwen3-VL, Emu3, and BAGEL/Janus constructions. We therefore rebuilt
per-layer VL DiffMean axes for both Chameleon models under the
input-faithful construction used everywhere else: $50$ images (either
the canonical COCO set or seed-fixed random-pixel images), each encoded
by Chameleon's own in-model VQ tokenizer into $1{,}024$ visual tokens
inside a full \texttt{<image>}+text sequence, with text/visual position
masks, estimator, hooks, and evaluation harness unchanged
($\alpha{=}1.0$ unless stated; \texttt{navigate}/\texttt{boolean}, $n{=}250$;
split-half stability $\geq 0.95$).

\begin{table}[h]
\centering
\footnotesize
\setlength{\tabcolsep}{3pt}
\renewcommand{\arraystretch}{1.05}
\caption{\textbf{Full-sequence VL axes on Chameleon-7B/30B: the
modality axis is causal in both models, under both image sources.}
Accuracies in \% (base $\to$ intervened); DD in pp
(DD${}>0$ = spatial-selective). Baselines: 7B
\texttt{navigate}/\texttt{boolean} $41.2/38.8$; 30B $57.6/55.6$.
The 30B late block (L24--35) is a clean null under both sources
(DD $-0.0$, CI $[-6.0,+5.6]$), so the effect is layer-localized, not a
global degradation. Random null: random directions exceeding the real axis; $z$ and $\Delta$ refer to \texttt{navigate} (7B $z$ at $\alpha{=}1.0$).}
\label{tab:cham-fullseq}
\begin{tabularx}{\linewidth}{@{}l l l c c >{\raggedright\arraybackslash}X@{}}
\toprule
\textbf{Model (block, $\alpha$)} & \textbf{Image source} & \textbf{Spatial acc.} & \textbf{Control acc.} & \textbf{DD [95\% CI]} & \textbf{Random null} \\
\midrule
Cham-7B (L0--7, $0.75$) & random pixels & $41.2 \to 28.4$ & $38.8 \to 40.8$ & $+14.8$ [$+4.0,+25.6$] & \\
Cham-7B (L0--7, $0.75$) & real COCO & $41.2 \to 4.8$ & $38.8 \to 39.6$ & $+37.2$ [$+26.8,+48.4$] & $0/5$; $z{=}-12.1$ \\
Cham-7B (L5 only, $1.0$) & real COCO & $41.2 \to 32.4$ & $38.8 \to 46.4$ & $+16.4$ [$+5.6,+27.2$] & \\
\addlinespace
Cham-30B (L12--23, $1.0$) & random pixels & $57.6 \to 47.2$ & $55.6 \to 57.2$ & $+12.0$ [$+4.0,+20.0$] & \\
Cham-30B (L12--23, $1.0$) & real COCO & $57.6 \to 24.0$ & $55.6 \to 55.2$ & $+33.2$ [$+21.2,+44.8$] & $0/3$; $z{=}-88.7$ \\
Cham-30B (L36--47, $1.0$) & random pixels & $57.6 \to 39.2$ & $55.6 \to 59.6$ & $+22.4$ [$+14.0,+30.8$] & $0/3$; all rand.\ $\Delta{=}0$ \\
Cham-30B (L36--47, $1.0$) & real COCO & $57.6 \to 39.2$ & $55.6 \to 59.2$ & $+22.0$ [$+13.6,+30.4$] & $0/3$; all rand.\ $\Delta{=}0$ \\
\bottomrule
\end{tabularx}
\end{table}

\textbf{Chameleon-7B: a third, independent causal direction.} The
full-sequence axis reproduces the canonical double dissociation
(early-block DD monotone in dosage: $+12.0/+33.2/+37.2/+30.8$ pp at
$\alpha{=}0.25/0.5/0.75/1.0$ with COCO images), yet it is nearly
orthogonal to both the embedding-centroid axis (L5 $\cos \approx
0.19$--$0.25$) and the single-token DiffMean axis (L5 $\cos \approx 0$).
Real images strengthen it substantially over random pixels
($+37.2$ vs.\ $+14.8$ pp), mirroring the BAGEL noise-vs-COCO gain.

\textbf{Chameleon-30B: the modality axis is causal after all.} The
full-sequence axis is rotated away from the synthetic-token axis of
method 7 ($\cos \approx 0.43$ against the random-pixel version, $0.56$
against the COCO version, layer mean) and from the embedding axis
($\cos \approx 0.01$), which explains why the original $19$-cell sweep
was null: it probed a different direction. Along the input-faithful
modality axis, Chameleon-30B shows the same spatial-selective,
direction-specific, layer-localized dissociation as every other
configuration --- \texttt{navigate} $-33.6$ pp with \texttt{boolean}
unharmed ($-0.4$ pp) at the mid block --- under both image sources, with
random directions producing exactly zero \texttt{navigate} change at the
final block. The implications for the Chameleon-30B boundary-case
narrative are discussed in Appendix~\ref{app:fail-cham30b}
(resolution paragraph): the spatial carrier at 30B has not decoupled
from the modality contrast; it relocates from the early block (7B, L5
hub) to mid/final blocks and is exposed only when the modality contrast
is measured on the model's actual multimodal input pathway.

\subsection{Intervention dosage choices}
  \label{app:alpha-dosage}

  \paragraph{Hook location.}
  Across Chameleon, Emu3, Liquid, Qwen3-VL, BAGEL, and Janus-Pro, the main
  intervention is implemented with a forward hook on the complete decoder
  block module (e.g., \texttt{model.layers[$\ell$]}), not a pre-hook and not
  a hook on \texttt{self\_attn} or the MLP. The hook modifies the first
  tensor returned by the block, after its MLP residual update. The modified
  tensor is passed to the next block before that block's input normalization;
  for the final decoder block it is passed to the model-level final
  normalization and then the language-model head. BAGEL's implementation
  also exposes an attention-output hook for diagnostic experiments, but all
  BAGEL results reported in the main paper use the default full-block-output
  position.

  The intervention strength $\alpha$ controls how much of the projection
  onto the tested direction is removed:
  $h \leftarrow h-\alpha(h^\top v)v$. We treat $\alpha$ as an intervention
  dosage, not as a learned parameter. For each architecture, we choose a
  dosage from coarse sweeps over layers and $\alpha$ values. The selected
  dosage is the smallest value that produces a stable task-selective effect
  at the causal hub while avoiding non-selective collapse of both spatial
  and control tasks. Once selected, the same dosage is fixed for the
  corresponding main-text analyses and random-direction controls.

  Table~\ref{tab:alpha-dosage} summarizes the dosages used in the main
  experiments. Most models use $\alpha{=}1.0$, corresponding to full
  projection removal. Some architectures require smaller dosages because
  the tested direction overlaps with a high-variance or broadly used
  component of the residual stream. In those cases, large $\alpha$ values
  damage general token prediction rather than selectively affecting
  spatial-imagery reasoning.

 \begin{table}[h]
  \centering
  \footnotesize
  \setlength{\tabcolsep}{4pt}
  \renewcommand{\arraystretch}{1.05}
  \caption{\textbf{Intervention dosages used in the main experiments.}
  $\alpha$ denotes the removal strength in
  $h \leftarrow h-\alpha(h^\top v)v$.}
  \label{tab:alpha-dosage}
  \begin{tabularx}{\linewidth}{@{} l l c >{\raggedright\arraybackslash}X @{}}
  \toprule
  \textbf{Model} & \textbf{Axis / path} & $\boldsymbol{\alpha}$ & \textbf{Selection note} \\
  \midrule
  Chameleon-7B & Embedding VL axis & $1.0$ & Full removal; strongest layer-localized DD at L5. \\
  Emu3 & Early / final VL axes & $0.5$ / $0.5$ & Same dosage for both carriers (L0--7, L24--31); $\alpha{=}1.0$ also degrades controls. \\
  Liquid & DiffMean axis (L27) & $1.0$ & Single layer. The L21--27 block has a sharp cliff: DD $+5.6$ pp at $\alpha{=}0.50$, $+40.0$ pp at $0.55$, non-selective collapse at $1.0$. \\
  BAGEL & Generation-path DiffMean & $0.5$ & $\alpha{=}0.70$ over-suppresses control tasks. \\
  Janus-Pro & Generation-path visual-token & $0.7$ & Most significant in the sweep ($z{=}3.86$, L22--28); significant for $\alpha{\in}[0.30,0.70]$, marginal at $1.0$ ($z{=}1.95$). \\
  Qwen3-VL (all four) & Per-layer DiffMean axis & $1.0$ & Full removal at the hub; 30B-A3B is a preliminary specificity check. \\
  Chameleon-30B & Task-contrastive axis & -- & VL axis null at every tested dosage (Table~\ref{tab:cham30b-failures}). \\
  \bottomrule
  \end{tabularx}
\end{table}
  We separate primary removal dosages from reinjection dosages used in the
  rescue experiments. The rescue intervention uses an additive reinjection
  magnitude $\beta$ measured in units of the target-layer projection
  standard deviation $\sigma$, and is reported separately in
  Appendix~\ref{app:rescue-sigma} and Appendix~\ref{app:rescue-dosage}.

\subsection{Chameleon-7B dose--response curve}
\label{app:cham7b-dose}

To characterize dosage sensitivity in the headline configuration, we ran
an 11-point $\alpha$ sweep on Chameleon-7B (L0--7 block, embedding VL
axis, $n{=}250$ per point per task). Table~\ref{tab:cham7b-dose} reports
absolute accuracy; $\alpha{=}0$ is the no-intervention reference within
this sweep.

\begin{table}[h]
\centering
\footnotesize
\setlength{\tabcolsep}{4.5pt}
\renewcommand{\arraystretch}{1.05}
\caption{\textbf{Chameleon-7B 11-point dose--response.} \texttt{navigate}
degrades smoothly with $\alpha$ and largely saturates beyond
$\alpha\approx0.6$, while \texttt{boolean\_expressions} stays flat or
improves. There is no cliff-like transition; Liquid remains the only
cliff-like case (Appendix~\ref{app:fail-liquid-cliff}).}
\label{tab:cham7b-dose}
\begin{tabular}{lccccccccccc}
\toprule
$\alpha$ & 0.0 & 0.1 & 0.2 & 0.3 & 0.4 & 0.5 & 0.6 & 0.7 & 0.8 & 0.9 & 1.0 \\
\midrule
\texttt{navigate} (\%) & 41.2 & 34.4 & 25.6 & 22.4 & 18.0 & 17.2 & 16.4 & 14.8 & 13.2 & 11.2 & 11.6 \\
\texttt{boolean} (\%) & 38.8 & 42.0 & 48.4 & 46.4 & 47.2 & 45.6 & 48.0 & 48.4 & 48.4 & 48.8 & 50.0 \\
\bottomrule
\end{tabular}
\end{table}

Relative to the $\alpha{=}0$ reference, \texttt{navigate} falls
monotonically to $-24.8$ pp by $\alpha{=}0.6$ and then changes by at most
$5$ pp more through $\alpha{=}1.0$, while \texttt{boolean\_expressions}
improves throughout. The selective effect is therefore present across a
broad dosage range rather than at a tuned operating point.

\subsection{Held-out validation of dosage selection}
\label{app:heldout-alpha}

Because the four swept configurations select $\alpha$ from a coarse sweep
on 250 evaluation examples, the selected dosage could in principle
overfit the sweep sample. To test this, we repeated the original sweep
under a pre-fixed 125/125 calibration/evaluation split: $\alpha$ is
selected on the 125 calibration examples only and then evaluated once on
the 125 disjoint held-out examples.

The dosage selected on the calibration half remains effective on the
disjoint evaluation half: held-out DD is $+20.0$ pp for Emu3 early,
$+19.2$ pp for Emu3 final, and $+9.6$ pp for Janus, and every 95\% CI
excludes 0; Liquid remains the dosage-sensitive exception
(Appendix~\ref{app:fail-liquid-cliff}). The effect is also not confined
to the selected dosage: Emu3 is significant at every swept
$\alpha\in[0.30,1.00]$ and Janus across $[0.30,0.70]$. Three further
checks involve no dosage selection. No external dataset entered $\alpha$
or hub selection, yet the frozen intervention transfers to external
benchmarks (Figure~\ref{fig:cross_benchmark}) and to the 27-task and
mental-simulation evaluations; the Chameleon-30B task-contrastive axis
remains effective on held-out tasks (DD $=+6.28$ pp, CI excludes 0;
Table~\ref{tab:cham30b-heldout}); and zero of $N$ norm-matched random
directions reproduce the effect, while on Chameleon-7B the real axis gives
$z=+4.8$ and exceeds all 20 energy-matched controls
(Table~\ref{tab:energy-matched-cham7b}).

Six of ten configurations use $\alpha{=}1.0$ and involve no dosage
selection at all; Chameleon-30B's seven-dose null sweep was a boundary
check, not a selection. Together with the broad significant dosage
ranges, these checks indicate that the reported effects are not artifacts
of tuning $\alpha$ to a particular sample or task pair.

\subsection{Random seeds}
\label{app:repro-seeds}

Table~\ref{tab:seeds} records all random-seed assignments across experiments. The convention throughout is \texttt{base + i} where \texttt{base} is the experiment's anchor and \texttt{i} indexes the random direction or axis-construction sample.

\begin{table}[h]
\centering
\footnotesize
\setlength{\tabcolsep}{4pt}
\renewcommand{\arraystretch}{1.05}
\caption{\textbf{Random seed assignments per experiment.} \texttt{base+i} convention indexed by random-direction or sample-construction id.}
\label{tab:seeds}
\begin{tabularx}{\linewidth}{@{} l c >{\raggedright\arraybackslash}X @{}}
\toprule
\textbf{Experiment} & \textbf{Base seed} & \textbf{Use} \\
\midrule
Cham-7B/30B random-direction control & 2026 & \texttt{seed = 2026 + i}, $i \in \{0, \ldots, 19\}$ \\
Cham-30B axis build & 2026 & single seed, default \\
Cham-7B task-contrastive random control & 4096 & \texttt{seed = 4096 + i}, $i \in \{0, \ldots, 59\}$ \\
EXP4 phase 0 split 1 & 1000 & \texttt{default\_rng(1000+idx)} per random image \\
EXP4 phase 0 split 2 (stability) & 2000 & \texttt{2000 + idx} per random image \\
BAGEL gen (main) axis & 4000 & \texttt{4000 + idx} per sample \\
BAGEL gen v2 axis & 2000 & \texttt{2000 + idx} per sample \\
BAGEL und axis & 1000 & \texttt{1000 + idx} per sample \\
Janus random control & 0--21 & 22 independent seeds \\
Cross-modal leakage analysis & 42 & \texttt{np.random.seed(42)}, PCA visualization only \\
Cham-30B rescue pilot & 1234 & \texttt{torch.Generator().manual\_seed(1234)} \\
\midrule
Bootstrap CI (all models) & -- & Stochastic in pilot scripts; CIs stable within $\pm 0.1$ pp. Release scripts fix and log the seed. \\
\bottomrule
\end{tabularx}
\end{table}

\paragraph{Bootstrap reproducibility caveat.}
The reported CIs are computed from per-sample correctness files. In the
pilot scripts, bootstrap resampling used stochastic seeds; repeated runs
change CIs by less than $\pm 0.1$ pp because the resampling pool is large.
The public release scripts fix and log the bootstrap seed to enable
bit-for-bit CI recomputation.

\subsection{Datasets}
\label{app:repro-data}

\paragraph{Primary benchmark.} BBH (BIG-Bench Hard), publicly released under Apache-2.0. We use 7 spatial-imagery tasks plus symbolic-manipulation controls; full task list in \S\ref{sec:setup}.

\paragraph{External validation.} SpartQA (yes/no and multi-choice subsets), StepGame (k-hop variants), bAbI tasks 17/19/20, GSM8K, LogiQA, MMLU-Pro. All publicly available; download URLs and version pins in \texttt{data/README.md}.

\paragraph{Axis-construction images.} 50 images from COCO val2017 (CC-BY-4.0) for BAGEL gen-path and Janus; synthetic random-pixel images for all other DiffMean-based axes (fully reproducible from seeds).

\paragraph{Models.} All models loaded via HuggingFace IDs (no weights distributed): \texttt{facebook/chameleon-7b}, \texttt{facebook/chameleon-30b}, \texttt{BAAI/Emu3-Chat}, \texttt{ByteDance-Seed/BAGEL-7B-MoT}, \texttt{deepseek-ai/Janus-Pro-7B}, \texttt{Qwen/Qwen3-VL-\{2B,4B,8B,32B,30B-A3B\}-Instruct}, \texttt{Qwen/Qwen2.5-7B}, \texttt{Qwen/Qwen2.5-VL-7B-Instruct}, \texttt{deepseek-ai/deepseek-llm-7b-base}. The Liquid-V1-7B checkpoint is loaded from its public HuggingFace release (\texttt{Junfeng5/Liquid\_V1\_7B}); no model weights are redistributed by us.

\subsection{Spatial-Imagery Score Annotation}
\label{app:spatial-score}

To test whether the intervention effect varies with spatial demand, we
assign each BBH task a spatial-imagery score from $0$ to $5$. A score of
$5$ denotes a task whose natural solution strongly relies on constructing
or manipulating an internal spatial scene, whereas a score of $0$ denotes
a logic-like task that can be solved primarily by symbolic manipulation.

The scores were independently annotated by four frontier LLMs: GPT,
Gemini, Claude, and DeepSeek. We use the mean LLM rating as the final
spatial-imagery score. To reduce the risk that the ordering is an artifact
of LLM self-annotation, two AI-background human annotators also performed
a blind sanity check. The annotators were highly consistent
(Kendall's $W\!\approx\!0.98$). Table~\ref{tab:spatial-score-annotation}
reports the full annotations.

\begin{table}[t]
\centering
\scriptsize
\setlength{\tabcolsep}{3.2pt}
\renewcommand{\arraystretch}{1.12}
\caption{\textbf{Spatial-imagery score annotation.}
Each task is rated on a $0$--$5$ scale. The final score is the mean of the
four LLM ratings. Human annotations are used as a blind sanity check rather
than for computing the final score.}
\label{tab:spatial-score-annotation}
\begin{tabular}{lcccccccc}
\toprule
Task & GPT & Gemini & Claude & DeepSeek & LLM Mean & Range & Human-A & Human-B \\
\midrule
\texttt{navigate}             & 5 & 5 & 5 & 5 & 5.00 & 0 & 5 & 5 \\
\texttt{relative\_position}   & 3 & 4 & 4 & 3 & 3.50 & 1 & 4 & 3 \\
\texttt{colored\_objects}     & 3 & 3 & 3 & 3 & 3.00 & 0 & 3 & 3 \\
\texttt{temporal\_sequences}  & 2 & 2 & 2 & 2 & 2.00 & 0 & 2 & 2 \\
\texttt{causal\_judgement}    & 1 & 1 & 1 & 1 & 1.00 & 0 & 1 & 1 \\
\texttt{boolean\_expressions} & 0 & 0 & 0 & 0 & 0.00 & 0 & 0 & 0 \\
\texttt{web\_of\_lies}        & 0 & 0 & 0 & 0 & 0.00 & 0 & 0 & 0 \\
\bottomrule
\end{tabular}
\end{table}

Six of the seven tasks receive unanimous ratings from all four LLM
annotators. The only disagreement is on \texttt{relative\_position}:
Gemini and Claude assign a score of $4$, whereas GPT and DeepSeek assign
a score of $3$. This disagreement is consistent with the task's mixed
nature: it uses a spatial left--right relation, but much of the reasoning
can also be implemented symbolically as ordered-list transitivity. The
human sanity check shows the same pattern, with Human-A assigning $4$ and
Human-B assigning $3$.

\paragraph{Mapping to the seven-task spectrum.} The plotted spectrum
(Figure~\ref{fig:seven_task_spectrum}) evaluates
\texttt{temporal\_simple}, a simplified programmatic variant of
\texttt{temporal\_sequences} that inherits its score ($2.0$), and
additionally includes \texttt{date\_understanding} (scored $1.5$ under
the same protocol); \texttt{web\_of\_lies} (score $0$) is used as a
symbol-manipulation control task in the selectivity analyses rather than
as a spectrum point.

We use the spatial-imagery score only as a coarse proxy for spatial demand,
not as ground truth about cognition. Therefore, our prediction is not that
every individual task must follow a strictly monotonic damage pattern.
Instead, tasks with higher spatial-imagery scores should, on average,
exhibit larger degradation under VL-axis ablation.

\subsection{Code release}
\label{app:repro-release}

\paragraph{Released layout.} The released repository is organised by responsibility rather than by experiment site. \texttt{src/core/} contains the shared math: DiffMean / embedding-centroid axis construction (\texttt{axis.py}), projection-removal, additive/multiplicative steering, rescue, and rank-$k$ subspace hooks (\texttt{hooks.py}), both BBH evaluation protocols (\texttt{eval\_bbh.py}), DD/bootstrap/McNemar statistics (\texttt{stats.py}), linear probes, and IO helpers. \texttt{src/models/} provides per-architecture loaders (Chameleon, Qwen3-VL dense and MoE, BAGEL, Emu3, Liquid, Janus). \texttt{src/tasks/} loads BBH and the supplementary task sets, with all dataset paths parameterised through \texttt{configs/paths.yaml} and environment variables. \texttt{src/pipelines/} exposes one runnable module per experiment type (axis construction, intervention, layer scan, random control, $\alpha$-sweep, task spectrum, text-only control, activation addition, rescue, linear probes, rank-$k$ subspace construction and intervention, cross-backbone text ablation, and the BAGEL generation-path axis and intervention); every file carries an \texttt{origin:} header naming the original experiment script it was ported from, so results can be audited line by line. \texttt{experiments/<model>/run.sh} provides one annotated end-to-end launcher per model site, and \texttt{analysis/} ships the figure scripts and the cross-model bootstrap driver. \texttt{tests/smoke\_cpu.py} verifies the intervention math (e.g., that $h-\alpha(h^{\top}v)v$ zeroes the projection at $\alpha{=}1$ and preserves the orthogonal component) without a GPU.

\paragraph{Public release package.}
The public release package contains the following.
\begin{enumerate}
\item \texttt{requirements.txt} pinning all Python dependencies (Python $\geq$ 3.10).
\item Top-level \texttt{README.md} with a step-by-step main-experiment reproduction recipe, and \texttt{reproduce.sh}, a one-command reproduction of the headline Chameleon-7B double dissociation (axis construction $\to$ L0--7 removal $\to$ DD with paired-bootstrap CI), including a \texttt{SMOKE=1} small-$n$ mode.
\item \texttt{configs/paths.yaml} (data and checkpoint roots, overridable via env vars) and \texttt{configs/models.yaml} (HuggingFace IDs and per-architecture layer accessors for every model used).
\item Per-model launchers under \texttt{experiments/<model>/run.sh} that drive the canonical sequence end-to-end on a single machine.
\item \texttt{data/download\_bbh.sh} and \texttt{data/download\_coco\_50.sh} for fetching BIG-Bench-Hard and the 50 COCO val2017 images used to build the BAGEL gen-path and Janus axes (the image bytes themselves are not redistributed), plus the small gradient-task \texttt{jsonl} files shipped in-repo.
\item Fixed seeds logged inline in every result JSON's \texttt{config} field; the released bootstrap scripts fix and log the resampling seed so CIs can be recomputed bit-for-bit from the per-sample correctness files.
\end{enumerate}

\paragraph{Items excluded from release.}
The release does not include model weights (loaded directly from the HuggingFace Hub via the IDs in \texttt{configs/models.yaml}), raw COCO image bytes (replaced by the download script), the SPACE benchmark items (CC-BY-NC-ND; obtainable from the original release~\citep{ramakrishnan2025space}), the upstream BAGEL repository that the BAGEL adapter depends on (pointed at via the \texttt{EVT\_BAGEL\_REPO} environment variable), preliminary pilot runs not used in main claims, and any local session-state artifacts. One component is documented rather than ported: the BAGEL-internal per-sample extraction behind the rank-$k$ phase-0 subspace construction depends on upstream BAGEL internals; the released \texttt{subspace\_construct} pipeline covers the Qwen3-VL path, and the PCA step itself is model-agnostic and included.

\subsection{Statistical testing standards}
\label{app:repro-stats}

For every primary DD claim in the paper, we report:

\begin{itemize}
\item \textbf{Effect size}: $\mathrm{DD} = \Delta_{\text{control}} - \Delta_{\text{spatial}}$ in percentage points, with paired bootstrap $95\%$ CI (paired-by-index, $n_{\text{boot}}{=}10000$).
\item \textbf{Direction specificity vs.\ random}: $z = (\mathrm{DD}_{\text{axis}} - \mu_{\text{rand}}) / \sigma_{\text{rand}}$, with $N_{\text{rand}} \in \{20, 22, 60, 100\}$ independent random unit directions per model. One-sided $p$ is the empirical fraction of random DDs $\geq \mathrm{DD}_{\text{axis}}$.
\item \textbf{Energy-matched random control} (Chameleon-7B and Cham-30B): $20$ random directions where the per-token removal $L_2$ norm is forced equal to the canonical-axis removal norm, isolating direction effects from magnitude effects (energy-matched protocol).
\item \textbf{Held-out validation} (Cham-30B task-contrastive axis): $5$ tasks not used in axis construction; held-out DD with $95\%$ CI to rule out within-axis circularity (Chameleon-30B task-contrastive analysis).
\end{itemize}

We report a model configuration as a positive EVT case only if it passes the four causal checks of \S\ref{sec:eval}; configurations that show a behavioral drop without passing direction specificity or task selectivity are recorded as diagnostic results, not main-claim evidence.

\subsection{Energy-matched random controls}
\label{app:energy-matched}

To rule out the possibility that the VL-axis effect is stronger simply
because it removes more hidden-state energy, we run an energy-matched
random-direction control on Chameleon-7B. For each random direction, we
match the per-token projection-removal norm to that of the canonical
VL-axis intervention. Thus, the control removes the same amount of
hidden-state energy, but along a random orientation.

Table~\ref{tab:energy-matched-cham7b} reports the result for the
Chameleon-7B VL embedding axis at the early block L0--7 with
$\alpha=1.0$. The VL axis remains substantially stronger than the
energy-matched random baseline, confirming that the effect is due to
direction specificity rather than removal magnitude.

\begin{table}[h]
\centering
\footnotesize
\setlength{\tabcolsep}{5pt}
\renewcommand{\arraystretch}{1.05}
\caption{\textbf{Energy-matched random controls on Chameleon-7B.}
Random directions are rescaled so that their per-token removal norm
matches the canonical VL-axis removal norm. The VL axis remains far
outside the energy-matched random distribution.}
\label{tab:energy-matched-cham7b}
\begin{tabular}{l c c c c}
\toprule
\textbf{Direction} & \textbf{DD (pp)} & \textbf{Bootstrap 95\% CI} &
\textbf{Random DD} & \textbf{$z$} \\
\midrule
VL axis, L0--7 & $\mathbf{+15.22}$ & $[+10.0,+20.5]$ &
$-1.88 \pm 3.56$ & $\mathbf{+4.80}$ \\
\bottomrule
\end{tabular}

\vspace{2pt}
\footnotesize
\emph{Note.} The random baseline uses $20$ energy-matched random
directions; none reaches the real-axis effect (empirical $p<0.05$;
one-sided $p<10^{-4}$ under a normal approximation of the control
distribution).
\end{table}

\section{Decodability vs.\ Causal Handle: Two Dissociation Patterns}
\label{app:probe-vs-hub}

Section~\ref{sec:cross-arch-dd} reports that decodability and causal handle are not interchangeable diagnostics. This appendix presents the per-layer probe data behind that claim for two complementary cases: Chameleon-7B (where decodability is uniform and causality is localized) and BAGEL's understanding path (where decodability is localized but $1$D causality is null). The two cases are not redundant: they document the two distinct ways in which probing can mislead the analyst, and together they motivate the alignment property in \S\ref{sec:representation-properties}.

\subsection{Chameleon-7B: uniform decodability, localized causality}
\label{app:probe-cham7b}

\paragraph{Setup.}
We trained two probes per layer on $250$ samples from \texttt{navigate} (spatial) and $250$ from \texttt{boolean\_expressions} (symbolic-control), with $5$-fold stratified cross-validation. The first is a full-dimensional logistic probe on the layer-$L$ hidden state at the last input token ($d{=}4096$). The second is a $1$D probe on the scalar projection $h \cdot \hat{v}$, where $\hat{v}$ is the canonical Chameleon-7B VL embedding-centroid axis used in the main intervention. As a direction-specificity control we trained $22$ matched $1$D probes on random-direction projections (each direction sampled once and L2-normalized; seeds $0$--$21$).

\paragraph{Full probe is saturated everywhere.}
The full-dimensional probe achieves $1.000$ accuracy at every one of the $32$ layers (mean $\pm$ std across folds: $1.000 \pm 0.000$). This reflects vocabulary divergence between the two tasks (\texttt{left}/\texttt{right}/\texttt{forward} vs.\ \texttt{True}/\texttt{False}/\texttt{and}/\texttt{or}) rather than learned spatial structure: any sufficiently expressive probe can separate the tasks at any depth. The full probe therefore provides no localization signal whatsoever.

\paragraph{1D VL-axis probe is informative at the hub.}
Table~\ref{tab:cham7b-probe} summarizes the 1D-axis probe across layer regions. The probe is informative through the early-to-mid layers (L0--L15, where the causal hub L5 sits), with peak accuracy of $1.000$ at L5 itself.

\begin{table}[h]
\centering
\footnotesize
\setlength{\tabcolsep}{6pt}
\renewcommand{\arraystretch}{1.05}
\caption{\textbf{Chameleon-7B 1D VL-axis probe.} Per-region mean accuracy and $z$-score range against $22$ random unit-norm direction probes. The hub at L5 reaches the maximum accuracy of $1.000$.}
\label{tab:cham7b-probe}
\begin{tabular}{l c c l}
\toprule
\textbf{Layer region} & \textbf{Mean acc} & \textbf{$z$ vs.\ random} & \textbf{Note} \\
\midrule
L0--L6 (early, includes hub) & $0.944$ & up to $+1.01$ & L5 (hub) $=1.000$ \\
L7--L15 (mid) & $0.982$ & up to $+1.45$ & sustained signal \\
L16--L31 (late) & $0.683$ & near or below baseline & axis no longer aligned \\
\bottomrule
\end{tabular}
\end{table}

\paragraph{Implication.}
At Chameleon-7B's causal hub (L5), the VL-axis 1D probe is fully informative ($\mathrm{acc}{=}1.000$) and intervention along the same axis produces the largest single-layer DD in the model (\S\ref{sec:cross-arch-dd}). \emph{Decodability and causal handle co-localize at the hub.} The standalone 1D probe is therefore a usable, but not a sufficient, localizer for the hub: it identifies a wide early region (L0--L15) within which the hub falls but does not single out L5 specifically and gives no information about whether intervention will succeed.

\subsection{BAGEL understanding path: localized decodability, null 1D causality}
\label{app:probe-bagel-und}

The complementary pattern appears in BAGEL's understanding path.
Table~\ref{tab:bagel-und-probe} reports $1$D probe accuracy across the hub
block; the probe peaks at L25--L27 with statistically reliable separation
from the random baseline.

\begin{table}[h]
\centering
\footnotesize
\setlength{\tabcolsep}{6pt}
\renewcommand{\arraystretch}{1.05}
\caption{\textbf{BAGEL und-path 1D probe at the hub block.} Per-layer accuracy with $z$-score against $22$ random direction probes. The probe peaks in the last three layers; 1D intervention along the same axis is null at every layer in this block, and rank-$k$ ablation is non-selective.}
\label{tab:bagel-und-probe}
\begin{tabular}{l c c}
\toprule
\textbf{Layer} & \textbf{1D probe acc} & \textbf{$z$ vs.\ random} \\
\midrule
L25 & $0.701$ & $+1.52$ \\
L26 & $0.736$ & $+1.33$ \\
\textbf{L27} & $\mathbf{0.729}$ & $\mathbf{+2.65}$ \\
\midrule
\multicolumn{3}{l}{\emph{1D DiffMean intervention (block L21--27, $\alpha{=}0.3$)}: DD $=-7.3$ pp ($z=-1.25$, null)} \\
\multicolumn{3}{l}{\emph{Rank-16 PCA ablation at hub}: damaging but control-dominated (spatial $-19.7$ vs.\ control $-32.3$ pp)} \\
\bottomrule
\end{tabular}
\end{table}

The probe is informative at late layers, but the 1D intervention is null there, and no rank-$k$ subspace exposes a selective handle either: at $k{=}16$, removal damages controls more than spatial tasks (\S\ref{app:anchors}, anchor 1). This is the geometric opposite of Chameleon-7B: decodability localizes late, while no tested linear handle---1D or rank-$k$---is spatially selective in this path.

\subsection{Synthesis: probing alone is not a sufficient diagnostic}
\label{app:probe-synthesis}

Table~\ref{tab:probe-vs-hub-summary} contrasts the two cases.

\begin{table}[h]
\centering
\footnotesize
\setlength{\tabcolsep}{4pt}
\renewcommand{\arraystretch}{1.05}
\caption{\textbf{Two dissociation patterns between probing and causal handle.} Probe accuracy at the hub layer alone does not predict intervention success: Chameleon-7B's hub probe is informative and 1D intervention works; BAGEL und's hub probe is informative but neither 1D nor rank-$k$ intervention yields a spatially selective effect.}
\label{tab:probe-vs-hub-summary}
\begin{tabularx}{\linewidth}{@{} l X X @{}}
\toprule
\textbf{Aspect} & \textbf{Chameleon-7B (L5 hub)} & \textbf{BAGEL und (L21--26 hub)} \\
\midrule
Full-dim probe & $1.000$ at every layer & $\geq 0.999$ at every layer \\
1D probe at the hub & $1.000$ (informative) & $0.729$ ($+2.65\sigma$, informative) \\
1D intervention at the hub & DD $=+33.2$ pp (L5; block $z=+12.22$) & DD $=-7.3$ pp, $z=-1.25$ (null) \\
Rank-$16$ intervention & not needed (1D succeeds) & control-dominated ($-32.3$ vs.\ $-19.7$ pp); not selective \\
Diagnostic shape & Probe and causality co-localize & Probe localizes; no selective handle at 1D or rank-$k$ \\
\bottomrule
\end{tabularx}
\end{table}

\paragraph{Two cautionary lessons.}
\textbf{(1)} A high probe accuracy does not imply a usable causal handle along the same direction (BAGEL und). \textbf{(2)} A uniform full-dimensional probe across all layers does not imply that all layers are causally equivalent (Chameleon-7B). Each lesson rules out one direction of inference from probing to causal structure. The four-check protocol in \S\ref{sec:eval} therefore requires intervention evidence in addition to probing; main-text \S\ref{sec:representation-properties} formalizes the geometric properties (concentration, alignment, semantic location) that explain when probing and intervention agree.

\paragraph{Practical heuristic for new architectures.}
A probe sweep is cheap; an intervention sweep is expensive. From the Chameleon-7B pattern, a usable heuristic is to locate the layers where a 1D probe along the candidate axis is informative for task identification, and intervene in that range first. From the BAGEL und pattern, a null result at the probe peak should not end the inquiry: rank-$k$ subspace intervention is the natural next step---although in BAGEL's understanding path even rank-$k$ removal turns out to be non-selective, which is itself diagnostic (the information is entangled with general capacity rather than absent).

\subsection{Within-task probes: spatial state, not task identity or lexical cues}
\label{app:within-task-probes}

The cross-task probes above (\texttt{navigate} vs.\
\texttt{boolean\_expressions}) could in principle exploit vocabulary or
input format rather than spatial content; indeed, the full-dimensional
cross-task probe is saturated at every layer partly because of vocabulary
divergence (\S\ref{app:probe-cham7b}). To remove this confound, we probe
\emph{only navigate examples}, holding task identity and input format
fixed, and predict parametric spatial state variables.

\paragraph{Exact spatial labels.}
Each of the 250 navigate instructions is parsed by an exact simulator
that replays the instruction sequence and computes the final state. The
simulated return-to-start answer matches the gold target on
\textbf{250/250} examples in both the Chameleon and BAGEL sets, so the
derived labels are exact rather than heuristic. The probe targets are
(i) \emph{final endpoint relation}: whether the endpoint lies ahead of or
behind the start (\texttt{dy\_sign}; $n{=}126$ examples with nonzero
displacement; majority class $0.67$), and (ii) \emph{final displacement
magnitude}: whether $|dx|+|dy|$ exceeds the median ($n{=}250$; majority
$0.50$). Both targets integrate the full instruction sequence. Probes are
logistic regressions with 5-fold stratified cross-validation on hidden
states at the last prompt token.

\begin{table}[h]
\centering
\scriptsize
\setlength{\tabcolsep}{3.5pt}
\renewcommand{\arraystretch}{1.08}
\caption{\textbf{Within-task probes across all 10 configurations.}
Full-dimensional probe accuracy for the endpoint ahead/behind target
(majority baseline $0.67$) and the displacement-magnitude target
(majority $0.50$), best probed layer; for Chameleon-7B and BAGEL-und the
causal-hub layer is shown (L5/L27). Because the ahead/behind label is
unbalanced, we also report balanced accuracy on pooled out-of-fold
predictions with paired-bootstrap 95\% CIs ($10{,}000$ resamples, fixed
seed); for the balanced displacement target, balanced and plain accuracy
coincide. Task identity and input format are fixed by construction, so
cross-task lexical separation cannot explain these accuracies.}
\label{tab:within-task-probes}
\begin{tabular}{lccc}
\toprule
\textbf{Configuration} & \textbf{Endpoint acc.} & \textbf{Endpoint balanced acc.\ [95\% CI]} & \textbf{Displacement acc.\ [95\% CI]} \\
\midrule
Chameleon-7B (L5 hub) & 0.842 & 0.807 [0.726, 0.881] & 0.664 [0.605, 0.722] \\
BAGEL-und (L27 hub)   & 0.825 & 0.795 [0.714, 0.872] & 0.784 [0.733, 0.835] \\
Qwen3-VL-8B           & 0.842 & 0.819 [0.741, 0.891] & 0.828 [0.781, 0.873] \\
Qwen3-VL-2B           & 0.881 & 0.849 [0.774, 0.917] & 0.764 [0.711, 0.816] \\
Qwen3-VL-4B           & 0.872 & 0.855 [0.784, 0.921] & 0.796 [0.745, 0.844] \\
Qwen3-VL-32B          & 0.889 & 0.867 [0.796, 0.931] & 0.832 [0.786, 0.877] \\
Qwen3-VL-30B-A3B      & 0.881 & 0.855 [0.782, 0.922] & 0.828 [0.780, 0.873] \\
Emu3-Chat             & 0.904 & 0.885 [0.817, 0.945] & 0.760 [0.706, 0.811] \\
Liquid-V1-7B          & 0.834 & 0.807 [0.728, 0.881] & 0.780 [0.728, 0.832] \\
Janus-Pro-7B (gen)    & 0.857 & 0.825 [0.748, 0.896] & 0.728 [0.673, 0.781] \\
\midrule
Majority baseline     & 0.67  & 0.50 (chance) & 0.50 \\
\bottomrule
\end{tabular}
\end{table}

Across all 10 configurations, the endpoint target is decoded at
$0.825$--$0.904$ against a $0.67$ majority baseline, and displacement
magnitude at $0.66$--$0.83$ against $0.50$. Balanced accuracy sits
$0.02$--$0.03$ below plain accuracy, and its lower CI bound remains well
above the majority baseline in every configuration, so the decodability
claim survives label-imbalance correction. One caveat: outside
Chameleon-7B and BAGEL, the best layer for the endpoint target is the
embedding layer (L0), whereas the displacement target peaks at late
layers; probe localization should therefore not be read as hub
localization. In contrast, the paper's 1D causal axes are weak readouts of these
same variables (mostly $z<3$ against 22 random directions). The axis is a
causal \emph{carrier}, not a content \emph{register}: within-task state
information is present in the residual stream, while the direction that
causally matters is not the direction that best reads that state out.

\paragraph{Interpreting model failure.}
This distinction changes how benchmark failures should be read. Emu3
decodes the endpoint state at $0.904$ while answering only 12--18\% of
the same navigate items correctly, so its failure cannot be explained by
a missing decodable representation; the bottleneck must lie in causal
access, routing, or downstream readout. Benchmark accuracy alone cannot
separate a model that lacks a representation from one that encodes it but
fails to use it; probing plus intervention can.

\subsection{The decodability--causality dissociation across three architectures}
\label{app:three-arch-dissociation}

To test whether the dissociation generalizes beyond Chameleon-7B and
BAGEL, we extended the probe--intervention comparison to Qwen3-VL-8B
(hidden states at the last prompt token, $n{=}250$ navigate $+$ 250
boolean examples, all 36 layers).

\begin{table}[h]
\centering
\footnotesize
\setlength{\tabcolsep}{5pt}
\renewcommand{\arraystretch}{1.08}
\caption{\textbf{Qwen3-VL-8B: the causally strongest direction is not the
most decodable one.} Task-identity decoding (navigate vs.\ boolean) at
the causal hub L8, against 1D probes on random unit directions.}
\label{tab:qwen8b-dissociation}
\begin{tabular}{lcc}
\toprule
\textbf{Direction (at hub L8)} & \textbf{Task-identity acc.} & \textbf{Causal role} \\
\midrule
Random 1D directions (mean) & 0.825 & none (0/100 match real axis) \\
Causal DiffMean axis & 0.540 ($z=-1.49$ vs.\ random) & $z=+38.3$ under intervention \\
Full-dimensional probe & 1.000 & --- \\
\bottomrule
\end{tabular}
\end{table}

Task identity is so lexically salient that \emph{random} 1D directions
decode it at $0.825$ on average, and a full-dimensional probe is
saturated at every layer. Yet the causal axis—extremely strong under
intervention at the hub ($z=+38.3$ against 100 norm-matched
directions)—decodes task identity at only $0.540$, \emph{below} the
random-direction mean. The direction that best separates the two tasks is
therefore not the direction that controls spatial reasoning. Conversely,
the 1D causal axes are weak within-task state readouts (mostly $z<3$;
\S\ref{app:within-task-probes}). Across Chameleon-7B, BAGEL, and
Qwen3-VL-8B, decodability and causal importance dissociate in both
directions, and causal interpretation requires intervention together with
the concentration, alignment, and semantic-location diagnostics of
\S\ref{sec:representation-properties}.

\section{Hub-Specific Recovery of Spatial Information}
\label{app:rescue}

\paragraph{What the rescue and amplification evidence does and does not establish.}
The rescue results support three distinct conclusions that should not be
conflated. \emph{(i) Chameleon-7B shows partial local recoverability.}
Hub removal reduces \texttt{navigate} from 47.2\% to 34.0\%, and L6/L7
reinjection restores $+8.0$/$+7.2$ pp (60.6\%/54.5\% of the loss;
Table~\ref{tab:rescue-layer}). Because 22 norm-matched random directions
recover $+6.8\pm6.8$ pp at the same dosage
(\S\ref{app:rescue-direction}), this establishes local recoverability
rather than unique direction-specific recovery; direction specificity
rests on the removal-side random and energy-matched controls, not on the
rescue. Stronger injections collapse (Table~\ref{tab:rescue-dosage}).
\emph{(ii) BAGEL's generation path exposes a compact 1D carrier:} its 1D
DiffMean axis is causally selective (DD $=+16.8$ pp, $z=+3.19$;
Table~\ref{tab:dual_path_geometry}). \emph{(iii) BAGEL's understanding
path is decodable but not selectively intervenable.} Its 1D DiffMean axis
is null (DD $=-7.3$ pp, $z=-1.25$), and rank-16 PCA ablation is damaging
but control-dominated (spatial $-19.7$ vs.\ control $-32.3$ pp;
DD $=-12.7$, CI $[-21.7,-4.0]$); cross-path rank-16 amplification
recovers spatial and control performance comparably ($+6.92$/$+5.07$ pp;
Table~\ref{tab:cross-path-rescue}). The information is present,
decodable, and partially recoverable, but entangled with general capacity
rather than organized as a substitute carrier.

\paragraph{Note on the three navigate-drop figures.}
The $-27.6$, $-20.7$, and $-13.2$ pp \texttt{navigate} drops reported for
Chameleon-7B correspond to three distinct protocols run in different
sessions with slightly different baselines: the single-layer scan at the
L5 hub ($n{=}250$, baseline 42.0\%), the frozen 27-task L0--7 run
($n{=}150$, baseline 43.3\%; Appendix~\ref{app:cham7b-bbh27}), and the L5
ablation inside the paired rescue run ($n{=}250$, baseline 47.2\%;
Table~\ref{tab:rescue-layer}), respectively.

This appendix expands on the rescue experiment in main-text \S\ref{sec:rescue}: removing the L5 axis component impairs spatial reasoning, and reinjecting that component at L6 restores a sizable fraction of the lost performance. The experiment establishes two facts: (i) spatial information is concentrated in the early hub---L5 is the strongest single-layer site ($-27.6$ pp in the layer scan, with L6/L7 also substantial at $-20.8$/$-21.2$ pp and L0--L4 $\leq 6.0$ pp), while the non-hub L20 control is null---and (ii) the lost performance can be partially recovered by reinjection in the immediately downstream layer, indicating that the information is in a recoverable form rather than destructively erased.

\subsection{Dosage calibration}
\label{app:rescue-sigma}

Reinjection dosage is reported in units of $\sigma$, the standard deviation of the per-token projection of hidden states onto the canonical VL axis at the hub layer. We calibrated $\sigma$ at L5 using $n{=}200$ \texttt{navigate} sequences, $13{,}236$ token positions in total, yielding $\sigma=8.17$ (a previous calibration with $n{=}50$ sequences gave $\sigma=8.16$, providing a consistency check). We use $\beta{=}0.5\sigma{=}4.08$ as the canonical reinjection dosage; larger values induce non-selective collapse (\S\ref{app:rescue-dosage}).

\subsection{Layer specificity: L5 is the strongest single-layer hub}
\label{app:rescue-layer}

Table~\ref{tab:rescue-layer} contrasts the L5 hub with a non-hub layer (L20) in the paired rescue run. In this run, L5 ablation drops \texttt{navigate} accuracy by $13.2$ pp; the same ablation operation at L20 produces no detectable effect ($+0.8$ pp, within baseline noise). The headline single-layer scan reports a larger $27.6$ pp drop, while this paired run is used for the rescue comparison.

\begin{table}[h]
\centering
\footnotesize
\setlength{\tabcolsep}{6pt}
\renewcommand{\arraystretch}{1.05}
\caption{\textbf{Hub specificity in the paired rescue experiment.} The same canonical-axis ablation at $\alpha{=}1.0$ produces a substantial drop at L5 (the causal hub) but is null at L20. Reinjection of the canonical axis at the immediately downstream layer restores spatial performance at L5--L6 but not at L20--L21, confirming that the recoverable information is hub-specific.}
\label{tab:rescue-layer}
\begin{tabular}{l c c}
\toprule
\textbf{Condition} & \textbf{navigate accuracy} & $\Delta$ vs.\ baseline \\
\midrule
Baseline (no intervention) & $47.2\%$ & --- \\
\midrule
\textbf{L5 ablation ($\alpha{=}1.0$, hub)} & $\mathbf{34.0\%}$ & $\mathbf{-13.2}$ \textbf{pp} \\
L5 ablation $+$ L6 reinjection ($\beta{=}0.5\sigma$) & $42.0\%$ & $-5.2$ pp (recovered $+8.0$ pp) \\
L5 ablation $+$ L7 reinjection ($\beta{=}0.5\sigma$) & $41.2\%$ & $-6.0$ pp (recovered $+7.2$ pp) \\
\midrule
L20 ablation ($\alpha{=}1.0$, non-hub) & $48.0\%$ & $+0.8$ pp (null) \\
L20 ablation $+$ L21 reinjection ($\beta{=}0.5\sigma$) & $45.6\%$ & $-1.6$ pp (no recovery effect) \\
\bottomrule
\end{tabular}
\end{table}

\paragraph{Reading the table.} The clean comparison is L5 ablation vs.\ L20 ablation: the same operation at the same dosage produces a $13.2$ pp drop at one layer and no effect at another. This is direct evidence that the spatial information targeted by the canonical axis is concentrated in the early hub region (strongest at L5) rather than present throughout the network. The reinjection rows further establish that the information at L5 is in a recoverable form: a $\beta{=}0.5\sigma$ injection at L6 restores $60.6\%$ of the lost performance, and the same restoration is unavailable when neither the ablation nor the reinjection target the hub (the L20--L21 control row).

\subsection{Dosage scan}
\label{app:rescue-dosage}

Table~\ref{tab:rescue-dosage} reports the $\beta$-sweep at the hub. The $\beta{=}0.5\sigma$ choice in the paired rescue experiment lies in the productive regime; larger doses overshoot and induce non-selective collapse, consistent with the general pattern that intervention magnitudes at $1\sigma$ or above approach the residual-norm budget at this layer.

\begin{table}[h]
\centering
\footnotesize
\setlength{\tabcolsep}{6pt}
\renewcommand{\arraystretch}{1.05}
\caption{\textbf{Reinjection dosage scan at L6 after L5 ablation.} The productive regime is at $\beta{=}0.5\sigma$.}
\label{tab:rescue-dosage}
\begin{tabular}{l c c}
\toprule
\textbf{$\beta$} & \textbf{navigate accuracy} & \textbf{Status} \\
\midrule
$0.5\sigma$ ($4.08$) & $42.0\%$ & productive (recovers $+8.0$ pp) \\
$1.0\sigma$ ($8.17$) & $0.0\%$ & overshoot (token-prediction collapse) \\
\bottomrule
\end{tabular}
\end{table}
 \subsection{Additive versus multiplicative steering}
  \label{app:additive-vs-multiplicative}

  We compare two forms of positive-direction intervention along the VL
  axis. Multiplicative amplification rescales each token's existing
  projection:
  \begin{equation}
    h \leftarrow h + \beta\,(h^\top v)\,v, \quad \beta > 0,
  \end{equation}
  where the magnitude of the perturbation varies per token according to
  that token's own projection. Additive steering instead applies a
  uniform offset independent of each token's current representation:
  \begin{equation}
    h \leftarrow h + \beta\, v.
  \end{equation}
  Both are tested on Chameleon-7B (early block L0--7) and Qwen3-VL-8B
  (early block L0--8; additionally at the identified causal hub L8 alone),
  at $n{=}150$ samples per task.

  \paragraph{Results.}
  Table~\ref{tab:additive-vs-multiplicative} summarises the outcomes.
  For reference, the multiplicative \emph{removal} condition
  ($h \leftarrow h - \alpha(h^\top v)v$, $\alpha{=}1.0$) is included in
  each block.

  \begin{table}[h]
  \centering
  \small
  \setlength{\tabcolsep}{5pt}
  \renewcommand{\arraystretch}{1.05}
  \caption{\textbf{Additive versus multiplicative steering along the VL
  axis.} $\Delta$ denotes accuracy change relative to baseline.
  Spatial task: navigate (Chameleon-7B) or
  colored-objects (Qwen3-VL-8B).
  Control task: boolean-expressions (Chameleon-7B) or
  web-of-lies (Qwen3-VL-8B).
  DD $= \Delta_\text{control} - \Delta_\text{spatial}$ (positive DD indicates spatial-selective damage, consistent with the main-text convention).
  Hub injection uses $n{=}247$ at L8 only.}
  \label{tab:additive-vs-multiplicative}
  \begin{tabularx}{\linewidth}{@{}l l r r r r@{}}
  \toprule
  \textbf{Model / block} & \textbf{Condition} &
  $\boldsymbol{\beta}$ &
  $\boldsymbol{\Delta_\text{spatial}}$ &
  $\boldsymbol{\Delta_\text{control}}$ &
  \textbf{DD} \\
  \midrule
  \multicolumn{6}{@{}l}{\textit{Chameleon-7B, early block L0--7
  (baseline: nav\,=\,37.3\%, bool\,=\,35.3\%)}} \\
  & Removal ($-\alpha$) & 1.0 & $-$22.0\,pp & $-$2.0\,pp & $+$20.0\,pp \\
  & Multiplicative ($+\beta$) & 0.3 & $-$15.3\,pp & $+$0.7\,pp & $+$16.0\,pp \\
  & Multiplicative ($+\beta$) & 0.5 & $-$30.7\,pp & $-$14.0\,pp & $+$16.7\,pp \\
  & Multiplicative ($+\beta$) & 1.0 & $-$37.3\,pp & $-$35.3\,pp & $+$2.0\,pp \\
  & Additive shift & 0.3 & $-$4.7\,pp & $-$4.7\,pp & $\phantom{+}$0.0\,pp \\
  & Additive shift & 0.5 & $-$24.0\,pp & $-$9.3\,pp & $+$14.7\,pp \\
  \midrule
  \multicolumn{6}{@{}l}{\textit{Qwen3-VL-8B, early block L0--8
  (baseline: co\,=\,81.3\%, wol\,=\,44.0\%)}} \\
  & Removal ($-\alpha$) & 1.0 & $-$28.7\,pp & $+$1.3\,pp & $+$30.0\,pp \\
  & Multiplicative ($+\beta$) & 0.1 & $-$0.7\,pp & $\phantom{+}$0.0\,pp & $+$0.7\,pp \\
  & Multiplicative ($+\beta$) & 0.3 & $\mathbf{+}$\textbf{3.3\,pp} & $-$2.0\,pp & $-$5.3\,pp \\
  & Multiplicative ($+\beta$) & 0.5 & $+$2.7\,pp & $+$2.0\,pp & $-$0.7\,pp \\
  & Multiplicative ($+\beta$) & 1.0 & $-$67.3\,pp & $-$44.0\,pp & $+$23.3\,pp \\
  \midrule
  \multicolumn{6}{@{}l}{\textit{Qwen3-VL-8B, hub injection at L8 only
  (baseline: co\,=\,82.2\%, wol\,=\,48.6\%, $n{=}247$)}} \\
  & Removal ($-\alpha$) & 1.0 & $-$75.7\,pp & $+$2.0\,pp & $+$77.7\,pp \\
  & Multiplicative ($+\beta$) & 0.05--0.5 & $\leq|{-}0.8|$\,pp & $\leq|{-}1.2|$\,pp & null \\
  & Additive shift & 0.5--10 & $\leq|{-}1.2|$\,pp & $\leq|{+}0.8|$\,pp & null \\
  \bottomrule
  \end{tabularx}
  \end{table}

  \paragraph{Summary of findings.}

  \textbf{Removal is strongly asymmetric.}
  Across both models and all layer configurations, multiplicative removal
  selectively disrupts spatial-imagery performance while leaving or
  improving symbol-manipulation controls, with effects of $-$20 to
  $-$76\,pp on the spatial task. Neither positive-direction mode matches
  this magnitude or selectivity.

  \textbf{No positive regime for Chameleon-7B.}
  On Chameleon-7B, multiplicative amplification damages the spatial task
  at every tested $\beta$ (navigate drops $-$15 to $-$37\,pp). Additive
  shift at $\beta{=}0.3$ is near-symmetric ($-$4.7\,pp on both spatial
  and control), while larger magnitudes produce non-selective degradation.
  Neither mode improves navigation accuracy.

  \textbf{Weak bidirectional signal for Qwen3-VL-8B (early block only).}
  On Qwen3-VL-8B, multiplicative amplification at $\beta{=}0.3$ yields a
  weak positive signal: colored-objects improves by $+$3.3\,pp while
  web-of-lies declines by $-$2.0\,pp (DD\,$=$\,$-$5.3\,pp under the
  main-text convention, i.e., the spatial task improves relative to the
  control). This is
  substantially smaller than the removal effect ($-$28.7\,pp;
  a ratio of approximately 1:9), and the effect disappears at
  $\beta{=}1.0$ where non-selective collapse occurs.

  \textbf{Hub injection is null for both modes.}
  Targeted injection at the identified causal hub (L8) with both
  multiplicative amplification ($\beta{=}0.05$--$0.5$) and additive shift
  ($\beta{=}0.5$--$10$) produces changes of at most $1.2$\,pp on any
  task—indistinguishable from zero—despite the same hub supporting a
  $-$75.7\,pp selective disruption under removal.

  \paragraph{Interpretation.}
  The hub-injection null result, combined with the large removal effect,
  is consistent with a saturation account: the model already exploits the
  VL-axis projection at or near capacity during spatial-imagery reasoning,
  so amplifying the existing signal yields no further gain. On
  Chameleon-7B, where the VL axis is derived from embedding centroids
  rather than per-layer DiffMean activations, the axis may carry broad
  modality-separation information rather than a precise spatial-imagery
  carrier; amplifying it then disrupts rather than supports spatial
  reasoning. The small positive effect observed for Qwen3-VL-8B
multiplicative amplification ($+$3.3\,pp) suggests that the DiffMean
axis is a more targeted spatial carrier in that model, but the strong
removal--amplification asymmetry indicates that the causal role of the
axis is primarily revealed by removal rather than by positive steering.
Together with the failure of additive shifts, this suggests that the
useful signal lies in the existing projection pattern along the axis, not
in a uniform global offset.

\subsection{Note on direction-specific recovery}
\label{app:rescue-direction}

For completeness, we tested whether reinjecting along randomly sampled directions at the hub-downstream layer (L6) reproduces the recovery effect. At $\beta{=}0.5\sigma$, with $22$ norm-matched random unit directions (seeds $0$--$21$), the mean recovery is $+6.8 \pm 6.8$ pp, with a wide spread across seeds. The canonical VL-axis recovery of $+8.0$ pp lies above this distribution mean, but the difference is within the seed-to-seed variance.

We interpret this as a feature of the dosage regime rather than a refutation of the canonical axis: at $\beta{=}0.5\sigma$ in $d{=}4096$, partial restoration of residual-stream energy at the hub-downstream layer suffices to relieve some of the L5 ablation damage, and the canonical axis is one of many directions that do this. The principal evidence in this experiment is the layer-specific contrast in Table~\ref{tab:rescue-layer}, not a claim of unique direction-specific recovery at this dosage.

\paragraph{Headline claim.} The experiment establishes that the spatial impairment is hub-localized (L5 is the strongest single-layer site; the non-hub L20 is null) and that the impairment is recoverable. Direction-specific recovery at lower dosages is a separate question that the present experiment does not adjudicate.

\section{Cross-Task Path Separation: Extended Evidence}
\label{app:cross_task_path}

Main-text \S\ref{sec:cross-task-path} reports that on BAGEL, the gen-path intervention damages performance on spatial-reasoning datasets several times more than the und-path intervention. This appendix gives the per-task data behind that claim and reports two further findings: a complexity-dependent compression of the path-separation ratio in multi-hop spatial reasoning, and an architecture-dependent variation in path dominance documented in Janus-Pro.

\subsection{BAGEL cross-path rescue}
\label{app:cross-path-rescue}

Main-text \S\ref{sec:gen-vs-und} reports an asymmetric rescue experiment
on BAGEL. After removing the compact generation-path carrier, we test
whether the understanding-path subspace can compensate for the lost
spatial information. We amplify the rank-$16$ PCA subspace from the
understanding path rather than reinjecting the removed generation-path
direction.

The first pilot used a stronger generation-path ablation
($\alpha_{\mathrm{gen}}=1.0$) and a $1$D understanding-path direction.
This setting caused near-complete collapse and did not provide a useful
rescue test. We therefore use a milder generation-path ablation
($\alpha_{\mathrm{gen}}=0.5$) and a rank-$16$ understanding-path basis.

\begin{table}[h]
\centering
\footnotesize
\setlength{\tabcolsep}{4pt}
\renewcommand{\arraystretch}{1.05}
\caption{\textbf{BAGEL cross-path rescue.} After generation-path ablation,
amplifying the rank-$16$ understanding-path subspace partially recovers
performance. The recovery is real but not spatial-selective, since
control tasks recover by a comparable amount.}
\label{tab:cross-path-rescue}
\begin{tabular}{l c c c c c}
\toprule
\textbf{Condition} & $\boldsymbol{\alpha_{\mathrm{gen}}}$ &
$\boldsymbol{\alpha_{\mathrm{und}}}$ &
\textbf{Navigate} & \textbf{Geometric} &
\textbf{Control avg.} \\
\midrule
Baseline & $0$ & $0$ & $58.0$ & $36.0$ & $77.6$ \\
Gen-path ablation & $0.5$ & $0$ & $3.6$ & $18.8$ & $44.0$ \\
Rescue & $0.5$ & $-0.10$ & $4.8$ & $33.6$ & $49.1$ \\
\bottomrule
\end{tabular}
\end{table}

Relative to the generation-path ablation condition, the rescue condition
recovers spatial performance by $+6.92$ pp with bootstrap CI
$[+4.53,+9.33]$, and recovers control performance by $+5.07$ pp with
bootstrap CI $[+1.87,+8.27]$. The two confidence intervals overlap,
indicating that the understanding path contains recoverable task-relevant
information, but in a broader and less selective form than the compact
generation-path carrier.

\subsection{BAGEL: gen-path is the dominant carrier on single-hop spatial tasks}
\label{app:cross-task-bagel}

Table~\ref{tab:cross-task-bagel} reports per-task interventions on BAGEL-7B-MoT for both paths at $\alpha{=}0.5$, final block (L21--27), $n{=}250$ per task. Each row pairs the gen-path and und-path effects on the same dataset; the rightmost column gives the ratio $|\Delta_{\text{gen}}| / |\Delta_{\text{und}}|$.

\begin{table}[h]
\centering
\footnotesize
\setlength{\tabcolsep}{4pt}
\renewcommand{\arraystretch}{1.05}
\caption{\textbf{BAGEL gen-path vs.\ und-path on three external spatial-reasoning benchmarks.} On single-hop tasks (bAbI 17/19/20, SpartQA, StepGame $k{=}1$), gen-path damage exceeds und-path damage by $6$--$31\times$. Multi-hop StepGame ($k{=}5, 9$) shows a compressed ratio --- discussed in \S\ref{app:cross-task-multihop}.}
\label{tab:cross-task-bagel}
\begin{tabular}{l l c c c}
\toprule
\textbf{Dataset} & \textbf{Subtask} & \textbf{$\Delta_{\text{gen}}$} & \textbf{$\Delta_{\text{und}}$} & \textbf{ratio} \\
\midrule
\multicolumn{5}{l}{\emph{Single-hop spatial reasoning}} \\
\midrule
bAbI & task 17 (positional reasoning) & $-6.8$ pp & $+0.8$ pp & --- (sign-reversed) \\
bAbI & task 19 (path finding) & $-12.4$ pp & $-0.4$ pp & $\mathbf{31\times}$ \\
bAbI & task 20 (agent motivation) & $-43.2$ pp & $-6.8$ pp & $\mathbf{6.4\times}$ \\
SpartQA & yes/no + multi-choice & $-12.0$ pp & $-0.4$ pp & $\mathbf{30\times}$ \\
StepGame & $k{=}1$ (single hop) & $-18.4$ pp & $-2.4$ pp & $\mathbf{7.7\times}$ \\
\midrule
\multicolumn{5}{l}{\emph{Multi-hop spatial reasoning}} \\
\midrule
StepGame & $k{=}5$ ($5$ hops) & $-16.8$ pp & $-13.2$ pp & $1.3\times$ \\
StepGame & $k{=}9$ ($9$ hops) & $-4.8$ pp & $-4.4$ pp & $1.1\times$ \\
\bottomrule
\end{tabular}
\end{table}

\paragraph{Single-hop tasks support the gen-path-dominant view.}
On all five single-hop variants (the top block), the gen-path intervention damages spatial reasoning $6$--$31\times$ more than the und-path intervention. The bAbI task 17 row, where und shows a small positive change ($+0.8$ pp) while gen drops by $-6.8$ pp, has a sign-reversed ratio that we report as ``---'' rather than $\infty$; the qualitative direction is consistent with the rest of the single-hop block.

\subsection{Multi-hop reasoning recruits both paths}
\label{app:cross-task-multihop}

On multi-hop StepGame ($k{=}5$, $k{=}9$), the gen/und ratio compresses to $1.1$--$1.3\times$. Two observations frame this finding.

First, the absolute gen-path damage on multi-hop variants ($-16.8$ pp at $k{=}5$, $-4.8$ pp at $k{=}9$) remains comparable to the single-hop case ($-18.4$ pp at $k{=}1$); the gen-path is still impacted. What changes is that the und-path damage rises in tandem.

Second, multi-hop spatial reasoning is task-structurally different from single-hop: reaching a $k{=}9$-step spatial conclusion requires iterated application of intermediate spatial relations rather than a single atomic spatial-mapping operation. This is consistent with iterative reasoning recruiting both representational paths --- the gen-path for the underlying spatial scene, the und-path for the symbolic transition rules --- in tandem rather than in sequence. Under this view, the path-separation effect we measure is sharpest where a single read-off from a constructed spatial scene suffices, and dilutes when the task has a large symbolic-reasoning component layered on top of the spatial substrate.

Note also that multi-hop StepGame baselines are low ($32.4\%$ at $k{=}5$, $19.2\%$ at $k{=}9$), so absolute $\Delta$ values carry larger relative noise than in the single-hop case; this contributes to the compressed ratio. We do not lean on this point: the structural argument in the previous paragraph is the substantive interpretation. We view this as a post-observation interpretation rather than a causal claim; controlled training-objective or path-isolation ablations are needed to test it directly.

\subsection{Janus-Pro 27-Task BBH Analysis}
\label{app:janus-bbh27}

Janus-Pro-7B provides an independent test of the generation-path
hypothesis on the full BBH task suite. We evaluate 27 BBH tasks with
$n=150$ examples per task, $\alpha=0.5$, layers L21--29, and 0-shot
decoding. This run uses an earlier fixed generation-path configuration
rather than the main-table one (L22--28, $\alpha=0.7$;
Table~\ref{tab:main-effect}); both dosages lie inside Janus's
significant range $\alpha\in[0.3,0.7]$
(Appendix~\ref{app:alpha-dosage}), and neither run is retuned per task. The generation-path axis is constructed from the VQ-16 visual
codes, whereas the understanding-path axis is constructed from the SigLIP
understanding encoder. Table~\ref{tab:janus-bbh27} reports the full
per-task results.

\begin{table}[t]
\centering
\scriptsize
\setlength{\tabcolsep}{3.2pt}
\renewcommand{\arraystretch}{1.08}
\caption{\textbf{Janus-Pro-7B gen-path vs. und-path intervention on 27 BBH tasks.}
The generation-path axis produces a strong drop on \texttt{navigate}
($-28.7$ pp) but only a mild drop on \texttt{boolean\_expressions}
($-7.3$ pp), reproducing the qualitative spatial-selective effect on a
broader task suite.}
\label{tab:janus-bbh27}
\resizebox{\textwidth}{!}{
\begin{tabular}{lrrrrl}
\toprule
Task & Baseline & Gen $\Delta$ & Und $\Delta$ & Gen--Und & Pattern \\
\midrule
\texttt{boolean\_expressions} & 98.7 & $-7.3$ & $-4.7$ & $-2.7$ & both \\
\texttt{causal\_judgement} & 64.0 & $-4.7$ & $-10.7$ & $+6.0$ & und $>$ gen \\
\texttt{date\_understanding} & 40.0 & $-0.7$ & $-6.7$ & $+6.0$ & und $>$ gen \\
\texttt{disambiguation\_qa} & 50.7 & $-1.3$ & $-2.7$ & $+1.3$ & $\sim$ \\
\texttt{dyck\_languages} & 12.0 & $0.0$ & $-3.3$ & $+3.3$ & $\sim$ \\
\texttt{formal\_fallacies} & 24.7 & $-10.0$ & $-2.7$ & $-7.3$ & gen $>$ und \\
\texttt{geometric\_shapes} & 100.0 & $-16.0$ & $0.0$ & $-16.0$ & gen $\gg$ und \\
\texttt{hyperbaton} & 32.7 & $+1.3$ & $-16.7$ & $+18.0$ & und $\gg$ gen \\
\texttt{logical\_deduction\_5\_obj} & 75.3 & $-24.0$ & $+1.3$ & $-25.3$ & gen $\gg$ und \\
\texttt{logical\_deduction\_7\_obj} & 70.0 & $-3.3$ & $+3.3$ & $-6.7$ & gen $>$ und \\
\texttt{logical\_deduction\_3\_obj} & 78.7 & $-0.7$ & $+0.7$ & $-1.3$ & $\sim$ \\
\texttt{movie\_recommendation} & 32.7 & $+5.3$ & $-4.7$ & $+10.0$ & und $>$ gen \\
\texttt{multistep\_arithmetic\_two} & 7.3 & $-2.0$ & $-5.3$ & $+3.3$ & $\sim$ \\
\texttt{navigate} & 81.3 & $-28.7$ & $-24.0$ & $-4.7$ & both, similar \\
\texttt{object\_counting} & 49.3 & $-6.7$ & $+1.3$ & $-8.0$ & gen $>$ und \\
\texttt{penguins\_in\_a\_table} & 31.5 & $-0.7$ & $-2.1$ & $+1.4$ & $\sim$ \\
\texttt{reasoning\_about\_colored\_objects} & 25.3 & $+0.7$ & $+1.3$ & $-0.7$ & $\sim$ \\
\texttt{ruin\_names} & 17.3 & $-0.7$ & $-8.7$ & $+8.0$ & und $>$ gen \\
\texttt{salient\_translation\_error} & 99.3 & $-34.0$ & $-56.0$ & $+22.0$ & und $\gg$ gen \\
\texttt{snarks} & 54.0 & $-2.7$ & $-20.7$ & $+18.0$ & und $\gg$ gen \\
\texttt{sports\_understanding} & 90.7 & $-5.3$ & $+6.0$ & $-11.3$ & gen $>$ und \\
\texttt{temporal\_sequences} & 68.0 & $-1.3$ & $-13.3$ & $+12.0$ & und $>$ gen \\
\texttt{tracking\_shuffled\_5\_obj} & 52.7 & $+6.7$ & $+20.7$ & $-14.0$ & und helps more \\
\texttt{tracking\_shuffled\_7\_obj} & 40.0 & $-7.3$ & $+16.7$ & $-24.0$ & und helps, gen hurts \\
\texttt{tracking\_shuffled\_3\_obj} & 37.3 & $-11.3$ & $+4.7$ & $-16.0$ & gen $>$ und \\
\texttt{web\_of\_lies} & 70.7 & $-9.3$ & $0.0$ & $-9.3$ & gen $>$ und \\
\texttt{word\_sorting} & 0.0 & $0.0$ & $0.0$ & $0.0$ & floor \\
\bottomrule
\end{tabular}
}
\end{table}

Overall, the generation-path intervention causes an average drop of
$-6.1$ pp across the 27 tasks, with 11 of 27 tasks dropping by more than
5 pp. The understanding-path intervention causes an average drop of
$-4.7$ pp, with 9 of 27 tasks dropping by more than 5 pp. Tasks with
$|\mathrm{Gen}-\mathrm{Und}|\ge 10$ pp show a balanced split: four are more
affected by the generation path
(\texttt{geometric\_shapes}, \texttt{logical\_deduction\_5\_obj},
\texttt{tracking\_shuffled\_3\_obj}, and
\texttt{tracking\_shuffled\_7\_obj}), and four are more affected by the
understanding path
(\texttt{salient\_translation\_error}, \texttt{hyperbaton},
\texttt{snarks}, and \texttt{temporal\_sequences}). This supports a
functional split between a compact generation-path carrier and a broader
understanding-path subspace.

The \texttt{tracking\_shuffled\_objects} family is unstable across object
counts: the generation path hurts the 3-object and 7-object variants, while
the understanding path improves the 5-object and 7-object variants. This
suggests that tracking tasks mix spatial tracking and symbolic ordering
strategies, making them less clean as a single functional category.

\subsection{Janus-Pro: a more balanced two-path architecture}
\label{app:cross-task-janus}

For comparison, on Janus-Pro the SpartQA intervention shows a different pattern: the gen-path early-block intervention has no net effect ($+2.0$ pp), while the und-path early-block intervention is mildly performance-enhancing ($+12.0$ pp), and the und-path final-block intervention is null ($-1.2$ pp). The standalone gen-path-dominance finding seen in BAGEL is not reproduced on Janus.

This is consistent with our 27-task BBH analysis (full per-task data in \S\ref{app:janus-bbh27}), where the across-task gap between gen-path and und-path effects is substantially smaller in Janus-Pro than in BAGEL. We interpret the comparison as evidence that the magnitude of path separation is architecture-dependent: BAGEL's training regime concentrates spatial information on the gen-path more sharply than Janus-Pro's. The qualitative finding that a generation-related axis is causal for spatial reasoning holds in both architectures (\S\ref{sec:gen-vs-und}), and indeed the Janus gen-path DiffMean axis remains the only carrier that passes all four causal checks in that model (\S\ref{sec:gen-vs-und}, \S\ref{app:fail-janus-early}). The Janus result does not refute the gen-path-as-carrier hypothesis; it places a quantitative caveat on cross-architecture extrapolation of the BAGEL ratio.

\section{Architecture-Specific Main-Effect Details}
\label{app:arch-specific-details}

\subsection{Emu3-Chat dual-carrier redundancy}
\label{app:emu3-dual-carrier}

Most positive cases exhibit a single causal hub, but Emu3-Chat is an
exception. Its spatial-selective effect appears in two separated layer
windows: an early block (L0--7) and a final block (L24--31), both with
the main effect at $\alpha=0.50$ (the task-spectrum runs used
$\alpha=0.10$ for the early carrier). The per-task accuracy drops induced by the two
carriers are highly correlated ($r=0.967$), and their spectrum slopes are
nearly matched: $-6.7$ pp/unit for the early carrier and $-6.0$ pp/unit
for the final carrier.

This suggests that the two blocks carry redundant copies of a similar
spatial representation rather than two unrelated task effects. Consistent
with this interpretation, ablating either block alone is sufficient to
produce a spatial-selective drop (early DD $=+20.8$ pp, final
DD $=+17.2$ pp). A complete removal of Emu3's EVT signal may therefore
require joint intervention on both layer windows, rather than the
single-hub intervention used for most other models.

\emph{Source-dependence caveat.} The real-image source swap
(Appendix~\ref{app:source-swap}) shows that the two carriers are not
equally robust. The early carrier replicates under COCO-image axes and
even strengthens at the L3 hub (single-layer DD $=+24.4$ pp,
CI $[+14.8,+34.4]$, versus $+14.0$ pp with the synthetic axis). The
final carrier does not replicate under the real-image axis
(DD $=-0.0$, CI $[-9.2,+8.8]$), consistent with the larger axis
divergence in the final window ($\cos$ falls to $0.61$ by L31). The
dual-carrier redundancy described above should therefore be read as a
property of the synthetic-token-ID axis; under input-faithful real-image
axes, Emu3 presents a single, early causal carrier.

\section{Task Construct Validity: Mental Simulation, the Visual$\times$Spatial Battery, and Matched Controls}
\label{app:construct-validity}

A central alternative explanation is that the spatial/symbolic contrast
reflects correlated task properties---baseline difficulty, reasoning
depth, reasoning structure, or the particular task pair---rather than
spatial demand. This appendix reports four complementary controls: (i) a design-based mental-simulation
battery with matched format and frozen interventions
(\S\ref{app:mental-sim}); (ii) a 2$\times$2 battery crossing spatial
demand with visual content (\S\ref{app:battery-2x2}); (iii) a
matched-depth, answer-completion-controlled dissociation
(\S\ref{app:matched-depth}); and (iv) a statistical control for baseline
accuracy across the task spectrum (\S\ref{app:spectrum-control}),
together with the complete 27-task BBH suite under a frozen intervention
(\S\ref{app:cham7b-bbh27}).

\subsection{Text-formatted mental-simulation battery}
\label{app:mental-sim}

We built text-formatted mental rotation~\citep{shepard1971mental}, paper
folding~\citep{shepard1972paperfolding}, and perspective-taking
tasks~\citep{hegarty2004dissociation} with programmatically generated,
solver-verified answers and balanced options, and additionally evaluated
mental rotation (MRT) and perspective taking (PTT) items \emph{verbatim}
from the SPACE benchmark~\citep{ramakrishnan2025space}. All tasks are
4-way multiple choice (chance $25\%$) with the same answer format,
prompt template, and scoring. Qwen3-VL-8B/32B are evaluated under the
paper's published configuration---per-layer DiffMean axes, early block,
$\alpha{=}1.0$---with no task-specific retuning
($n{=}150$ for 8B, $n{=}100$ for 32B).

\paragraph{Interpretability floor.} Cells whose baseline falls below a
pre-specified chance$+10$ pp threshold ($35\%$) are marked \emph{floor},
reported without inferential claims, and excluded from inference;
$\dagger$ marks a baseline marginally below the threshold, reported for
completeness. Every interpretable drop has a paired item-level bootstrap
95\% CI ($10{,}000$ resamples, fixed seed) and an exact two-sided
McNemar test.

\begin{table}[h]
\centering
\scriptsize
\setlength{\tabcolsep}{3.6pt}
\renewcommand{\arraystretch}{1.1}
\caption{\textbf{Mental-simulation battery under the frozen VL-axis
intervention.} Accuracy is baseline$\to$intervened (\%); $\Delta$ in pp
with paired bootstrap 95\% CI and exact McNemar $p$. Floor cells
(baseline $<35\%$) are excluded from inference.}
\label{tab:mental-sim}
\begin{tabular}{llccccl}
\toprule
\textbf{Model} & \textbf{Task} & $\boldsymbol{n}$ & \textbf{Base$\to$Int.} & $\boldsymbol{\Delta}$ \textbf{(pp)} & \textbf{95\% CI} & \textbf{McNemar $p$} \\
\midrule
Qwen3-VL-32B & Mental rotation (easy) & 100 & $78.0\to23.0$ & $-55.0$ & $[-66,-44]$ & $1.4\mathrm{e}{-13}$ \\
Qwen3-VL-32B & Mental rotation (hard) & 100 & $61.0\to15.0$ & $-46.0$ & $[-57,-34]$ & $1.2\mathrm{e}{-10}$ \\
Qwen3-VL-32B & Paper folding (easy) & 100 & $89.0\to50.0$ & $-39.0$ & $[-49,-29]$ & $2.2\mathrm{e}{-10}$ \\
Qwen3-VL-32B & Paper folding (hard) & 100 & $36.0\to31.0$ & $-5.0$ & $[-17,+7]$ & $0.52$ (n.s.) \\
Qwen3-VL-32B & Perspective taking (easy) & 100 & $54.0\to32.0$ & $-22.0$ & $[-31,-13]$ & $2.7\mathrm{e}{-5}$ \\
Qwen3-VL-32B & Perspective taking (hard) & 100 & $62.0\to36.0$ & $-26.0$ & $[-36,-16]$ & $6.2\mathrm{e}{-6}$ \\
Qwen3-VL-32B & SPACE MRT (verbatim) & 100 & $43.0\to26.0$ & $-17.0$ & $[-31,-3]$ & $2.4\mathrm{e}{-2}$ \\
Qwen3-VL-32B & SPACE PTT (verbatim) & 100 & $59.0\to33.0$ & $-26.0$ & $[-39,-13]$ & $3.1\mathrm{e}{-4}$ \\
Qwen3-VL-32B & SPACE MPFB (verbatim) & 100 & floor & --- & --- & excluded \\
\midrule
Qwen3-VL-8B & Mental rotation (easy)$^\dagger$ & 150 & $32.0\to19.3$ & $-12.7$ & $[-20.7,-4.7]$ & $5.4\mathrm{e}{-3}$ \\
Qwen3-VL-8B & Mental rotation (hard) & 150 & floor & --- & --- & excluded \\
Qwen3-VL-8B & Paper folding (easy) & 150 & $79.3\to30.7$ & $-48.7$ & $[-58.0,-39.3]$ & $8.3\mathrm{e}{-17}$ \\
Qwen3-VL-8B & Paper folding (hard) & 150 & floor & --- & --- & excluded \\
Qwen3-VL-8B & Perspective taking (easy) & 150 & $59.3\to27.3$ & $-32.0$ & $[-41.3,-22.7]$ & $1.2\mathrm{e}{-9}$ \\
Qwen3-VL-8B & Perspective taking (hard) & 150 & $68.7\to26.0$ & $-42.7$ & $[-52.0,-33.3]$ & $6.3\mathrm{e}{-15}$ \\
Qwen3-VL-8B & SPACE MRT (verbatim) & 150 & floor & --- & --- & excluded \\
Qwen3-VL-8B & SPACE PTT (verbatim) & 150 & $49.3\to27.3$ & $-22.0$ & $[-31.3,-12.0]$ & $3.8\mathrm{e}{-5}$ \\
Qwen3-VL-8B & SPACE MPFB (verbatim) & 150 & floor & --- & --- & excluded \\
\bottomrule
\end{tabular}
\end{table}

\paragraph{Direction specificity.} Five norm-matched random directions
leave perspective taking within $|\Delta|\leq4$ pp (real axis: $-22$ to
$-43$ pp across the two models) and give a mean of $-7.2$ pp (range
$-18$ to $+6$) on 32B mental rotation, where the real axis causes
$-55.0$ pp. The mental-simulation deficits are therefore axis-specific.

\paragraph{Deficit, not disruption.} Perspective taking provides the
cleanest reading: under both conditions, both models produce complete,
well-formed answers to well-posed questions (answer completion
$\approx100\%\to99\%$ under ablation), so the deficit consists of
\emph{incorrect answers}, not a failure to answer.

\paragraph{Answer-completion caveat for the other cells.} Completion is
not fully preserved everywhere: at 32B, mental rotation's completion
falls from $77\%$ to $40\%$ and paper folding's from $89\%$ to $50\%$
under ablation, so part of those accuracy drops reflects failure to
produce a parseable answer within the generation budget. We therefore
treat perspective taking (and the matched-depth battery of
\S\ref{app:matched-depth}) as the anchor evidence for the
mental-simulation claim, with the remaining cells as convergent but
completion-confounded support.

\paragraph{Capability floor and architecture coverage.} Chameleon-7B is
at or below chance on every mental-simulation task, including simplified
variants; its outputs are complete and well formed within the generation
budget but incorrect, indicating a genuine capability floor rather than a
formatting failure. The measurable mental-simulation evidence therefore
comes from the Qwen3-VL configurations that clear the interpretability
threshold, while the canonical VL-axis phenomenon
(Table~\ref{tab:main-effect}) spans architectures.

\subsection{The visual$\times$spatial 2$\times$2 battery}
\label{app:battery-2x2}

To separate ``visual'' from ``spatial,'' we cross \{visual, non-visual\}
with \{spatial, non-spatial\} while matching the 4-way multiple-choice
format, prompt template, scoring, and frozen intervention. The
visual-spatial and non-visual-spatial quadrants are covered by the
mental-simulation battery above and by \texttt{navigate}, SpartQA, and
StepGame (\S\ref{sec:cross-arch-dd}--\ref{sec:four-checks}). The critical
new cell is \emph{visual non-spatial} attribute binding: recalling
colors and materials from a scene description with no spatial relations.
Non-visual non-spatial controls (\texttt{web\_of\_lies},
\texttt{boolean\_expressions}, MMLU-Pro, GSM8K) complete the design.

\begin{table}[h]
\centering
\footnotesize
\setlength{\tabcolsep}{4.5pt}
\renewcommand{\arraystretch}{1.1}
\caption{\textbf{Visual non-spatial attribute binding under the frozen
intervention.} At 7--8B the visual non-spatial control is largely
preserved (Chameleon-7B even improves; CI includes 0), while spatial
tasks drop by 22--49 pp; at 32B the same cell collapses, indicating a
broader functional range at larger scale.}
\label{tab:attribute-binding}
\begin{tabular}{llcccl}
\toprule
\textbf{Model} & $\boldsymbol{n}$ & \textbf{Base$\to$Int.} & $\boldsymbol{\Delta}$ \textbf{(pp)} & \textbf{95\% CI} & \textbf{McNemar $p$} \\
\midrule
Chameleon-7B & 250 & $38.0\to44.0$ & $+6.0$ & $[-0.4,+12.4]$ & $0.091$ (n.s.) \\
Qwen3-VL-8B & 150 & $100.0\to96.0$ & $-4.0$ & $[-7.3,-1.3]$ & $3.1\mathrm{e}{-2}$ \\
Qwen3-VL-32B & 100 & $100.0\to29.0$ & $-71.0$ & $[-80,-62]$ & $8.5\mathrm{e}{-22}$ \\
\bottomrule
\end{tabular}
\end{table}

At 7--8B, the intervention strongly disrupts spatial reasoning while
largely preserving non-spatial visual-attribute recall: Chameleon-7B
changes by $+6.0$ pp (CI includes 0; all five norm-matched random
directions give $|\Delta|\leq2.8$ pp), and Qwen3-VL-8B changes by only
$-4.0$ pp from ceiling versus 22--49 pp spatial drops. Chameleon-7B
starts at only 38\%, so preservation of this control is not explained by
an easy baseline.

\paragraph{The 32B cell and its robustness.} At 32B, attribute binding
falls $100.0\to29.0$ while \texttt{web\_of\_lies} remains at
$58.0\to56.0$ (within the random-direction band) and
\texttt{boolean\_expressions} moves mildly ($88.0\to77.0$);
\texttt{navigate} moves moderately ($70.7\to57.3$). The attribute-binding
drop is axis-specific and robust: a single-layer (hub L8) intervention
gives $100.0\to9.0$; $\alpha{=}0.5$ gives $100.0\to20.0$; and a 3-shot
short-answer format gives $100.0\to32.0$ (with \texttt{boolean\_expressions} at
$88.0\to77.0$ under the same 3-shot configuration); three matched-norm
random directions all remain at $100.0$. Although both attribute-binding
cells start at ceiling, their intervention sensitivity differs sharply
across scale ($100.0\to96.0$ at 8B versus $100.0\to29.0$ at 32B).

We read this as a \emph{configuration-dependent broadening of the
functional range}: the carrier is spatially concentrated within the
visual domain at 7--8B and extends to non-spatial visual attribute
binding at 32B, while remaining selective relative to symbolic controls.
This broadening is consistent with the widening of the affected layer
window at scale ($\approx4$ to $\approx11$ layers;
Figure~\ref{fig:panel-qwen-scaling}) and is why we state the construct
(\emph{vision-shaped}) independently of any single task family.

\subsection{Matched-depth, answer-completion-controlled dissociation}
\label{app:matched-depth}

Spatial tasks might degrade more simply because they require more
sequential state updates. We therefore cross spatial demand with update
load on Qwen3-VL-8B, using a non-spatial color-tracking task (tracking
which color each named object currently has through a sequence of
repaint updates) with matched format and frozen intervention.

\begin{table}[h]
\centering
\footnotesize
\setlength{\tabcolsep}{5pt}
\renewcommand{\arraystretch}{1.1}
\caption{\textbf{Crossing spatial demand with sequential update load
(Qwen3-VL-8B).} Completion denotes the fraction of items with a final
answer produced within the generation budget (baseline$\to$intervened
where both measured).}
\label{tab:matched-depth}
\begin{tabular}{llccc}
\toprule
\textbf{Task} & \textbf{Update load} & \textbf{Base$\to$Int.} & $\boldsymbol{\Delta}$ \textbf{(pp)} & \textbf{Completion} \\
\midrule
Perspective taking (spatial; easy/hard) & 0 & $59.3\to27.3$ / $68.7\to26.0$ & $-32.0$ / $-42.7$ & $\geq0.94$ \\
Color tracking (non-spatial) & 3 & $100.0\to79.3$ & $-20.7$ & $0.98$ \\
Color tracking (non-spatial) & 5 & $100.0\to71.3$ & $-28.7$ & $0.94$ \\
Color tracking (non-spatial) & 8 & $99.3\to55.3$ & $-44.0$ & $0.83$ \\
Color tracking (non-spatial) & 12 & $96.0\to44.0$ & $-52.0$ & $0.81$ \\
\bottomrule
\end{tabular}
\end{table}

Three observations separate the two factors. First, mental rotation and
perspective taking each present one static scene and one question, with
\emph{zero} update load, yet show large drops ($78.0\to23.0$ at 32B;
$59.3\to27.3$ at 8B), so sequential update load cannot be the primary
explanation. Second, at matched moderate depth (3--5 updates) the
non-spatial cells are less impaired ($100\to79/71$) than the zero-update
spatial cells, and items are answered within the generation budget on
$\geq94\%$ of trials in these cells, so the spatial deficit is not a
failure to produce answers. The non-spatial effect is also axis-specific:
five random directions leave color tracking at $100\%$. Third, at longer
horizons (8--12 updates) the non-spatial drop grows to $-44/-52$ pp,
revealing partial overlap between the same carrier and long-horizon
non-spatial state tracking. Together, these results identify a spatially
concentrated carrier with a secondary role in sufficiently long
non-spatial state tracking.

\subsection{Statistical control for baseline accuracy}
\label{app:spectrum-control}

Because tasks differ in baseline difficulty, we regress each
configuration's per-task intervention effect on the 0--5 spatial-imagery
score while controlling for baseline accuracy across the seven-task
spectrum (\S\ref{sec:four-checks}).

\begin{table}[h]
\centering
\footnotesize
\setlength{\tabcolsep}{5pt}
\renewcommand{\arraystretch}{1.05}
\caption{\textbf{Baseline-controlled spatial-score slopes} (pp per score
point) across the seven-task spectrum. The slope remains negative in 6
of 7 configurations after controlling for baseline accuracy. We report
these descriptively and make no pooled-significance claim.}
\label{tab:spectrum-control}
\begin{tabular}{lcl}
\toprule
\textbf{Configuration} & \textbf{Baseline-controlled slope} & \textbf{Note} \\
\midrule
Qwen3-VL-32B & $-7.28$ & \\
Chameleon-7B & $-5.70$ & Spearman $\rho=-0.83$, $p=0.042$ \\
Emu3 & $-5.25$ & \\
Janus-Pro & $-3.79$ & \\
Liquid & $-3.59$ & \\
BAGEL & $-1.60$ & \\
Qwen3-VL-8B & $+1.93$ & sole positive slope \\
\bottomrule
\end{tabular}
\end{table}

\subsection{Task selectivity across all 27 BBH tasks}
\label{app:cham7b-bbh27}

Would the double dissociation survive a broader set of algorithmic and
structured-reasoning controls than \texttt{boolean\_expressions} and
\texttt{web\_of\_lies}? We evaluated Chameleon-7B on all 27 BBH tasks
under one frozen intervention (embedding VL axis, L0--7, $\alpha{=}1.0$,
$n{=}150$ per task), with no per-task retuning.
Table~\ref{tab:cham7b-bbh27} reports the full results; Janus-Pro's
independent 27-task run under its own frozen generation-path
configuration is in Table~\ref{tab:janus-bbh27}.

\begin{table}[h]
\centering
\scriptsize
\setlength{\tabcolsep}{4pt}
\renewcommand{\arraystretch}{1.05}
\caption{\textbf{Chameleon-7B on all 27 BBH tasks under the frozen L0--7
intervention} ($\alpha{=}1.0$, $n{=}150$). The two largest drops are the
two most spatial tasks; algorithmic and structured-logic controls are
stable or improve. Tasks with near-floor baselines should be read
qualitatively.}
\label{tab:cham7b-bbh27}
\begin{tabular}{lccr@{\hspace{18pt}}lccr}
\toprule
\textbf{Task} & \textbf{Base} & \textbf{Int.} & $\boldsymbol{\Delta}$ &
\textbf{Task} & \textbf{Base} & \textbf{Int.} & $\boldsymbol{\Delta}$ \\
\midrule
\texttt{navigate} & 43.3 & 22.7 & $-20.7$ & \texttt{date\_understanding} & 22.0 & 26.7 & $+4.7$ \\
\texttt{tracking\_shuffled\_3\_obj} & 29.3 & 18.7 & $-10.7$ & \texttt{boolean\_expressions} & 50.0 & 54.7 & $+4.7$ \\
\texttt{movie\_recommendation} & 42.7 & 36.7 & $-6.0$ & \texttt{ruin\_names} & 24.0 & 28.0 & $+4.0$ \\
\texttt{snarks} & 47.3 & 41.3 & $-6.0$ & \texttt{logical\_deduction\_7\_obj} & 11.3 & 15.3 & $+4.0$ \\
\texttt{tracking\_shuffled\_5\_obj} & 22.7 & 18.0 & $-4.7$ & \texttt{formal\_fallacies} & 84.0 & 86.7 & $+2.7$ \\
\texttt{salient\_translation\_error} & 16.0 & 12.0 & $-4.0$ & \texttt{reasoning\_colored\_objects} & 15.3 & 18.0 & $+2.7$ \\
\texttt{logical\_deduction\_3\_obj} & 7.3 & 4.0 & $-3.3$ & \texttt{geometric\_shapes} & 2.0 & 4.0 & $+2.0$ \\
\texttt{sports\_understanding} & 38.7 & 36.0 & $-2.7$ & \texttt{temporal\_sequences} & 8.0 & 10.0 & $+2.0$ \\
\texttt{disambiguation\_qa} & 24.0 & 22.0 & $-2.0$ & \texttt{penguins\_in\_a\_table} & 30.8 & 32.2 & $+1.4$ \\
\texttt{hyperbaton} & 43.3 & 42.7 & $-0.7$ & \texttt{tracking\_shuffled\_7\_obj} & 13.3 & 14.7 & $+1.3$ \\
\texttt{multistep\_arithmetic\_two} & 4.7 & 4.0 & $-0.7$ & \texttt{dyck\_languages} & 20.0 & 20.7 & $+0.7$ \\
\texttt{object\_counting} & 2.0 & 1.3 & $-0.7$ & \texttt{word\_sorting} & 0.0 & 0.0 & $0.0$ \\
\texttt{causal\_judgement} & 20.0 & 28.0 & $+8.0$ & \texttt{logical\_deduction\_5\_obj} & 10.0 & 16.7 & $+6.7$ \\
\texttt{web\_of\_lies} & 21.3 & 46.0 & $+24.7$ & & & & \\
\bottomrule
\end{tabular}
\end{table}

Under the frozen intervention, the two largest drops are
\texttt{navigate} ($-20.7$ pp) and the 3-object tracking variant
($-10.7$ pp), while algorithmic and structured controls---
\texttt{dyck\_languages} ($+0.7$), \texttt{formal\_fallacies} ($+2.7$ at
an 84\% baseline), \texttt{hyperbaton} ($-0.7$), \texttt{word\_sorting}
($0.0$), and \texttt{logical\_deduction} ($-3.3$/$+6.7$/$+4.0$ for
3/5/7 objects)---remain stable or improve; \texttt{web\_of\_lies}
improves by $+24.7$ pp. Selectivity therefore does not depend on the
particular \texttt{navigate}/\texttt{boolean} pair. On Janus-Pro's
independent 27-task run, the spatial-group mean drops $-14.5$ pp,
roughly three times the symbolic-group mean ($-4.6$ pp), replicating the
group-level contrast on a second architecture
(Appendix~\ref{app:janus-bbh27}). Two transparency notes bound this
table. First, several of the 27 tasks
have near-floor baselines for this model (e.g., \texttt{word\_sorting},
\texttt{object\_counting}, \texttt{geometric\_shapes}), so their
individual deltas are not individually interpretable, and we do not
claim a graded spatial-score trend over the full 27-task suite. Second,
for binary yes/no tasks whose Chameleon-7B baselines sit near or below
chance (\texttt{navigate}, \texttt{boolean\_expressions},
\texttt{web\_of\_lies}, \texttt{causal\_judgement}), accuracy changes
partly reflect shifts in the output-answer distribution rather than pure
reasoning changes---\texttt{web\_of\_lies}' large $+24.7$ pp
``improvement,'' for example, coincides with a high no-answer rate at
baseline. We therefore read this table as a frozen-intervention
selectivity check, and rest the strongest causal claims on the
answer-completion-controlled batteries of
\S\ref{app:mental-sim}--\ref{app:matched-depth}.

\section{When EVT Does Not Appear: Failure Modes and Diagnostic Lessons}
\label{app:failures}

EVT is not a uniform phenomenon. Across 10 multimodal configurations and 4 cross-family transfer experiments, we observed several modes in which the canonical $1$D VL-axis intervention fails, succeeds only weakly, or requires a non-modality contrast to expose. This appendix consolidates all such cases. We organize them by the diagnostic lesson rather than by model family, because the same model can fail under one method and succeed under another (Chameleon-30B is the clearest example). Each subsection states the observation, the apparent cause, and the methodological correction --- if one was found.

\paragraph{Roadmap.}
\S\ref{app:fail-cham30b} documents Chameleon-30B, where seven distinct modality-axis methods all return null and a task-contrastive axis recovers a strong effect, demonstrating that the EVT signal can dissociate from the modality axis at scale. \S\ref{app:fail-janus-early} documents two early failed attempts on Janus-Pro before the generation-path DiffMean axis was identified. \S\ref{app:fail-liquid-cliff} documents the dosage cliff on Liquid. \S\ref{app:fail-qwen2b} contextualizes the smaller-scale Qwen3-VL-2B effect within the cross-scale narrative. \S\ref{app:fail-transfer} documents four cross-family backbone-transfer experiments, three of which produce null or non-selective damage and one of which produces a positive transfer that constrains the attribution argument. Together, these cases sharpen the boundary of the canonical method and motivate the diagnostic protocol of the axis-construction details earlier in this appendix.

\subsection{Chameleon-30B: A modality-axis null with a task-contrastive rescue}
\label{app:fail-cham30b}

Chameleon-30B shares architecture and training data with Chameleon-7B; the only difference is capacity ($d_{\text{hid}}{=}8192$ vs.\ $4096$, $48$ vs.\ $32$ layers). The 7B configuration is the cleanest positive case in the paper (single-layer L5 DD $=+33.2$ pp; block-level $z=+12.22$). On the 30B version, seven modality-axis methods produced null effects --- all seven, however, construct the modality contrast from the embedding table or from synthetic token-ID inputs; an eighth, input-faithful modality-axis construction (full multimodal sequences through the model's own VQ tokenizer) later succeeds, resolving the boundary case (Appendix~\ref{app:cham-fullseq}, Table~\ref{tab:cham-fullseq}, and the resolution paragraph at the end of this section).

\paragraph{The seven failed modality-axis methods.}
Table~\ref{tab:cham30b-failures} enumerates each method, its expected effect (extrapolated from the 7B result), and the observed null.

\begin{table}[h]
\centering
\footnotesize
\setlength{\tabcolsep}{4pt}
\renewcommand{\arraystretch}{1.05}
\caption{\textbf{Seven modality-axis methods on Chameleon-30B, all null.} Each row reports a distinct intervention; all yielded $\mathrm{DD}\approx 0$ on \texttt{navigate}/\texttt{boolean} pair. The eighth row (task-contrastive axis, last row, separated) was the first construction to succeed. A ninth construction --- the input-faithful full-sequence \emph{modality} axis --- also succeeds (mid-block DD $=+33.2$ pp; Table~\ref{tab:cham-fullseq}) and supersedes the ``modality axis is inert'' reading of rows 1--7; see the resolution paragraph. Diagnostic conclusion in the rightmost column.}
\label{tab:cham30b-failures}
\begin{tabularx}{\linewidth}{@{} l >{\raggedright\arraybackslash}p{0.34\linewidth} >{\raggedright\arraybackslash}p{0.16\linewidth} >{\raggedright\arraybackslash}X @{}}
\toprule
\textbf{\#} & \textbf{Method} & \textbf{Effect} & \textbf{Apparent cause} \\
\midrule
1 & Embedding axis, $\alpha{=}1.0$, 7 layer configs & DD $\approx 0$ & Axis exists geometrically (separation $0.1065 > $ 7B's $0.0851$) but is causally inert \\
2 & Alpha sweep $\alpha\in[1.5, 2.2]$ & DD $\approx 0$ & Stronger removal still inert \\
3 & Alpha sweep $\alpha \geq 2.5$ & non-selective collapse & $\alpha$ exceeds residual-norm budget \\
4 & PCA subspace $k\in\{1,\ldots,200\}$, two depth ranges & DD $\approx 0$ & Position-PCA is not task-causal-PCA; 2.4\% variance subspace contains no causal direction \\
5 & Single-task probe direction & non-selective & Probe fits prompt-template distribution, not spatial content \\
6 & Multi-task LOTO probe (12 layers tested) & 0\% across all layers & No linearly shared spatial-reasoning feature across tasks \\
7 & Per-layer DiffMean, 4 blocks $\times$ 4 $\alpha$ values plus a 7-dose early-block fine sweep & DD $\approx 0$ on all 19 cells & $\cos$(DiffMean axis, embedding axis)$\approx 0$; modality-position axis $\perp$ task-causal direction \\
\midrule
\textbf{8} & \textbf{Task-contrastive axis} (\texttt{nav} $-$ \texttt{bool}, L24--35) & \textbf{DD $=+16.7$ pp}, $z=+6.4$ & Task-contrast direction is causal; modality-position direction is not (\S\ref{sec:representation-properties}, semantic location)  \\
\bottomrule
\end{tabularx}
\end{table}

\paragraph{Held-out validation of the task-contrastive axis.}
The task-contrastive axis in Table~\ref{tab:cham30b-failures} is built
from the \texttt{navigate}--\texttt{boolean\_expressions} contrast. To
rule out circularity, we evaluate the same axis on held-out tasks that
were not used for axis construction. The held-out spatial tasks are
\texttt{relative\_position},
\texttt{reasoning\_about\_colored\_objects}, and
\texttt{temporal\_sequences}; the held-out control tasks are
\texttt{web\_of\_lies} and \texttt{causal\_judgement}. The intervention
uses the late block L24--35 with $\alpha=1.0$.

\begin{table}[h]
\centering
\footnotesize
\setlength{\tabcolsep}{5pt}
\renewcommand{\arraystretch}{1.05}
\caption{\textbf{Held-out validation of the Chameleon-30B task-contrastive
axis.} The axis is constructed from \texttt{navigate} and
\texttt{boolean\_expressions}, but evaluated on held-out tasks.}
\label{tab:cham30b-heldout}
\begin{tabular}{l l c c c}
\toprule
\textbf{Task} & \textbf{Group} & \textbf{Baseline} &
\textbf{Intervened} & $\boldsymbol{\Delta}$ \\
\midrule
\texttt{relative\_position} & spatial & $62.8$ & $62.8$ & $\phantom{+}0.0$ pp \\
\texttt{reasoning\_about\_colored\_objects} & spatial & $27.9$ & $26.7$ & $-1.2$ pp \\
\texttt{temporal\_sequences} & spatial & $28.4$ & $28.8$ & $+0.4$ pp \\
\texttt{web\_of\_lies} & control & $48.8$ & $51.2$ & $+2.4$ pp \\
\texttt{causal\_judgement} & control & $58.3$ & $67.9$ & $+9.6$ pp \\
\bottomrule
\end{tabular}
\end{table}

The held-out DD remains positive at $+6.28$ pp, with paired-bootstrap
CI $[+0.13,+12.37]$ pp. Energy-matched random directions give
$-1.08 \pm 3.28$ pp, yielding $z=+2.24$ against the random baseline
(one-sided normal-approximation $p\approx0.013$). Thus, the task-contrastive axis retains a
directional double-dissociation signal beyond the task pair used to
construct it.

\paragraph{Geometry of modality and task axes.}
We further compare the canonical VL axis with the task-contrastive axis
across layers. The VL axis is the embedding-level text--visual centroid
difference. The task axis at layer $L$ is defined as the difference
between the mean final-token hidden states of \texttt{navigate} and
\texttt{boolean\_expressions} examples.

\begin{table}[h]
\centering
\footnotesize
\setlength{\tabcolsep}{5pt}
\renewcommand{\arraystretch}{1.05}
\caption{\textbf{VL axis and task axis are nearly orthogonal in
Chameleon.} The two axes have small cosine similarity across the network,
supporting the interpretation that Chameleon-30B decouples modality
separation from task-causal spatial reasoning.}
\label{tab:cham-axis-cos}
\begin{tabular}{l c c c}
\toprule
\textbf{Model} & \textbf{Layers} & \textbf{Mean $|\cos|$} &
\textbf{Max $|\cos|$} \\
\midrule
Chameleon-7B & $32$ & $0.0118$ & $0.0390$ at L9 \\
Chameleon-30B & $48$ & $0.0145$ & $0.0360$ at L47 \\
\bottomrule
\end{tabular}
\end{table}

In both models, $|\cos|<0.04$ throughout the network, showing that the VL
axis and the task-contrastive axis are almost orthogonal. The important
difference is semantic location. In Chameleon-7B, the small cosine peak
overlaps the early VL-axis causal hub, whereas in Chameleon-30B the
task-axis rescue hub lies in L24--35 and is separated from the VL-axis
geometric peak. (Note that the VL axis in this comparison is the
embedding-centroid estimate; the input-faithful full-sequence modality
axis, which \emph{is} causal at 30B, is itself near-orthogonal to the
embedding axis ($\cos\approx 0.01$) --- see the resolution paragraph
below and Appendix~\ref{app:cham-fullseq}.)

The full data for methods 1--7 ships with the released per-sample result files (\S\ref{app:repro-release}); the rescue evidence is summarized in Table~\ref{tab:cham30b-failures}
and validated on held-out tasks in Table~\ref{tab:cham30b-heldout}.
\paragraph{Original diagnostic conclusion: capacity-driven decoupling (superseded; retained for the methodological record).}
Methods 1--7 all parameterize the intervention along a \emph{modality} contrast (text vs.\ visual). Method 8 parameterizes it along a \emph{task} contrast (\texttt{navigate} vs.\ \texttt{boolean\_expressions}). The fact that the same hidden state, at the same layer block, with the same $\alpha$, supports a $z=+6.4$ effect under method 8 but DD $\approx 0$ under method 7 originally suggested that 30B retains spatial-reasoning structure while \emph{decoupling} it from the modality direction, with the doubled hidden dimension (8192 vs.\ 4096) relieving the dimensional pressure that, in the 7B model, forces modality and spatial-reasoning circuits to share representational space.

\paragraph{Resolution: the modality axis is causal under an input-faithful construction.}
The decoupling interpretation is superseded by the full-sequence
source-swap experiment (Appendix~\ref{app:cham-fullseq}). Methods 1--7
share a hidden confound: none of them measures the modality contrast on
the model's actual multimodal input pathway. Method 7's synthetic axis
(16 random token IDs, no image, no VQ encoding) turns out to be rotated
$\cos \approx 0.43$--$0.56$ away from the axis obtained when real or
random images are passed through Chameleon's own VQ tokenizer inside a
full \texttt{<image>}+text sequence. Along that input-faithful modality
axis, Chameleon-30B exhibits the full causal signature: mid-block
(L12--23) DD $=+33.2$ pp with \texttt{boolean} unharmed
(CI $[+21.2,+44.8]$; \texttt{navigate} $z=-88.7$ against random
directions, $0/3$), final-block (L36--47) DD $=+22.0$/$+22.4$ pp under
COCO/random-pixel sources respectively (random directions produce
exactly zero \texttt{navigate} change), and a clean null at the late
block (L24--35). Three revisions follow. (i) The spatial carrier at 30B
has \emph{not} decoupled from the modality contrast; relative to the 7B
hub (L5, early), it has \emph{relocated} to mid/final blocks. (ii) What
scaling changes is the carrier's depth and its sensitivity to axis
construction, not its existence --- so the cross-family contrast with
Qwen3-VL (\S\ref{sec:evt}) is about \emph{where} the carrier lives, not
\emph{whether} it survives scale. (iii) The task-contrastive rescue
(method 8) and its held-out validation remain valid evidence, but they
are no longer the only handle on 30B; the semantic-location lesson of
\S\ref{sec:representation-properties} sharpens into a construction
lesson: a null along one \emph{estimate} of the modality contrast does
not establish that the modality direction itself is inert.

\subsection{Janus-Pro-7B: Two failed early stages before generation-path success}
\label{app:fail-janus-early}

Janus-Pro was initially classified as a failure case under two methods before the generation-path DiffMean axis was tried. Both failures were architecture-specific and instructive.

\paragraph{Stage 1 --- Embedding centroid axis.}
We applied the Chameleon recipe directly: $v = \mathrm{normalize}(\mu_{\text{text}} - \mu_{\text{visual}})$ from the embedding table, with $\alpha{=}1.0$ on five layer configurations. All cells produced $|z| < 1$. Diagnostic: in Janus-Pro, the \texttt{<image\_N>} placeholder tokens used to define the visual centroid are \emph{functional dead weights} --- during real inference, the SigLIP aligner overwrites those positions, so the embedding rows are never updated by training. The centroid difference therefore reflects initialization noise, not a learned visual direction. This is a generic risk for any architecture where visual tokens enter the residual stream by replacement rather than by table lookup.

\paragraph{Stage 2 --- SigLIP understanding-path DiffMean.}
Following BAGEL precedent, we collected hidden states at \texttt{<image\_N>} positions from real COCO images passed through the SigLIP aligner, then computed per-layer DiffMean. Axis diagnostic immediately flagged a problem: $\|\mu_{\text{text},L}\| \approx 14$ while $\|\mu_{\text{visual},L}\| \approx 114$ at $L{=}0$, with projection ratio $0.83$. The axis captured an input-source norm gap, not a clean modality-semantic direction. Applying $\alpha{=}0.5$ at L0--7 collapsed the model: \texttt{navigate} acc spiked to $0.984$ (a false-positive saturation pattern; the model emits \texttt{"Yes"} unconditionally), \texttt{boolean} dropped to $0.000$ with degenerate output (e.g., \texttt{"Yes-NoLet'sthinkstepbystep.A:Yes-NoLet'sthinkstepbystep.A:..."}~repeating). The intervention destroyed token prediction rather than spatial reasoning specifically.

\paragraph{Stage 3 --- generation-path DiffMean (success).}
The third method --- per-layer DiffMean computed on hidden states at VQ-16 generation tokens --- passed all four causal checks (DD $=+8.0$ pp at $\alpha{=}0.7$, $z=+3.58$, $0/22$ random hits, paired bootstrap CI $[+2.93, +13.00]$ pp). This stage is documented in the main text \S\ref{sec:generation-carrier}; the failures of stages 1 and 2 are recorded here both as a methodological warning and as evidence that the generation-path axis is not the only construction tried but the only one that survives diagnostic checks.

\subsection{Liquid-V1-7B: A narrow dosage cliff at block level}
\label{app:fail-liquid-cliff}

Liquid passes all four causal checks in the single-layer configuration:
the headline result is L27-only removal at $\alpha{=}1.0$
(DD $=+21.6$ pp; \texttt{navigate} $-12.8$, \texttt{boolean} $+8.8$;
Table~\ref{tab:main-effect}). The dosage cliff appears in the
\emph{block-level} experiment (L21--27), whose full sweep is shown in
Table~\ref{tab:liquid-cliff}.

\begin{table}[h]
\centering
\footnotesize
\setlength{\tabcolsep}{5pt}
\renewcommand{\arraystretch}{1.05}
\caption{\textbf{Liquid block-level dosage sweep (L21--27).} The
transition from a weak effect to \texttt{navigate} collapse occurs
inside the $\alpha\in[0.50,0.55]$ window. Large post-cliff DDs are
saturation artifacts: \texttt{navigate} accuracy is at or near zero, so
no usable selectivity remains. Accuracy in \%, $\Delta$ relative to the
\texttt{navigate}/\texttt{boolean} baselines (54.8/54.8).}
\label{tab:liquid-cliff}
\begin{tabular}{cccccl}
\toprule
$\alpha$ & navigate & boolean & $\Delta$nav (pp) & DD (pp) & status \\
\midrule
0.40 & 51.2 & 51.6 & $-3.6$ & $+0.4$ & weak \\
0.50 & 44.8 & 50.4 & $-10.0$ & $+5.6$ & block-level sweet spot \\
0.55 & 8.8 & 48.8 & $-46.0$ & $+40.0$ & cliff: navigate collapses \\
0.60 & 0.0 & 49.2 & $-54.8$ & $+49.2$ & navigate at zero \\
0.65 & 0.0 & 48.4 & $-54.8$ & $+48.4$ & plateau (nav dead, bool stable) \\
0.70 & 0.0 & 34.4 & $-54.8$ & $+34.4$ & boolean begins to collapse \\
1.00 & 0.4 & 0.0 & $-54.4$ & $\approx 0$ & full non-selective collapse \\
\bottomrule
\end{tabular}
\end{table}

The cliff is two-staged: \texttt{navigate} collapses between
$\alpha{=}0.50$ and $0.55$, then \texttt{boolean\_expressions} collapses
around $\alpha{=}0.70$. We emphasize that the post-cliff region is not a
usable selective-intervention window: with \texttt{navigate} at 0--9\%
accuracy, the large DD values reflect one task being dead rather than
graded selectivity. The best block-level selectivity is only
DD $=+5.6$ pp at $\alpha{=}0.50$, far below the single-layer L27
configuration ($+21.6$ pp), which concentrates the removal at the hub
and avoids the cliff entirely.

\paragraph{Apparent cause.}
Liquid was created by VQGAN fine-tuning on top of a Gemma-7B text backbone. Visual information is injected only at the final layer block, with no joint training of earlier layers. This produces a maximally localized causal hub: spatial information sits in a thin sliver of representational space at L27, and the block-level effect appears or disappears together as $\alpha$ crosses a threshold. This is the geometric opposite of Emu3, where the same causal pathway is duplicated at early and final blocks
(\S\ref{app:emu3-dual-carrier}). The single-layer headline configuration does not exhibit the cliff at the synthetic-axis dosage ($\alpha{=}1.0$); the real-image axis, which projects more strongly onto L27 hidden states, shifts the same cliff to lower $\alpha$ (\S\ref{app:source-swap}). The block-level cliff is a methodological warning for block-scale interventions on late-fusion architectures.

\subsection{Qwen3-VL-2B: A weaker but coherent positive case}
\label{app:fail-qwen2b}

Qwen3-VL-2B passes all four causal checks with $z=+27.7$, showing that
the Qwen3-VL effect is already present at the smallest dense scale we test.
Together with Qwen3-VL-4B/8B ($z=+39.1/+38.3$) and Qwen3-VL-32B ($z=+262.5$),
this supports an overall amplification trend across scale, while keeping
the small-scale result as a clean positive case rather than a mixed
scaling point.

\paragraph{Note on 30B-A3B (MoE).} The MoE configuration achieves $z=+48.8$ (stronger than 2B). Its random-direction control was run at $N=5$ rather than the $N=22$ used for other models, and as the only MoE in our suite ($n{=}1$), conclusions about MoE-specific effects carry the standard sample-size caveat.

\subsection{Cross-family backbone transfer: Three failures and one positive}
\label{app:fail-transfer}

To test whether VL-axis effects are intrinsic to the source VLM or merely reflect a generic property of the underlying backbone, we transferred axes from VLMs to their text-only counterparts and to unrelated text backbones. Four experiments were run; three produced null or non-selective damage, and one produced a positive transfer that constrains the attribution argument.
\begin{table}[t]
\centering
\footnotesize
\setlength{\tabcolsep}{3.8pt}
\renewcommand{\arraystretch}{1.14}
\caption{\textbf{Cross-family backbone transfer.}
Source VLM axes are transferred to target text backbones under the same
intervention recipe. Only the matched Qwen family yields positive transfer.}
\label{tab:transfer}

\newcolumntype{L}[1]{>{\raggedright\arraybackslash}p{#1}}
\newcolumntype{Y}{>{\raggedright\arraybackslash}X}

\begin{tabularx}{\linewidth}{@{}
L{0.25\linewidth}
L{0.27\linewidth}
L{0.13\linewidth}
Y
@{}}
\toprule
\textbf{Transfer}
& \textbf{Effect}
& \textbf{Verdict}
& \textbf{Takeaway} \\
\midrule

Janus $\to$ DeepSeek-LLM-7B
& DD $=+2.4$--$3.6$ pp; $z=+0.32$
& null
& No task-causal coupling in the unrelated text backbone. \\

BAGEL $\to$ Qwen2.5-7B base
& navigate $\Delta=-5.3$ pp; $z=-5.63$; DD $\approx 0$
& non-selective
& Active direction, but no spatial selectivity. \\

Liquid text-mode self-transfer
& navigate $\Delta=+3.6$ pp; $z=+9.04$
& sign-reversed
& Text-mode axis amplifies rather than suppresses spatial reasoning. \\

\midrule

\textbf{Qwen2.5-VL $\to$ Qwen2.5-7B base}
& \textbf{DD $=+18.5$ pp}; $z=-18.6$
& \textbf{positive}
& \textbf{Matched-family transfer supports amplification of a pre-existing Qwen spatial scaffold.} \\

\bottomrule
\end{tabularx}
\end{table}
\paragraph{The positive transfer constrains the attribution claim.}
The Qwen2.5-VL $\to$ Qwen2.5-7B result (last row) is what allows \S\ref{sec:attribution-control} to distinguish \emph{introduction} (Chameleon: 13$\times$ magnitude gap between VLM and matched text backbone) from \emph{amplification} (Qwen family: aligned scaffold pre-exists, multimodal training strengthens it). Without this transfer experiment, the Qwen3-VL $\to$ Qwen3-8B base result alone would be ambiguous because Qwen3 pretraining data is not fully documented; the Qwen2.5 variant has fully published training documentation and shows the same DD$\approx +18$pp pattern, consistent with amplification rather than introduction
(\S\ref{sec:attribution-control}).

\paragraph{The three nulls bound the universality claim.}
Janus $\to$ DeepSeek and BAGEL $\to$ Qwen2.5 both rule out a generic-direction hypothesis: there is no single ``spatial reasoning direction'' portable across pretrained text backbones. An axis must either (i) be extracted from the same family ancestor or (ii) target a feature already present in the backbone. The Liquid self-transfer further shows that even the same-backbone embedding-space axis behaves qualitatively differently in text-only mode --- a methodological warning that axis construction is recipe-specific.

\subsection{Summary: failure modes and the boundary of the canonical method}

Across the cases above, three distinct failure modes account for the observed nulls:

\paragraph{(M1) Wrong axis semantics.}
Cham-30B (modality axis null but task-contrastive succeeds), Janus stage 1 (embedding axis on dead-weight visual tokens), Janus stage 2 (SigLIP path axis dominated by norm gap). In each case the candidate axis exists geometrically but does not align with the causal direction. Diagnostic: report axis projection ratio (proj/h), centroid separation, and split-half cosine before claiming a null.

\paragraph{(M2) Wrong dosage window.}
Liquid cliff. The axis and layer are correct; only a narrow $\alpha$ band exposes the effect. Diagnostic: scan $\alpha$ at fine resolution near any apparent transition before concluding null.

\paragraph{(M3) Wrong target backbone.}
Cross-family transfer (3 of 4 cases). The axis is correct in its source VLM but does not couple causally to spatial reasoning in an unrelated text backbone. Diagnostic: same-family transfer is the strongest control; cross-family transfer measures something different (geometric portability) and is null by default rather than informative.

The Qwen3-VL-2B cases are not failures under any of M1--M3; they are quantitative cases of the canonical method at smaller scale, consistent with the introduction-vs-amplification gradient documented in \S\ref{sec:attribution-control}.

\smallskip
\noindent
The methodological correction implied by these cases is the diagnostic protocol of axis-construction validation: before reporting a null on a new architecture, validate axis construction (axis projection ratio, separation, split-half cosine), scan dosage at the cliff resolution, and, for dual-path or cross-family cases, attempt the task-contrastive axis as a fallback (Cham-30B precedent).

\section{Three Regime Anchor Cases: Extended Evidence}
\label{app:anchors}

\S\ref{sec:representation-properties} introduces three geometric regimes for the alignment between the readout (DiffMean) axis and the dominant variance subspace: \emph{concentric}, \emph{projected-diffuse}, and \emph{mixed}. Each regime is anchored by one configuration in our experiments. This appendix consolidates the underlying data --- per-layer $|\cos(\mathrm{DiffMean}, \mathrm{PCA\text{-}top\text{-}1})|$, cumulative explained variance (EV), and rank-$k$ DD curves --- for the three anchors. All data come from EXP4 (released as \texttt{src/pipelines/subspace\_intervention.py}). One provenance note for auditability: the raw EXP4 logs store a \texttt{dd} field computed as $\Delta_{\text{spatial}}-\Delta_{\text{control}}$, the \emph{opposite} sign of the Methods convention DD $=\Delta_{\text{control}}-\Delta_{\text{spatial}}$; all rank-$k$ DD values in this paper are reported in the Methods convention (sign-corrected from the logs), and the released pipeline documents the discrepancy in its docstring. A full classification of the remaining 7 model configurations into regimes is deferred to future work, since systematic rank-$k$ ablation requires a phase-0 PCA construction that is beyond the scope of this work (\S\ref{app:failures}).

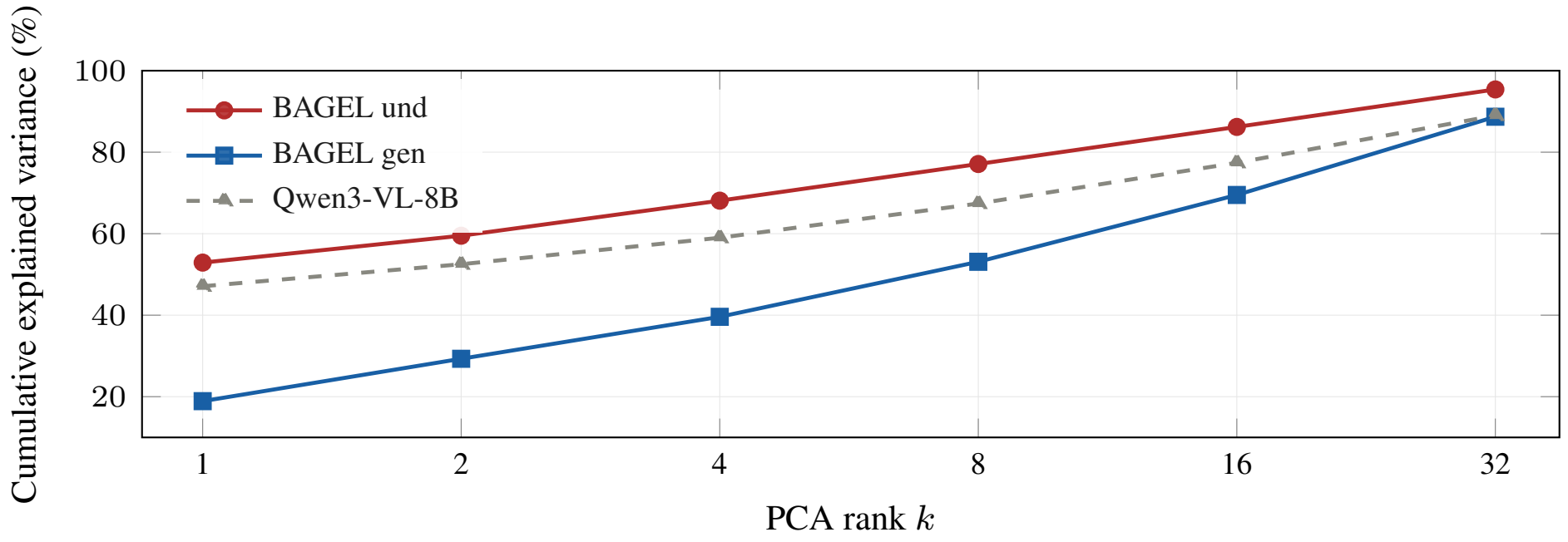
\begin{figure}[h]
\centering
\begin{tikzpicture}
\begin{axis}[
  width=0.92\linewidth, height=4.5cm,
  xlabel={\footnotesize PCA rank $k$},
  ylabel={\footnotesize Cumulative explained variance (\%)},
  xmode=log, log basis x=2,
  xmin=0.85, xmax=38,
  ymin=10, ymax=100,
  xtick={1,2,4,8,16,32}, xticklabels={1,2,4,8,16,32},
  grid=major, grid style={black!10, very thin},
  tick label style={font=\scriptsize},
  label style={font=\footnotesize},
  legend style={font=\scriptsize, draw=none, fill=white, fill opacity=0.9, at={(0.02, 0.98)}, anchor=north west},
  legend cell align=left,
  every axis plot/.append style={mark size=1.6pt, line width=0.9pt}
]
\addplot[color=accent, mark=*] coordinates {
  (1, 52.9) (2, 59.5) (4, 68.1) (8, 77.1) (16, 86.2) (32, 95.4)
};
\addlegendentry{BAGEL und}
\addplot[color=accent2, mark=square*] coordinates {
  (1, 18.9) (2, 29.3) (4, 39.6) (8, 53.1) (16, 69.5) (32, 88.7)
};
\addlegendentry{BAGEL gen}
\addplot[color=neutral, mark=triangle*, dashed] coordinates {
  (1, 47.1) (2, 52.5) (4, 59.0) (8, 67.4) (16, 77.4) (32, 89.0)
};
\addlegendentry{Qwen3-VL-8B}
\end{axis}
\end{tikzpicture}
\caption{\textbf{Cumulative explained variance of the per-sample DiffMean PCA subspace, hub-block average.} BAGEL und reaches $86\%$ by $k{=}16$, indicating a relatively concentrated dominant variance structure. BAGEL gen requires $k{=}32$ to reach the same coverage --- consistent with the projected-diffuse pattern. Qwen3-VL-8B is intermediate.}
\label{fig:anchor-ev}
\end{figure}

\paragraph{Anchor summary.}
Table~\ref{tab:anchor-summary} gives the headline numbers for the three regimes; Figures~\ref{fig:anchor-cos}--\ref{fig:anchor-dd} visualize the underlying curves.

\begin{table}[h]
\centering
\footnotesize
\setlength{\tabcolsep}{4pt}
\renewcommand{\arraystretch}{1.05}
\caption{\textbf{Three regime anchor cases.} Hub-block mean $|\cos(\mathrm{DiffMean}, \mathrm{PCA\text{-}top\text{-}1})|$, cumulative explained variance at $k{=}1$ and $k{=}16$, and the most informative rank-$k$ DD value with $95\%$ CI. ``ctrl crash'' marks DD values that are driven by control collapse rather than spatial selectivity (see text).}
\label{tab:anchor-summary}
\setlength{\tabcolsep}{2.5pt}
\begin{tabular}{l c c c}
\toprule
 & \textbf{BAGEL und} & \textbf{BAGEL gen} & \textbf{Qwen3-VL-8B} \\
 & (L21--26, hub) & (L21--27, hub) & (L0--8, hub) \\
\midrule
Regime & concentric & projected-diffuse & mixed \\
Mean $|\cos|$ in hub & 0.94 & 0.42 & 0.44 (layer-bimodal) \\
Cumulative EV at $k{=}1$ & 52.9\% & 18.9\% & 47.1\% \\
Cumulative EV at $k{=}16$ & 86.2\% & 69.5\% & 77.4\% \\
DD @ $k{=}1$ (pp) & $+1.7$ \scriptsize{[$-5.0$,$+8.3$]} & $-36.7$ \scriptsize{ctrl crash} & $-24.0$ \scriptsize{ctrl crash} \\
DD @ $k{=}16$ (pp) & $-12.7$ \scriptsize{[$-21.7$,$-4.0$] ctrl-dom.} & $-28.3$ \scriptsize{ctrl crash} & $-15.3$ \scriptsize{[$-24.0$,$-6.3$] ctrl-dom.} \\
1D causal carrier & none (all ranks non-sel.) & DiffMean (1D, $z{=}{+}3.19$) & per-layer DiffMean (1D) \\
Ctrl.-adj.\ slope (pp/unit) & --- & $-1.6$ & $+1.9$ \\
\bottomrule
\end{tabular}
\end{table}

\paragraph{Concentration diagnostic.}
Table~\ref{tab:concentration} reports the unified concentration diagnostic
used in the main text. For a unit axis $\hat a$, raw concentration is
computed as
\[
c_{\mathrm{raw}} =
\mathbb{E}_{x,t}
\left[
\frac{(h_{\ell}(x,t)^\top \hat a)^2}
{\|h_{\ell}(x,t)\|_2^2}
\right].
\]
We also report a centered version, $c_{\mathrm{cen}}$, computed after
removing the layer mean from the hidden states. The centered diagnostic
separates sample-varying signal from a global mean offset along the axis.

\begin{table}[h]
\centering
\small
\setlength{\tabcolsep}{4pt}
\renewcommand{\arraystretch}{1.05}
\caption{\textbf{Unified concentration diagnostic.}
$c_{\mathrm{raw}}$ measures total axis-aligned hidden-state energy, while
$c_{\mathrm{cen}}$ removes the layer mean before computing the same
quantity. Chameleon has low centered concentration but a selective causal
effect. BAGEL has very high raw concentration due to mean alignment, and
still remains above random after centering.}
\label{tab:concentration}
\begin{tabular}{lcccccc}
\toprule
\textbf{Model / axis} &
\textbf{Layer} &
$\mathbf{c_{\mathrm{raw}}}$ &
$\mathbf{c_{\mathrm{cen}}}$ &
$\mathbf{|\cos(\mu,a)|}$ &
\textbf{Random} &
$\mathbf{c_{\mathrm{cen}}/rand}$ \\
\midrule
Chameleon-7B / VL axis & L3 & $0.104\%$ & $0.380\%$ & $0.032$ & $0.018\%$ & $20.9\times$ \\
Chameleon-7B / VL axis & L4 & $0.174\%$ & $0.091\%$ & $0.042$ & $0.022\%$ & $4.1\times$ \\
Chameleon-7B / VL axis & L5 & $0.131\%$ & $0.137\%$ & $0.036$ & $0.026\%$ & $5.3\times$ \\
Chameleon-7B / VL axis & L6 & $0.215\%$ & $0.045\%$ & $0.047$ & $0.024\%$ & $1.9\times$ \\
Chameleon-7B / VL axis & L7 & $0.249\%$ & $0.038\%$ & $0.051$ & $0.024\%$ & $1.6\times$ \\
Chameleon-7B / VL axis & L20 ctrl & $0.123\%$ & $0.007\%$ & $0.038$ & $0.019\%$ & $0.4\times$ \\
\midrule
BAGEL-7B / gen DiffMean & L21 & $49.2\%$ & $0.426\%$ & $0.718$ & $0.029\%$ & $14.5\times$ \\
BAGEL-7B / gen DiffMean & L22 & $57.4\%$ & $0.929\%$ & $0.778$ & $0.029\%$ & $32.5\times$ \\
BAGEL-7B / gen DiffMean & L23 & $49.6\%$ & $1.522\%$ & $0.725$ & $0.028\%$ & $55.1\times$ \\
BAGEL-7B / gen DiffMean & L24 & $58.3\%$ & $3.163\%$ & $0.786$ & $0.027\%$ & $118\times$ \\
BAGEL-7B / gen DiffMean & L25 & $47.3\%$ & $2.298\%$ & $0.709$ & $0.026\%$ & $87.7\times$ \\
BAGEL-7B / gen DiffMean & L26 & $58.4\%$ & $0.936\%$ & $0.785$ & $0.026\%$ & $36.4\times$ \\
\bottomrule
\end{tabular}
\end{table}

For Chameleon, $|\cos(\mu,a)|$ is small, so raw and centered
concentrations are of the same order; the signal is not dominated by a
global mean offset. For BAGEL, $|\cos(\mu,a)|$ is large, so the raw
concentration is dominated by mean alignment. After centering, BAGEL still
has a stronger axis-aligned signal than Chameleon, but the gap is moderate
rather than several hundred-fold.

\subsection{Anchor 1 (concentric): BAGEL understanding path}
\label{app:anchor-und}

The understanding path of BAGEL exhibits the cleanest \emph{concentric} alignment: the DiffMean axis lies inside the dominant variance subspace ($|\cos|$ peaks at $0.978$ at L21 and stays $> 0.85$ through L26 before collapsing at L27). Despite this near-alignment with the top variance direction, the rank-1 PCA intervention is null (DD$=+1.7$ pp, CI $[-5.0, +8.3]$), and expanding the rank does not expose a selective handle: at $k{=}16$ (cumulative EV $86.2\%$), removal is clearly damaging but \emph{control-dominated} --- control tasks drop by $32.3$ pp versus $19.7$ pp for spatial tasks (DD$=-12.7$ pp, CI $[-21.7,-4.0]$). No tested rank yields spatial selectivity in this path.

\paragraph{Diagnostic implication.}
Probing accuracy is high at L27 (1D probe $0.729 \pm 0.017$, $+2.65\sigma$ above norm-matched random; full per-layer probe data in Table~\ref{tab:bagel-und-probe}; cross-architecture comparison with Chameleon-7B's complementary pattern in \S\ref{app:probe-vs-hub}), yet 1D intervention along the same direction is null and rank-$k$ removal is non-selective. This is the canonical \emph{probe--intervention dissociation} in its strongest form: linear decodability does not imply any linearly intervenable causal handle. The spatial information carried by this path is entangled with general representational capacity rather than organized as a removable carrier.

\subsection{Anchor 2 (projected-diffuse): BAGEL generation path}
\label{app:anchor-gen}

The generation path of BAGEL exhibits the \emph{projected-diffuse} pattern: the DiffMean axis sits well off the top PCA direction across the hub (per-layer $|\cos|$ between $0.26$ and $0.51$ over L21--L26, mean $0.42$; the all-layer mean is $0.19$, and L27 is a known LayerNorm-dominated outlier at $0.819$). The 1D DiffMean axis is nonetheless causal (DD $=+16.8$ pp, $z=+3.19$; main text \S\ref{sec:gen-vs-und}, Table~\ref{tab:dual_path_geometry}). The PCA subspace, in contrast, is broadly entangled with general representational capacity: rank-$k$ ablation collapses controls by $56$--$74$ pp at every $k$, more than the spatial tasks (DD $-16.7$ to $-39.0$), so removing variance-dominant subspaces produces broad damage rather than a selective effect.

\paragraph{Diagnostic implication.}
The two interventions probe orthogonal axes. The DiffMean intervention recovers a compact one-dimensional spatial carrier; the PCA intervention recovers a broader general-purpose subspace whose removal damages everything. The qualitative contrast between BAGEL's two paths --- concentric on und (1D null, rank-$k$ effect) vs.\ projected-diffuse on gen (1D causal, rank-$k$ non-selective) --- is the geometric core of the \emph{compression vs.\ distribution} hypothesis in main text \S\ref{sec:gen-vs-und}.

\subsection{Anchor 3 (mixed): Qwen3-VL-8B}
\label{app:anchor-mixed}

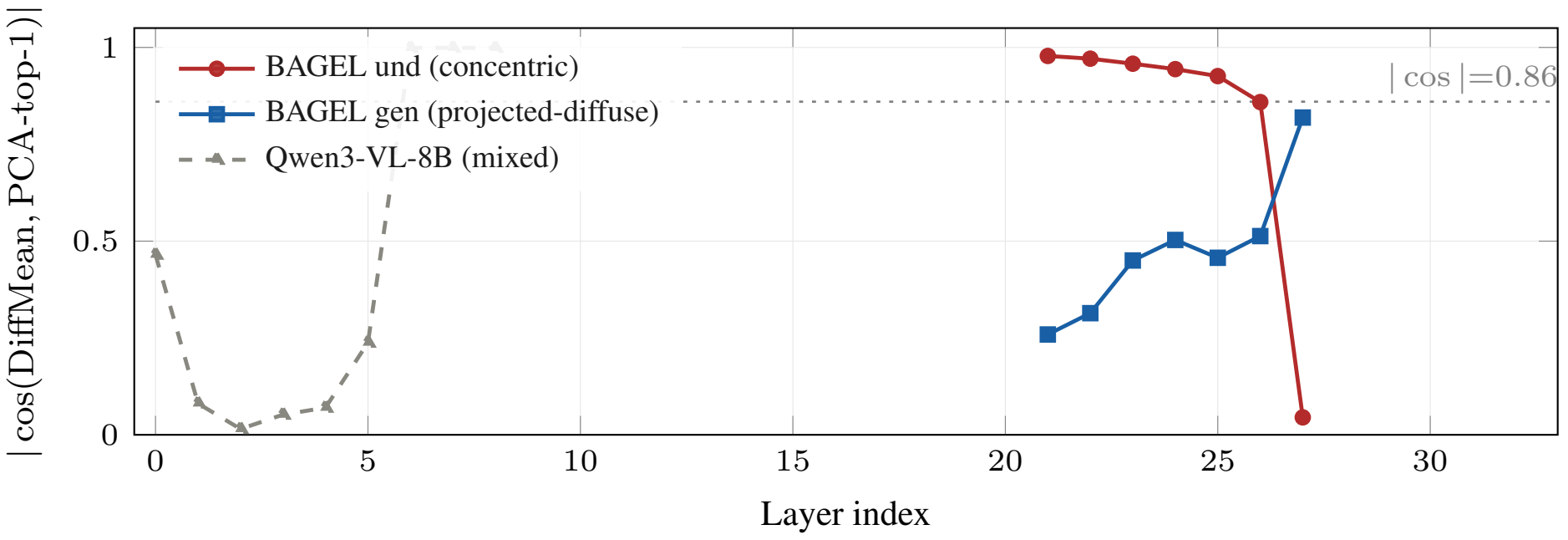
\begin{figure}[h]
\centering
\begin{tikzpicture}
\begin{axis}[
  width=0.92\linewidth, height=4.8cm,
  xlabel={\footnotesize Layer index},
  ylabel={\footnotesize $|\cos(\mathrm{DiffMean}, \mathrm{PCA\text{-}top\text{-}1})|$},
  xmin=-0.5, xmax=33,
  ymin=0, ymax=1.05,
  grid=major, grid style={black!10, very thin},
  tick label style={font=\scriptsize},
  label style={font=\footnotesize},
  legend style={font=\scriptsize, draw=none, fill=white, fill opacity=0.9, at={(0.02, 0.98)}, anchor=north west},
  legend cell align=left,
  every axis plot/.append style={mark size=1.4pt, line width=0.9pt}
]
\addplot[color=accent, mark=*] coordinates {
  (21, 0.978) (22, 0.971) (23, 0.958) (24, 0.944) (25, 0.926) (26, 0.859) (27, 0.045)
};
\addlegendentry{BAGEL und (concentric)}
\addplot[color=accent2, mark=square*] coordinates {
  (21, 0.259) (22, 0.314) (23, 0.450) (24, 0.503) (25, 0.457) (26, 0.513) (27, 0.819)
};
\addlegendentry{BAGEL gen (projected-diffuse)}
\addplot[color=neutral, mark=triangle*, dashed] coordinates {
  (0, 0.465) (1, 0.081) (2, 0.015) (3, 0.052) (4, 0.070) (5, 0.238) (6, 0.999) (7, 0.999) (8, 0.999)
};
\addlegendentry{Qwen3-VL-8B (mixed)}
\addplot[domain=0:33, samples=2, dotted, gray, line width=0.5pt] {0.86};
\node[font=\scriptsize, gray] at (axis cs:31, 0.92) {$|\cos|{=}0.86$};
\end{axis}
\end{tikzpicture}
\caption{\textbf{Per-layer $|\cos(\mathrm{DiffMean}, \mathrm{PCA\text{-}top\text{-}1})|$ for the three anchor cases.} BAGEL und stays near alignment ($|\cos|>0.85$) across the hub block before a layer-norm-dominated drop at L27 (excluded from hub mean). BAGEL gen remains well off the top PC across the hub (per-layer $0.26$--$0.51$, mean $0.42$; the all-layer mean is $0.19$). Qwen3-VL-8B is bimodal: near-orthogonal at L0--L5, saturated at L6--L8 corresponding to ViT injection. Dotted reference at $|\cos|{=}0.86$.}
\label{fig:anchor-cos}
\end{figure}

Qwen3-VL-8B is the \emph{mixed} case: $|\cos|$ is layer-dependent and bimodal (Figure~\ref{fig:anchor-cos}). At L0--L5 the DiffMean axis is nearly orthogonal to the top PCA direction ($|\cos|\in[0.015, 0.465]$); from L6 onwards $|\cos|$ jumps to $\approx 0.999$ and remains saturated through L33 before collapsing at L34. This bimodal profile is consistent with the architectural fact that the ViT projection feeds visual information into the LM stream around L6, after which the dominant variance direction and the modality-contrast direction converge.

\begin{figure}[h]
\centering
\begin{tikzpicture}
\begin{axis}[
  name=plotA,
  width=0.34\linewidth, height=4.5cm,
  title={\footnotesize\textbf{(A) BAGEL und}},
  title style={yshift=-2pt},
  xlabel={\footnotesize $k$},
  ylabel={\footnotesize DD (pp)},
  xmode=log, log basis x=2,
  xmin=0.85, xmax=38,
  ymin=-50, ymax=25,
  xtick={1,2,4,8,16,32}, xticklabels={1,2,4,8,16,32},
  grid=major, grid style={black!10, very thin},
  tick label style={font=\scriptsize},
  label style={font=\footnotesize},
  every axis plot/.append style={mark size=1.6pt, line width=0.9pt},
  axis line style={black!50}
]
\addplot+[color=accent, mark=*, error bars/.cd, y dir=both, y explicit] coordinates {
  (1, 1.67) +- (0, 6.66)
  (2, -4.67) +- (0, 8.17)
  (4, 5.67) +- (0, 6.0)
  (8, -9.33) +- (0, 7.83)
  (16, -12.67) +- (0, 8.83)
  (32, 3.67) +- (0, 8.5)
};
\addplot[domain=0.85:38, samples=2, dashed, gray, line width=0.5pt] {0};
\end{axis}
\begin{axis}[
  name=plotB, at={(plotA.east)}, anchor=west, xshift=10pt,
  width=0.34\linewidth, height=4.5cm,
  title={\footnotesize\textbf{(B) BAGEL gen}},
  title style={yshift=-2pt},
  xlabel={\footnotesize $k$},
  xmode=log, log basis x=2,
  xmin=0.85, xmax=38,
  ymin=-50, ymax=25,
  xtick={1,2,4,8,16,32}, xticklabels={1,2,4,8,16,32},
  yticklabels={},
  grid=major, grid style={black!10, very thin},
  tick label style={font=\scriptsize},
  label style={font=\footnotesize},
  every axis plot/.append style={mark size=1.6pt, line width=0.9pt},
  axis line style={black!50}
]
\addplot+[color=accent2, mark=square*, error bars/.cd, y dir=both, y explicit] coordinates {
  (1, -36.67) +- (0, 7.67)
  (2, -38.0) +- (0, 7.5)
  (4, -16.67) +- (0, 8.34)
  (8, -31.33) +- (0, 8.34)
  (16, -28.33) +- (0, 8.34)
  (32, -39.0) +- (0, 8.67)
};
\addplot[domain=0.85:38, samples=2, dashed, gray, line width=0.5pt] {0};
\end{axis}
\begin{axis}[
  name=plotC, at={(plotB.east)}, anchor=west, xshift=10pt,
  width=0.34\linewidth, height=4.5cm,
  title={\footnotesize\textbf{(C) Qwen3-VL-8B}},
  title style={yshift=-2pt},
  xlabel={\footnotesize $k$},
  xmode=log, log basis x=2,
  xmin=0.85, xmax=38,
  ymin=-50, ymax=25,
  xtick={1,2,4,8,16,32}, xticklabels={1,2,4,8,16,32},
  yticklabels={},
  grid=major, grid style={black!10, very thin},
  tick label style={font=\scriptsize},
  label style={font=\footnotesize},
  every axis plot/.append style={mark size=1.6pt, line width=0.9pt},
  axis line style={black!50}
]
\addplot+[color=neutral!70!black, mark=triangle*, error bars/.cd, y dir=both, y explicit] coordinates {
  (1, -24.0) +- (0, 7.67)
  (2, -6.33) +- (0, 9.5)
  (4, 0.67) +- (0, 9.0)
  (8, 2.33) +- (0, 8.17)
  (16, -15.33) +- (0, 8.84)
  (32, -21.0) +- (0, 8.67)
};
\addplot[domain=0.85:38, samples=2, dashed, gray, line width=0.5pt] {0};
\end{axis}
\end{tikzpicture}
\caption{\textbf{DD vs.\ rank-$k$ subspace ablation, three anchor cases}
(DD $=\Delta_{\text{control}}-\Delta_{\text{spatial}}$; positive would
mean spatial-selective damage). Error bars are paired-bootstrap $95\%$
CIs. \textbf{(A)} BAGEL und: null at $k{=}1$; at $k{=}16$ removal is
damaging but control-dominated ($-12.7$ pp). \textbf{(B)} BAGEL gen:
$\Delta_{\text{control}} \leq -56$ pp at every $k$ --- broad collapse,
not a selective effect (the selective handle is the 1D DiffMean axis,
main text \S\ref{sec:gen-vs-und}). \textbf{(C)} Qwen3-VL-8B:
non-monotonic but never spatially selective; the $k{=}1$ dip is driven
by control collapse and $k\in\{16,32\}$ remain control-dominated.}
\label{fig:anchor-dd}
\end{figure}
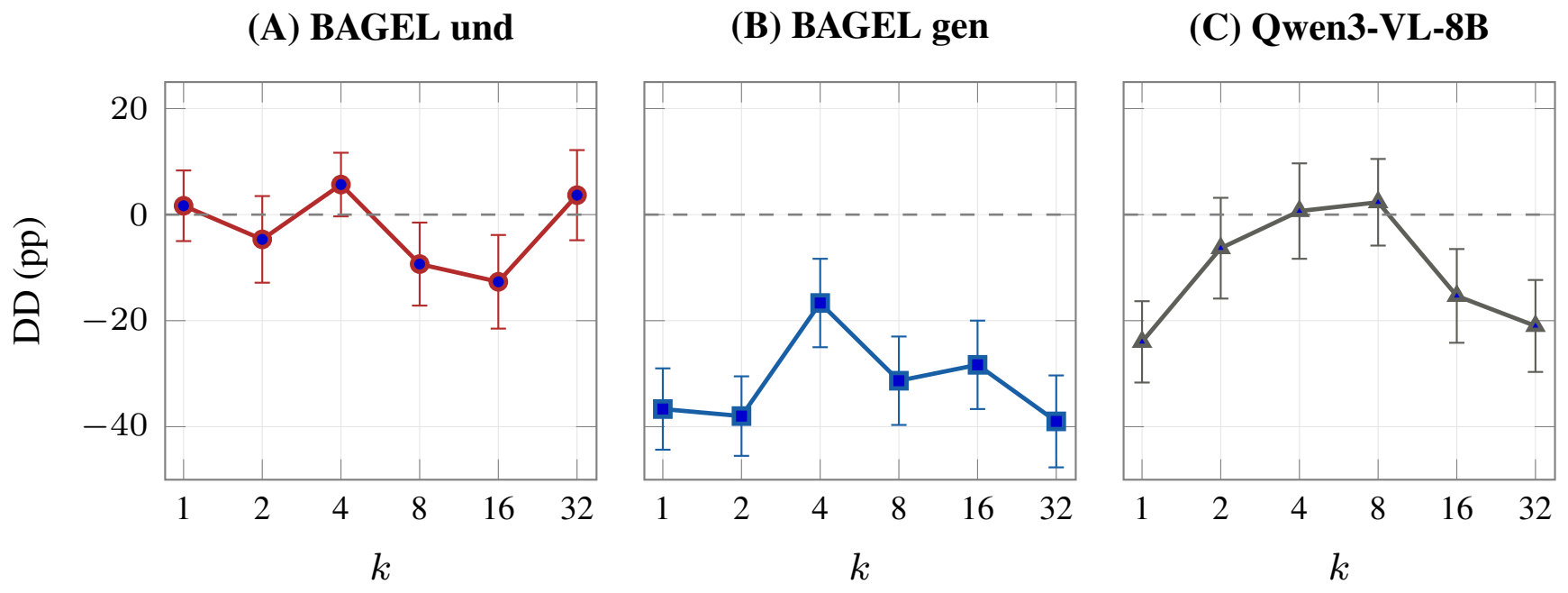

The rank-$k$ DD curve is correspondingly non-monotonic (Figure~\ref{fig:anchor-dd}, panel C), but at no rank does it become spatially selective. At $k{=}1$, controls collapse ($\Delta_{\text{control}}=-65$ pp; DD$=-24$). At $k{=}4$ and $k{=}8$, effects are small and near-symmetric (DD $+0.7$/$+2.3$). At $k{=}16$ and $k{=}32$, controls again drop more than spatial tasks (DD $-15.3$ and $-21.0$). The selective handle in Qwen3-VL-8B is therefore its per-layer 1D DiffMean axis (Table~\ref{tab:main-effect}), not any variance-dominant PCA subspace.

\paragraph{Diagnostic implication.}
A model can simultaneously satisfy ``concentric'' alignment in some layers and ``projected-diffuse'' in others. When evaluating a new architecture, neither the rank-1 result nor any single rank-$k$ value is decisive; the non-monotonic, never-selective DD-vs-$k$ curve is itself diagnostic of the mixed regime, and rank-$k$ removal should always be checked against control damage before any selectivity claim.

\subsection{Figures}
\label{app:anchor-figs}

\subsection{Scope of this analysis}
\label{app:anchor-scope}

The three anchor cases establish that the alignment regimes named in \S\ref{sec:representation-properties} are realized geometries with testable diagnostic signatures, not abstract possibilities. The methodological scope of this appendix is to characterize the regimes in detail through these anchors; a complete taxonomy of all $10$ configurations into regimes is a separate question and lies outside the present paper's scope. Within our framework, each new architecture can be classified by running the EXP4 phase-0 and phase-1 protocols.

\end{document}